\RequirePackage{fix-cm}
\documentclass[twocolumn]{svjour3} 
\smartqed  
\usepackage{graphicx}
\usepackage{subfigure}
\usepackage{amssymb}
\usepackage{amsmath}
\usepackage{algorithm}
\usepackage{algorithmic}
\usepackage[utf8]{inputenc}
\usepackage{url}
\DeclareMathOperator*{\argmin}{arg\,min}

\begin{document}

\title{The Robot Skin Placement Problem: a New Technique to Place Triangular Modules inside Polygons}
\titlerunning{A New Technique to Place Triangular Modules inside Polygons}

\author{Xuenan Guo \and Fulvio Mastrogiovanni}
\authorrunning{Guo and Mastrogiovanni}

\institute{
X. Guo
\at
Department of Computer Science and Technology, Tsinghua University\\
Beijing, China\\
\email{gxn12@mails.tsinghua.edu.cn}
\and
F. Mastrogiovanni
\at
Department of Informatics, Bioengineering, Robotics and Systems Engineering, University of Genoa\\
Genoa, Italy\\
\email{fulvio.mastrogiovanni@unige.it}
}

\date{Received: date / Accepted: date}

\maketitle

\begin{abstract}
Providing robots with large-scale robot skin is a challenging goal, especially when considering surfaces characterised by different shapes and curvatures.
The problem originates from technological advances in tactile sensing, and in particular from two requirements: (i) covering the largest possible area of a robot's surface with tactile sensors, and (ii) doing it using cheap, replicabile hardware modules.
Given modules of a specific shape, the problem of optimally placing them requires to maximise the number of modules that can be fixed on the selected robot body part. 
Differently from previous approaches, which are based on methods inspired by computational geometry (e.g., \emph{packing}), we propose a novel layout design method inspired by physical insights, referred to as \textit{Iterative Placement} (\textsc{ItPla}), which arranges modules as if physical forces acted on them.
A number of case studies from the literature are considered to evaluate the algorithm.
\end{abstract}


\section{Introduction}
\label{sec:introduction}

In the past few years, human-robot interaction (HRI) emerged as a major research field and design paradigm to enable novel robot behaviours, considering aspects related to perception, reasoning and action \cite{GoodrichSchultz2007}\cite{Dautenhahn2007}.
Two research strands can be identified.
On the one hand, \textit{cognitive} HRI aims at understanding how human and robot behaviours can be integrated to obtain advanced forms of assistance or cooperation, e.g., in case of household robots \cite{Mataric2006}\cite{Darvishetal2017}.
On the other hand, \textit{physical} HRI investigates how humans and robots can exploit haptic information to carry out a set of tasks where physical contact is essential, e.g., for robot co-workers \cite{ArgallBillard2010}\cite{Beetzetal2015}.

In physical HRI, special attention is devoted to the use of large-scale robot skin to provide robots with tactile information originating from their whole body. 
The \textit{sense of touch} is expected to play a fundamental role and to act as a key enabling technology to implement novel robot behaviours, specifically in humanoid robots employed in service scenarios \cite{Dahiyaetal2010}\cite{Deneietal2010}\cite{Deneietal2015}.

Although different technological solutions to deploy large-scale robot skin have been presented (please refer to \cite{DahiyaValle2013} and the references therein for a comprehensive analysis of the literature), only a few approaches consider general-purpose methods adaptable to different robots and robot's surfaces.
Modular robot skin designs are a good trade-off between \textit{cost} and \textit{quality}, specifically in terms of covered robot's surface \cite{Schmitzetal2011}\cite{Mittendorferetal2015}, adaptability of models for data processing \cite{Muscarietal2013}\cite{Deneietal2015}\cite{Kabolietal2015}, as well as hardware and software infrastructure \cite{Youssefietal2015a}\cite{Youssefietal2015b}. 
Among these solutions, ROBOSKIN \cite{Schmitzetal2011}\cite{Maiolinoetal2013} proves to be adaptable to different robot mechanical designs and shapes \cite{Maiolinoetal2013b}\cite{Baglinietal2014}\cite{Maiolinoetal2015}.  
ROBOSKIN modules are triangles of equal side, connected together to form an integrated mesh with just a limited number of entry points.

\begin{figure}[t!]
	\centering
	\subfigure[]{
		\includegraphics[width=0.475\hsize]{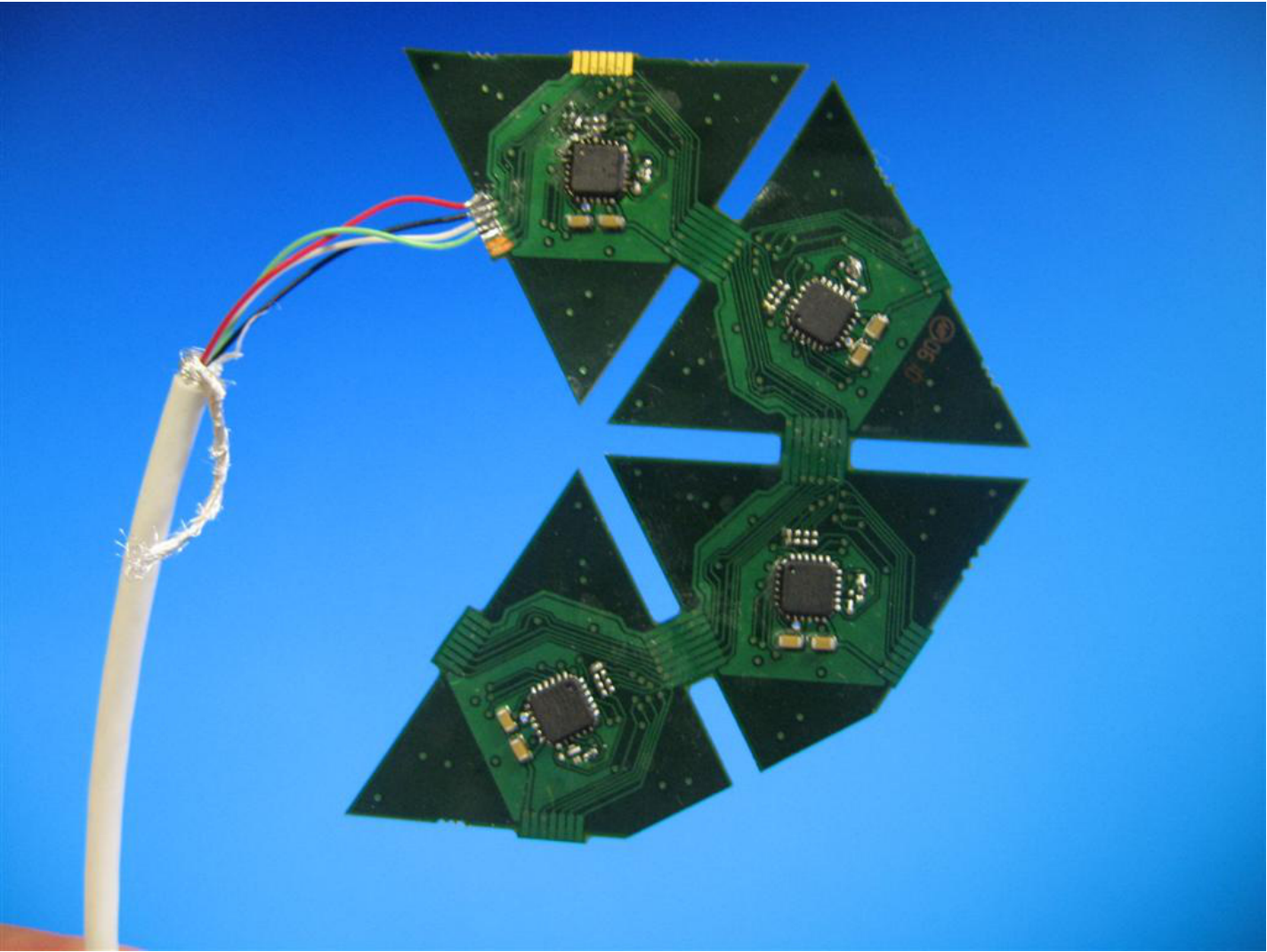}
	}
	\subfigure[]{
		\includegraphics[width=0.432\hsize]{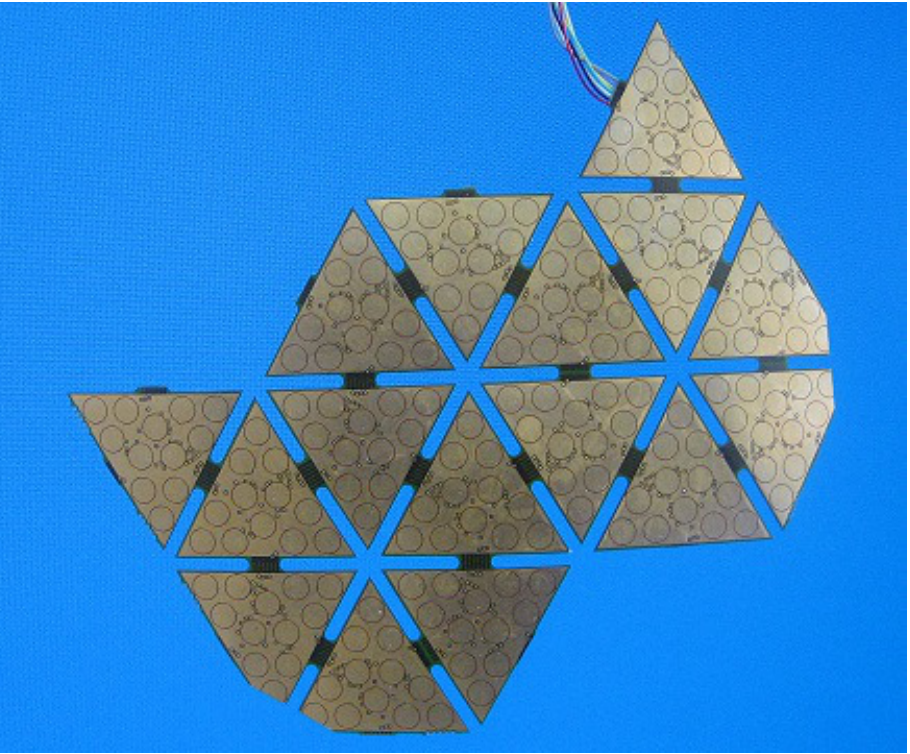}
	}
	\caption{Two examples of ROBOSKIN layouts: (a) bottom view of layout for a robot's hand palm; (b) top view of a layout for a robot's torso coverage.}
	\label{fig:roboskin_examples}
\end{figure}
In this paper we consider the problem of \emph{optimally placing robot skin modules on a given robot's surface}, a problem defined in \cite{Anghinolfietal2013} and further developed in \cite{Weietal2015}\cite{Weietal2017}.
We refer to the ROBOSKIN technology and, as a consequence, we consider triangular modules (Figure \ref{fig:roboskin_examples}).
However, the technique we present in this paper is not specific to the triangular shape, and an extension to different shapes can be derived.
In the problem we consider, \textit{optimality} refers to the actual amount of robot's surface covered by robot skin, subject to a number of mechanical and design constraints characterising it.
The typical mathematical formulation induces a NP-HARD class problem, for which a closed-form solution in polynomial time subject to certain conditions has been derived in \cite{Weietal2017}.

Apart from related problems, i.e., \emph{point placement} \cite{Eades1984} and \emph{microchip placement} \cite{QuinnBreuer1979}, only a few techniques have been proposed to address the problem we consider in this paper.
Anghinolfi and colleagues first define the \emph{robot skin placement} problem and solve it using a set of heuristics, which are based on placing a polygon (representing the border of the robot's surface to cover, which has been appropriately flattened) over an isometric grid, in order to maximise the number of grid's cells contained by the polygon itself \cite{Anghinolfietal2013}.
They show that this formulation is equivalent to fit the highest number of triangles on the robot's surface.
Although they propose seven heuristics (integrated in a multi-start approach), they argue that one seems to always outperform the others.
In their approach, since the isometric grid defines the mutual position of triangular modules, the relative displacement is fixed and rigid.
Furthermore, all modules in the resulting placement must be strictly located \textit{inside} the polygon's border.
As a consequence, the approach by Anghinolfi and colleagues does not take into account any \emph{soft} violation of the geometrical constraints.
An extension to the original algorithm has been proposed by Wei and colleagues \cite{Weietal2015}\cite{Weietal2017}, who adopt a greedy strategy to improve the original results.
Their algorithm repeats the polygon placement steps several times.
Each time, it gets the most \textit{stable} placement part and consider it as part of the final placement, and separates the corresponding area from the polygon's border.
The relative rigid displacement between modules is relaxed, and this leads to better placement results.
However, the approach by Wei and colleagues cannot tolerate constraint violations, and because of the adopted greedy strategy, a number of (possibly good) placements may not be generated.

The paper by Wang and colleagues \cite{Wangetal2014}, although not directly related to the problem we consider here, shares some similarity with our work.
They tackle a triangle-based \textit{packing} problem, where an area must be filled with the maximum number of triangles allowed by the area size.
According to their approach, triangles enter the area one at a time, and they are positioned considering also the locations of already packed triangles.
If no triangles can be added anymore, the packing process is terminated.
However, there are important differences with respect to our work:
(i) we allow triangles to reposition (according to certain conditions) before the final placement is reached at any time, whereas Wang and colleagues fix the position of each triangle, incrementally;
(ii) we set a few rules to move triangles in order to obtain the final placement, whereas Wang and colleagues use strategies inspired by the way humans would perform the packing problem.

The main contribution of this paper is a new physics-based placement algorithm for the automated layout design of large-scale robot skin, referred to as Iterative Placement (\textsc{ItPla}), which is available open source\footnote{Web: \url{https://github.com/boy2000-007man/ITPLA}.}.
Given generic robot body parts, a general-purpose process is described to optimally place robot skin modules to cover such body parts.
Each module is subject to a number of pseudo-forces, which move it according to mutual positions of other modules.

The paper is organised as follows.
Section \ref{sec:problem} introduces the automated robot skin placement problem and outlines the proposed algorithm, and it poses a number of definitions.
Section \ref{sec:circ} details a simpler version of our problem, which is based on circular modules, to describe basic concepts.
Section \ref{sec:tri} discusses our solution to the robot skin placement problem for triangular modules.
Case studies are discussed in Section \ref{sec:cases}.
Conclusions follow.



\section{The Robot Skin Placement Problem}
\label{sec:problem}

\subsection{Constraints and Problem Statement}

The robot skin placement problem for triangular modules has been introduced in \cite{Anghinolfietal2013}. 
As previously pointed out, different aspects of the problem have been discussed also in \cite{Anghinolfietal2012}\cite{Anghinolfietal2014}.
The problem originates from novel technological solutions to the development of large-scale robot skin.
ROBOSKIN \cite{Schmitzetal2011}\cite{Maiolinoetal2013} is a modular robot skin where each module, of triangular shape, has a $3$ $cm$ side and hosts a number of tactile elements (i.e., \textit{taxels}) representing discrete sensing locations.
Triangular modules are mechanically connected with each other at the mid points of their sides.
Such connections can be slightly bent (or even cut) but cannot be stretched nor twisted in an operational arrangement of modules (i.e., a \textit{patch}).
Therefore, triangular modules must not overlap (constraint $C_1$) nor slide ($C_2$) with each other because this would twist the connectors to a breaking point.
It is noteworthy that connectors host wires to allow data signals to be exchanged between any two modules. 

The whole problem can be divided in three steps, namely (i) \textit{body part surface flattening}, in which a 2D version of the 3D surface to cover is obtained through a proper topographic mapping in the form of a (possibly distorted) polygon $P$, (ii) \textit{optimal body part coverage}, which aims at maximising the area of $P$ covered with robot skin, i.e., finding the maximum number of triangles contained in $P$, and (iii) \textit{optimal data signals routing in patches}, where data buses are routed via connectors to enforce redundancy and fault tolerance \cite{Anghinolfietal2012}. 

As described in \cite{Anghinolfietal2013}, the problem of determining the optimal coverage when fixing robot skin on robot's body parts can be modelled as finding an appropriate placement of a polygon (representing a \emph{flattened} version of the surface to cover) over an isometric grid, subject to the two constraints $C_1$ and $C_2$ outlined above, to maximise the number of triangles which can be contained in it.
As discussed in \cite{Anghinolfietal2013}, this proves to be equivalent to cover the original 3D surface in so far as the distortions induced by the flattening procedure are minimised. 
In such a model, the isometric grid represents an arrangement of triangles of equal side consistent with the ROBOSKIN layout.
In this paper, we focus on the second step outlined above, i.e., optimal body part coverage.

\subsection{Outline of the Iterative Placement Algorithm}
In this Section, we formalise the problem we address more precisely.
In our case, \textit{optimality} refers to the amount of robot's surface that is actually covered.
In \emph{Iterative Placement}, body part coverage is divided in two steps:
first, the algorithm \emph{places} the highest possible number of triangles inside $P$ (computed based on the ratio between the polygon's and a triangle's area);
second, it iteratively \textit{adjusts} triangle's poses to satisfy constraints $C_1$ and $C_2$.

\begin{figure}[t]
	\centering
	\includegraphics[width=0.7\hsize]{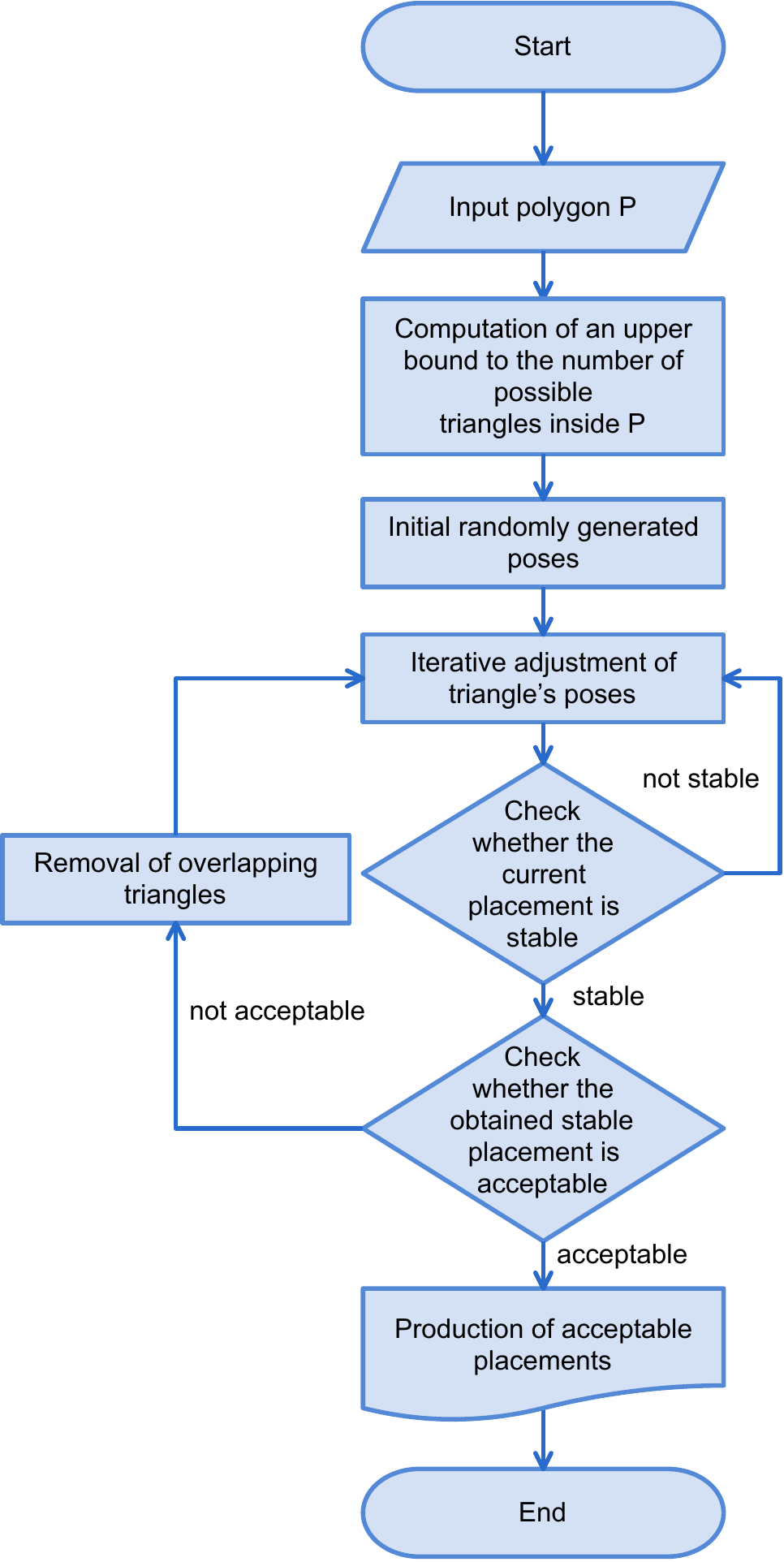}
	\caption{A flowchart of the \textsc{ItPla} algorithm.}
	\label{fig:procedure}
\end{figure}

Before outlining the overall algorithm's behaviour, a few informal definitions must be introduced.
With triangle's \emph{pose}, we refer to the location and orientation of a triangle-centred reference frame with respect to an external (i.e., absolute) reference frame. 
Given a polygon $P$, we refer to a \textit{placement} as a configuration of any number of triangle's poses within $P$.
We label as \emph{acceptable} a placement free from mutual overlaps between triangles, and as \emph{stable} a placement where triangle's poses tend to stabilise as a result of an iterative pose adjustment process aimed at satisfying constraints $C_1$ and $C_2$.
Two notions of stability are used in \textsc{ItPla}. 
A placement is \emph{absolutely stable} if poses are not updated in the iterative process, whereas we refer to an \emph{approximately stable} placement as a placement where each pose varies of a very small amount (i.e., the displacement is below a given threshold), which is negligible in practice. 
The steps of \textsc{ItPla} are described as follows (Figure \ref{fig:procedure}).

\begin{figure*}[t!]
	\centering
	\subfigure[]{
		\includegraphics[width=0.18\hsize]{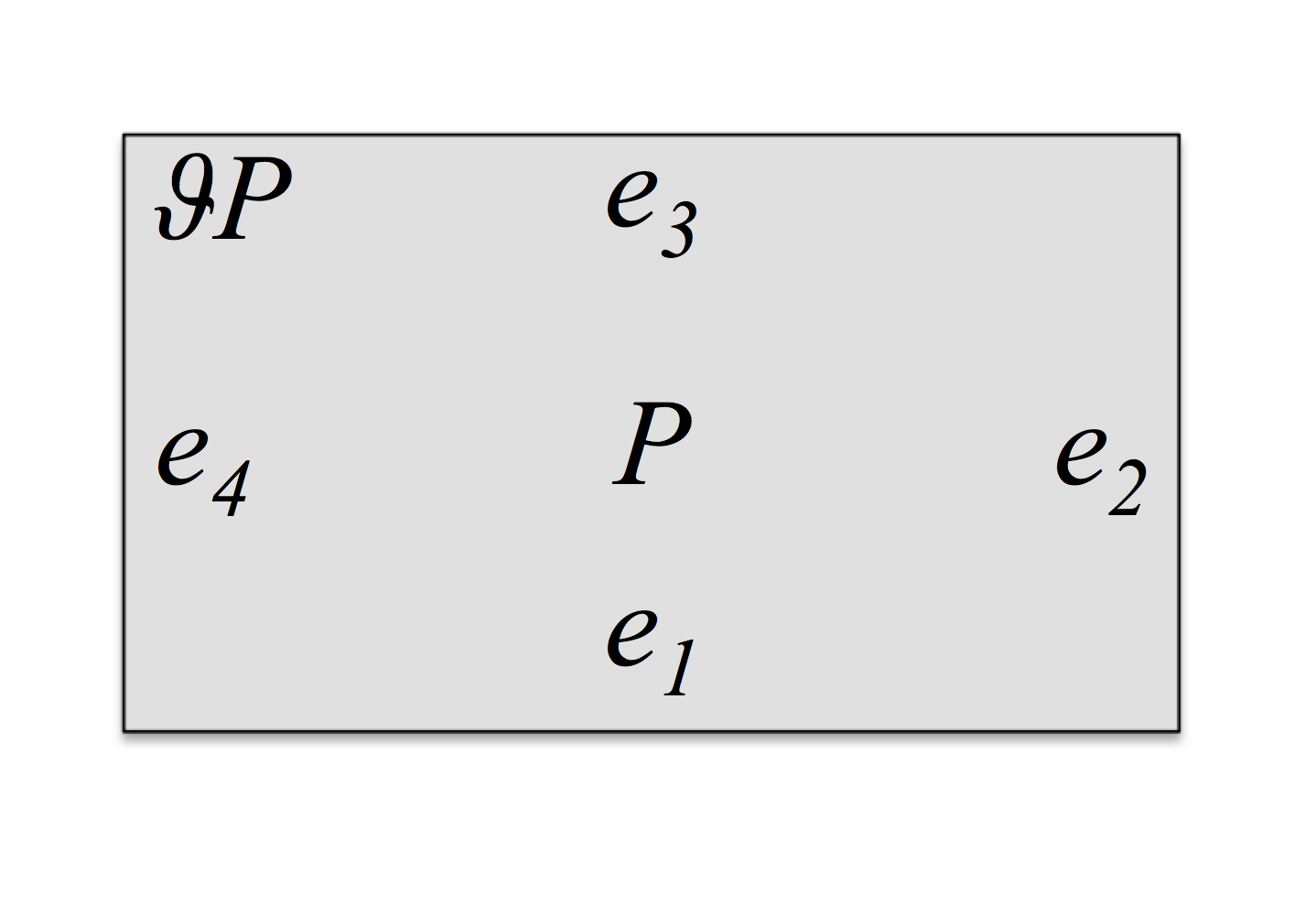}
	}
	\subfigure[]{
		\includegraphics[width=0.18\hsize]{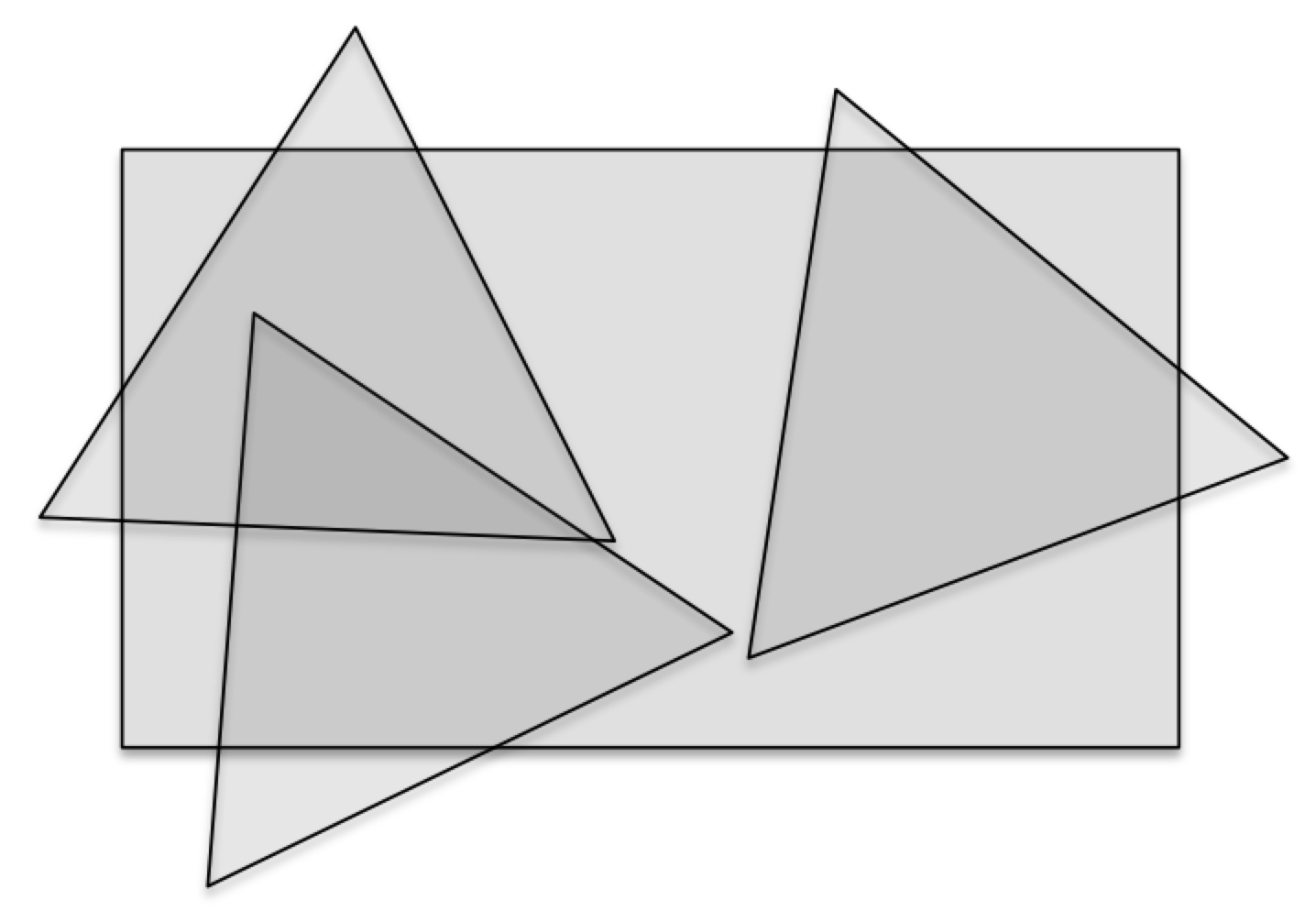}
	}
	\subfigure[]{
		\includegraphics[width=0.18\hsize]{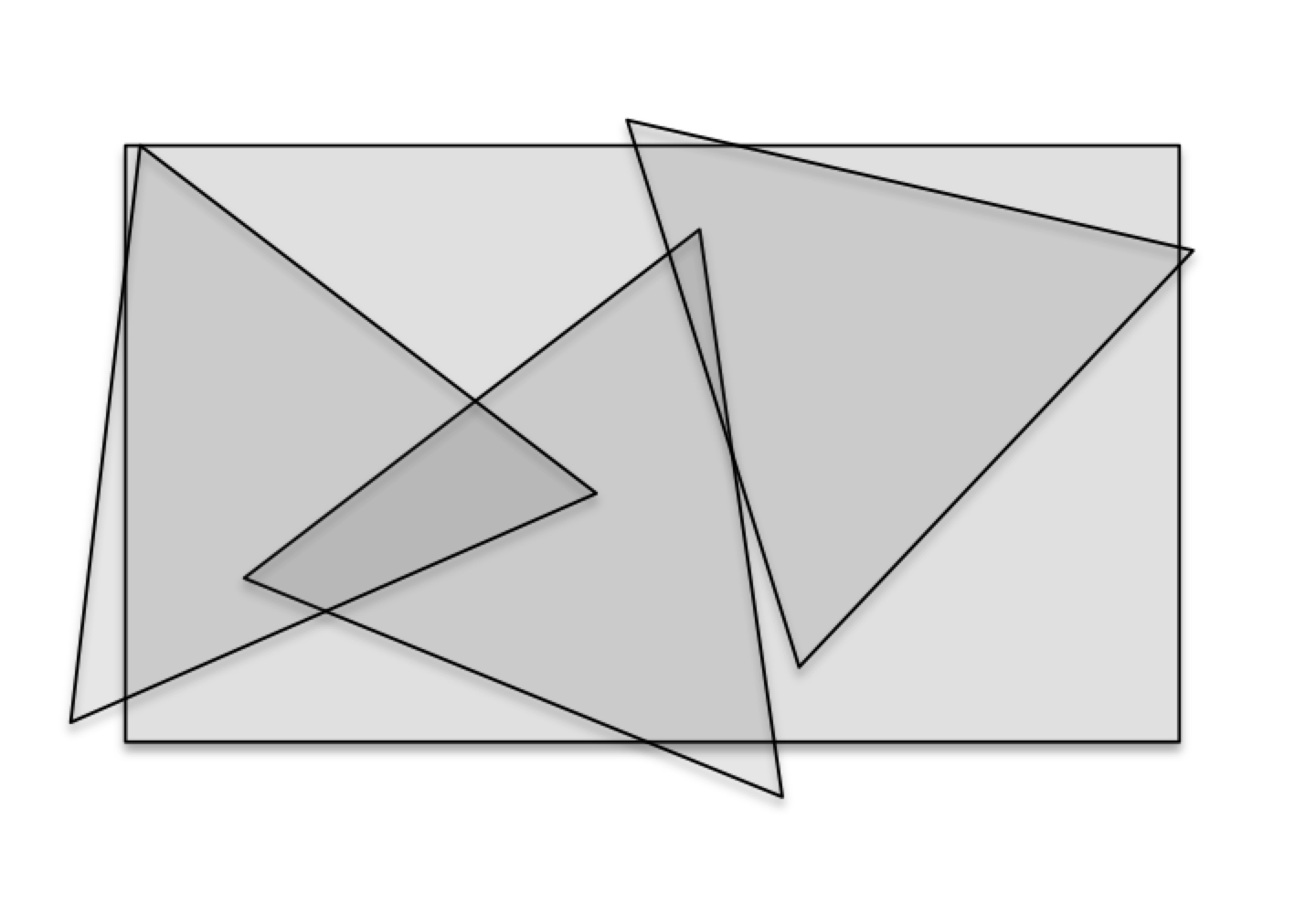}
	}
	\subfigure[]{
		\includegraphics[width=0.18\hsize]{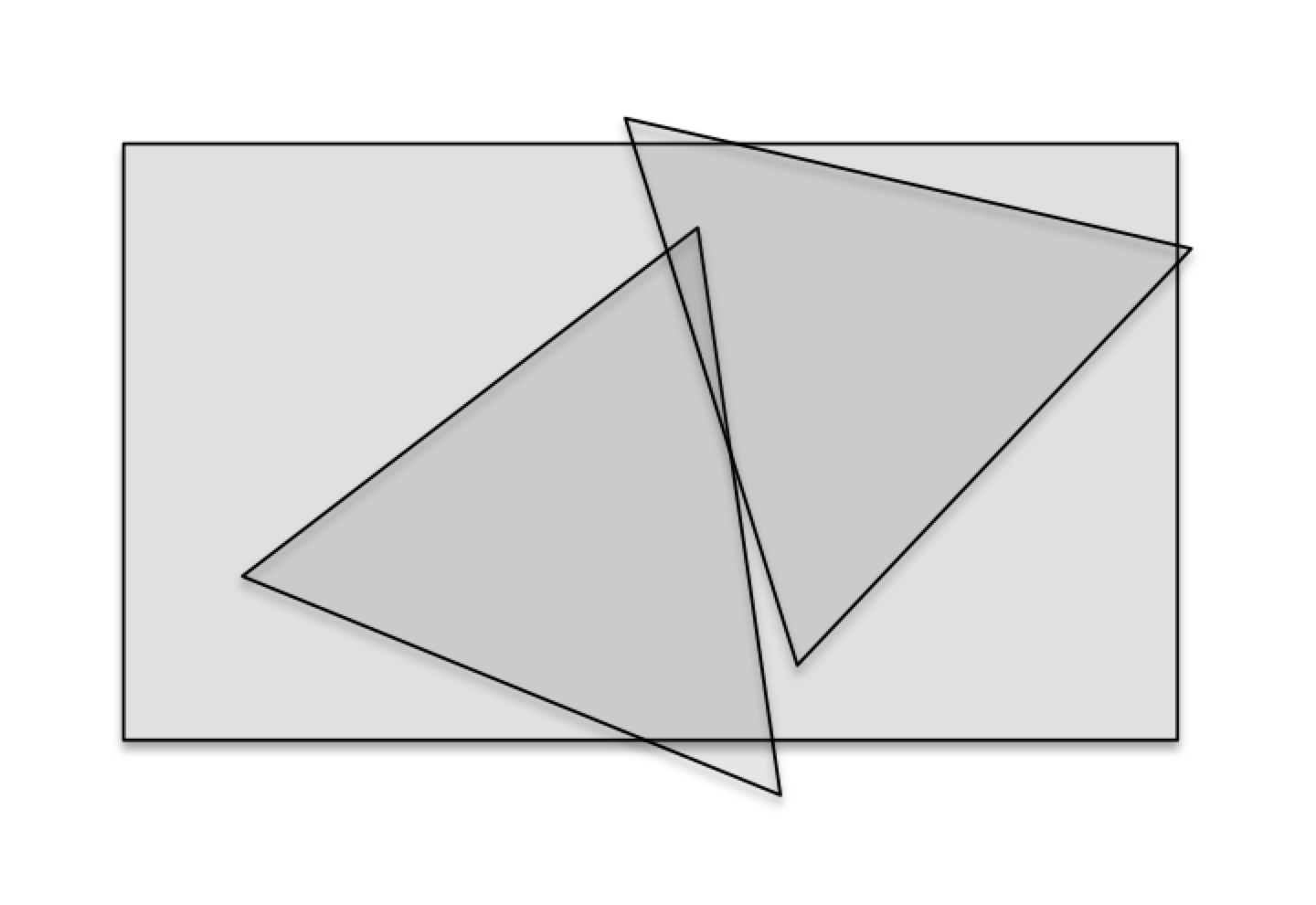}
	}
	\subfigure[]{
		\includegraphics[width=0.18\hsize]{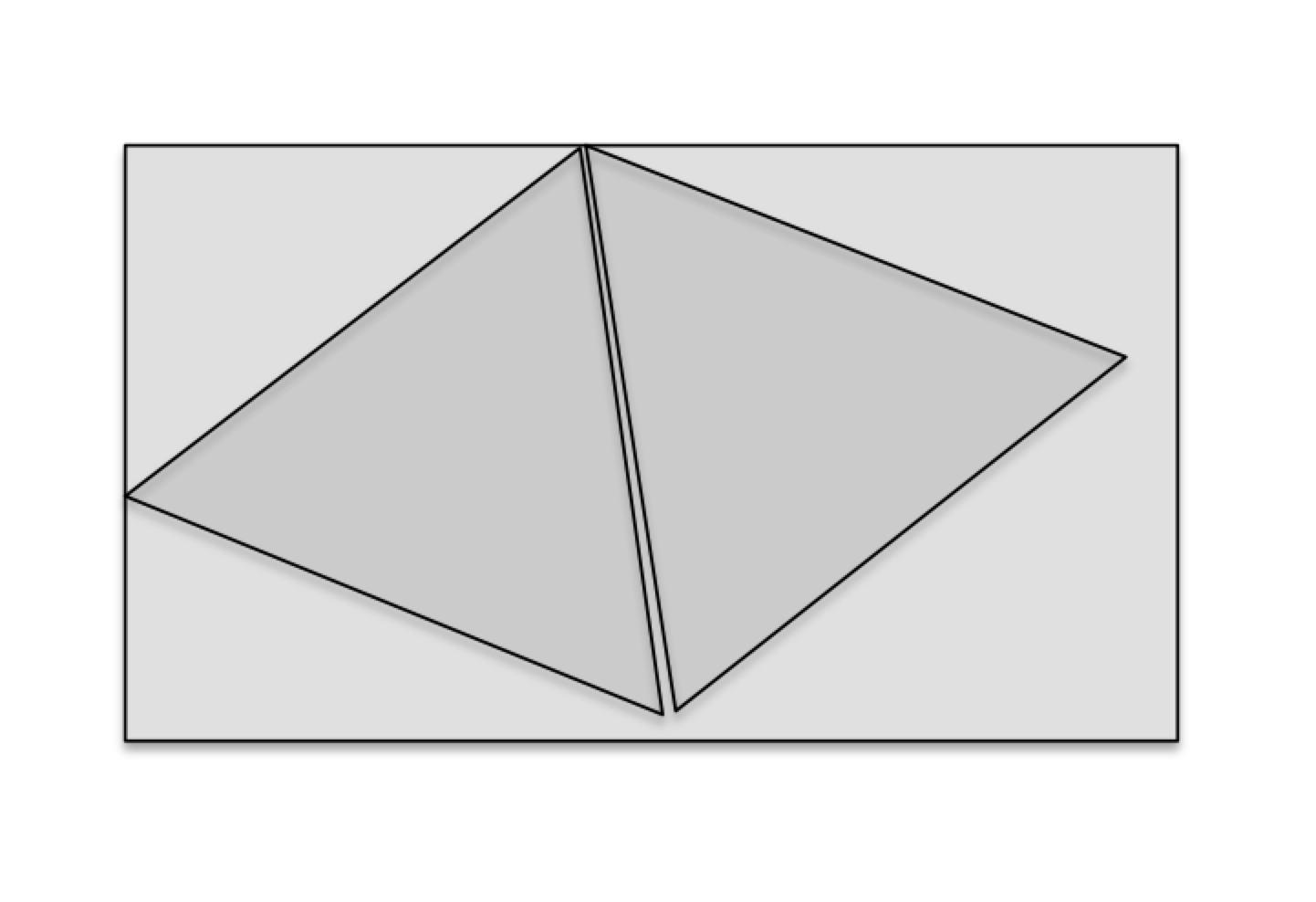}
	}
	\caption{An example of the whole coverage process: (a) a polygon $P$; (b) triangles are randomly placed in $P$; (c) an invalid placement is generated; (d) an overlapping triangle is removed; (e) a valid placement is obtained.}
	\label{fig:whole_process_example}
\end{figure*}
\begin{enumerate}
\item \textit{Computation of an upper bound to the number of possible triangles inside} $P$ \textit{and initial randomly generated poses}.
Given a polygon $P$ (Figure \ref{fig:whole_process_example}a), the upper bound is computed using the ratio between the areas of $P$ and a module's.
A corresponding number of triangles is placed randomly inside $P$ (Figure \ref{fig:whole_process_example}b).
It is noteworthy that, initially, triangles can overlap with each other as well as with the border of $P$.
The final result of the process may vary according to the initial placement.
In practice, it is possible to adopt a \emph{multi-start approach} and select the best placement result.
\item \textit{Iterative adjustment of triangle's poses}.
Given the initial placement, \textsc{ItPla} adjusts poses to obtain an overall acceptable placement.
The algorithm proceeds by iteratively arranging triangles using pseudo-forces (i.e., associating each triangle's reference frame with a velocity vector), therefore making triangles translate and rotate.
Pseudo-forces are designed to \emph{tend} to reduce the overlap between triangles and to properly align them.
As a consequence, it can be noticed that solutions are not guaranteed to satisfy the two constraints $C_1$ and $C_2$.
\item \textit{Check whether the current placement is stable}.
After a number of iterations, triangle's poses will tend to balance each other and stabilise (Figure \ref{fig:whole_process_example}c).
In ideal conditions (and with a theoretically infinite number of iterations), such a force balance leads to an absolutely stable placement. 
However, in order to obtain realistic solutions in a limited amount of time, we deem sufficient to consider placements in which approximate stability is reached.
This is a good trade-off between placement stability, solution accuracy and processing time. 
To this aim, \textsc{ItPla} employs simulated annealing to determine whether the placement is approximately stable according to a metric related to the degree of triangles overlap.
\item \textit{Check whether the obtained stable placement is acceptable}.
After a stable placement is obtained, \textsc{ItPla} evaluates whether it is acceptable.
Acceptability mainly depends on modelled physical constraints between triangular modules. 
In the reference robot skin technology, constraints are related to mechanical connectors, which are located at the middle of triangular module's sides.
Furthermore, end corners of triangular modules can be cut to a certain extent to accommodate for situations in which triangles overlap with borders of $P$.
Since $P$ originates from a 3D surface, it is subject to distortions.
As a consequence, the algorithm considers the placement to be acceptable when the amount of overlap between neighbouring triangles is below a given threshold.
\item \textit{Removal of overlapping triangles}.
When a stable placement is not acceptable, a triangle in overlap is chosen and removed (Figure \ref{fig:whole_process_example}d).
Then, a new iterative adjustment process starts considering the remaining triangles until a final acceptable placement is obtained.
\item \textit{Production of acceptable placements}.
When a stable placement is acceptable, it is shown to the user for evaluation (Figure \ref{fig:whole_process_example}e).
However, if a multi-start approach is employed, \textsc{ItPla} starts again from step $2$, until the selected number of algorithm iterations is reached.
\end{enumerate}

\subsection{Preliminary Definitions}

The unfolded surface of the robot body part to cover is modelled as a \emph{simple} and closed polygon $P$, whose boundary $\partial P$ is characterised by a finite set $\mathbb{E}_P$ of $\phi$ edges, such that $\mathbb{E}_P = \{e_1, \ldots, e_\phi\}$ (Figure \ref{fig:whole_process_example}a).
A polygon is labelled as \emph{simple} if its edges do not pairwise intersect, with the obvious exception of their vertexes. 
Since the goal of the robot skin placement is to cover as much of a robot's surface as possible, it will be convenient to refer to $\mathcal{A}(\cdot)$ as the area of any generic planar shape.
$P$ must be covered with the highest number $\mathcal{B}^*_P$ of modules, such that $\mathcal{B}^*_P \leq \mathcal{B}_P$, where $\mathcal{B}_P$ is a theoretical \emph{upper bound} to the number of possible modules fitting in, defined as the floor of the ratio between the areas of $P$ and a module's $m$, as follows:
\begin{equation}
\mathcal{B}_P = \left\lfloor\frac{\mathcal{A}(P)}{\mathcal{A}(m)}\right\rfloor.
\label{eq:UpperBound}
\end{equation}
For instance, with respect to Figure \ref{fig:whole_process_example}b, the upper bound $\mathcal{B}_P$ is equal to $3$.
In this paper, we refer to the set of modules covering $P$ as $\mathbb{S}_P = \{m_1, \ldots, m_{\mathcal{B}^*_P}\}$.
Given a module $m$, we refer to its \emph{pose} as $p_m$.
A placement $\mathbb{P}$ of a polygon $P$ can therefore be expressed as a set of module's poses, as follows:
\begin{equation}
\mathbb{P}_P = \{p_{m_1}, \ldots, p_{m_{\mathcal{B}^*_P}}\}.
\end{equation}

The physical constraints $C_1$ and $C_2$ described above, which are imposed by the target robot skin technology, are modelled in \textsc{ItPla} as rules in the algorithm.

From constraint $C_1$, given a polygon $P$, we have that modules belonging to $\mathbb{S}_P$ should not overlap with each other (Figure \ref{fig:whole_process_example}c).
However, since $P$ is a flattened planar representation of a robot body part, it is likely that $P$ is distorted with respect to the original 3D surface, either when unfolding \cite{Anghinolfietal2013} or surface parameterisation techniques are used \cite{Deneietal2015}.
Therefore, results where small overlaps are present may still be acceptable in practice (Figure \ref{fig:whole_process_example}d).
In case even minor overlaps cannot be accepted, the reference technology may still allow one to cut (to a limited extent) a module's end corner, which may solve a few issues associated with overlapping.  
In \textsc{ItPla}, we introduce an overlapping threshold $\tau_o$, which is an upper bound to the overlapping degree of any acceptable placement.

From constraint $C_2$, module's layouts are hard constraints.
However, for the same reasons considered for $C_1$, small violations of these constraints still may lead to acceptable solutions in practice.
\textsc{ItPla} minimises misplacements among neighbouring modules, and uses a misplacement threshold $\tau_m$ for acceptable placements.

The two thresholds $\tau_o$ and $\tau_m$
may be tuned to allow for more or less permissive violations of the constraints.




\section{\textsc{ItPla-Circ}: Circular Modules Placement and the Role of Distances}
\label{sec:circ}

If we consider the ROBOSKIN layout, we may observe that triangular modules are arranged in an hexagonal pattern.
This Section considers a set $\mathbb{S}_P$ of circular modules to be arranged in an hexagonal structure in order to cover at best the polygon $P$.
We focus on circular modules to discuss a few basic properties of \textsc{ItPla}.

It has been demonstrated, first by Gauss and then by Fejes T\'{o}th \cite{Toth1972}, that \textit{hexagonal packing} is the densest packing in the plane, as it forms a \emph{lattice} with a packing density $\eta = 0.9068$.        
Such a packing lattice replicates a number of configurations which can be found in Nature, e.g., bee's honeycombs and the graphite structure \cite{Williams1979}.
It does make sense to understand how such properties can be adapted to our purposes.

As anticipated in the previous Section, \textsc{ItPla} proceeds by iteratively arranging modules using a simulated pseudo-force field, which models attraction and repulsion forces among modules.
The algorithm tends to produce stable placements because attraction and repulsion forces tend to equate as by the \textit{energy loss} caused by simulated friction. 
In our case, friction is mimicked considering the level of overlap between modules.

While repulsion forces act to separate two neighbouring modules, attraction forces allow them to aggregate.
In the algorithm, a repulsion force occurs when modules overlap with each other.
The placement resulting from a distributed local effect of attracting and repulsing pseudo-forces is expected to resemble an hexagonal packing.
In order to obtain such placements, however, a faithful simulation of attraction and repulsion forces is not necessary.
Indeed, as pointed out by Eades \cite{Eades1984}, it is possible to adopt pseudo-forces, which can be used to compute velocity vectors associated with module's poses each time quantum $\Delta t$, whereas \emph{real} forces would induce accelerations.
In our case, this distinction is important because real forces lead to dynamic equilibria, whereas we are interested in static equilibria.

Given the geometric nature of the problem, let us define the \emph{overlap} and \emph{neighbour} relationships between modules.
In case of circular modules, the concept of pose reduces to their positions.
For the overlap, a working definition is straightforward.

\begin{definition}
Given two modules $m_i$, $m_j \in \mathbb{S}_P$, let us consider the open subset of $m_i$ as $\wp m_i = m_i \backslash \partial m_i$ and $m_j$ as $\wp m_j = m_j \backslash \partial m_j$.
An overlap exists between $m_i$ and $m_j$, referred to as $O(m_i, m_j)$ or $O(m_j, m_i)$, when $\wp m_i \cap \wp m_j \neq \emptyset$.
\end{definition}
It is noteworthy that overlaps can occur also between modules and a polygon's edges.

\begin{definition}
Given a module $m_i \in \mathbb{S}_P$ and a polygon's edge $e_j \in \mathbb{E}_P$, let us consider the open subset of $m_i$ as $\wp m_i$.
An overlap exists between $m_i$ and $e_j$, referred to as $O(m_i, e_j)$ or $O(e_j, m_i)$, when $\wp m_i \cap e_j \neq \emptyset$.
\end{definition}

We define $\mathbb{O}_{m_i}$ as the set of all $m_j$ for which an $O(m_i, m_j)$ relationship holds, plus the set of all $e_j$ for which an $O(m_i, e_j)$ relationship holds.
In order to quantify the level of overlap between a module $m_i$ and an edge $e_j \in \mathbb{E}_P$, it is possible to refer to $\mathcal{P}(\cdot)$ as the area of $m_i$ outside $P$, i.e., $\mathcal{A}(m_i \backslash P)$.
Ideally, the total overlap area $\mathcal{O}_{\mathbb{P}}$ associated with a placement $\mathbb{P}_P$ can be computed as:
\begin{equation}
\mathcal{O}_{\mathbb{P}} = \frac{1}{2} \sum_{i=1}^{\mathcal{B}^*_P} \mathcal{A}\left({O(m_i, m_j)}\right) + \sum_{i=1}^{\mathcal{B}^*_P} \mathcal{P}\left({O(m_i, e_j)}\right),
\end{equation}
which can be reduced to:
\begin{equation}
\mathcal{O}_{\mathbb{P}} = \sum_{i=1}^{\mathcal{B}^*_P} \mathcal{A}(m_i) - \mathcal{A}\left(\bigcup_{i=1}^{\mathcal{B}^*_P} m_i \cap P \right).
\end{equation}
The total overlap area is used to determine whether the associated placement $\mathbb{P}_P$ is acceptable.
This occurs if $\mathcal{O}_{\mathbb{P}}$ is lower than or equal to the threshold $\tau_o$:
\begin{equation}
\mathcal{O}_{\mathbb{P}} \leq \tau_o.
\end{equation}

\begin{figure}[t!]
	\centering
	\includegraphics[width=0.40\hsize]{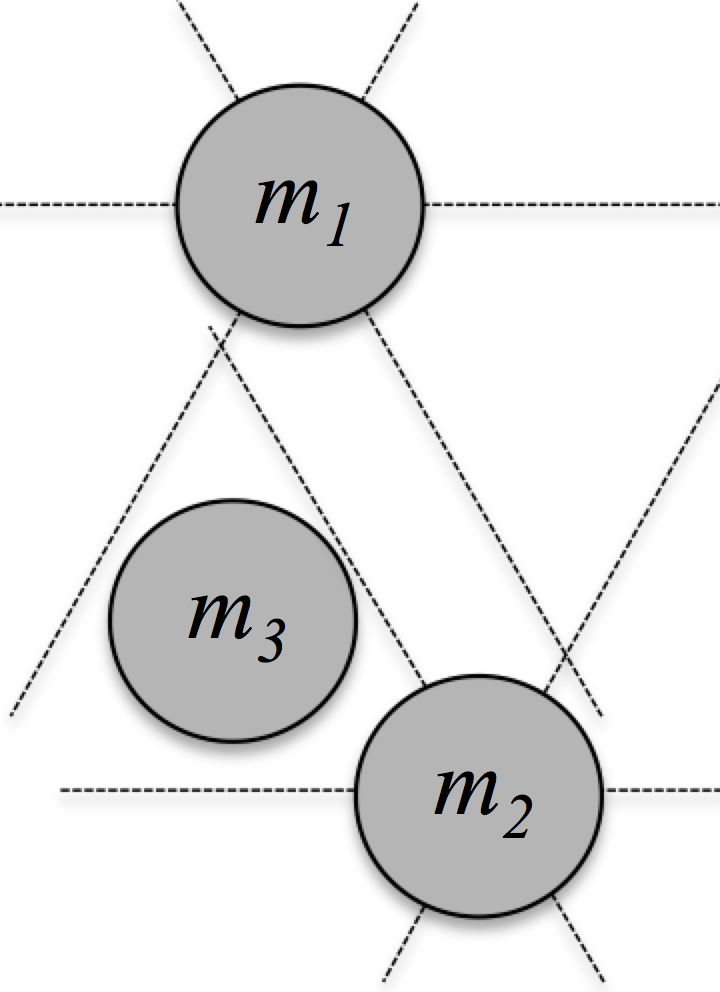}
	\caption{The neighbour relationship is not always reciprocal.}
	\label{fig:one-way}
\end{figure}
In our case, neighbouring relationships are tied to the peculiar geometrical nature of the problem.
A circular module can have at most $6$ neighbours, which we want to be arranged in an hexagonal structure to enforce the covered area.
To this aim, we ideally partition the planar space around a given module $m_i$ in $6$ equal semi-spaces $s_{i, k}$, with $k=1, \ldots, 6$, and we consider as the neighbour module in each semi-space $s_{i, k}$ the one \textit{closest} to $m_i$. 
Of course, this implies the notion of \emph{distance} between modules, which is introduced and discussed later.
It is noteworthy that, according to this statement, if $m_i$ is a neighbour of $m_j$, does not necessarily hold that $m_j$ is a neighbour of $m_i$, i.e., the relationship is not reciprocal. 
An example is shown in Figure \ref{fig:one-way}.
Here, $m_1$ is a neighbour of $m_2$, but the opposite does not hold.
This is due to the fact that $m_1$ is the only module belonging to one of the semi-spaces of $m_2$, whereas both $m_2$ and $m_3$ lie in the same semi-space of $m_1$, and $m_3$ is closer to $m_1$. 
As a consequence, because the interaction caused by pseudo-forces depends on neighbour relationships, they are not expected to be necessarily mutual.

\begin{definition}
Given a module $m_i \in \mathbb{S}_P$, let us consider the open subset of $m_i$ as $\wp m_i$ and its pose $p_i$.
For each module $m_j \in \mathbb{S}_P \backslash m_i$, let us consider its open subset $\wp m_j$ and its pose $p_j$.
For each semi-space $s_{i, k}$ around $m_i$, we classify each $m_j$ as belonging to strictly one semi-space $s_{i, k}$, i.e., $m_j \in s_{i, k}$, depending on its pose. 
Then, for each $s_{i, k}$, we define $m^*_j$ as a neighbour of $m_i$, and we write $N_k(m^*_j, m_i)$, if and only if:
\begin{equation}
m^*_j = \argmin_j d(p_j, p_i),
\end{equation}
where $d$ is an Euclidean distance function of module's poses, i.e., their centres in this case, and $m_j \in s_{i, k}$.
\end{definition}

\begin{figure}[t!]
	\centering
	\subfigure[]{
		\includegraphics[width=0.35\hsize]{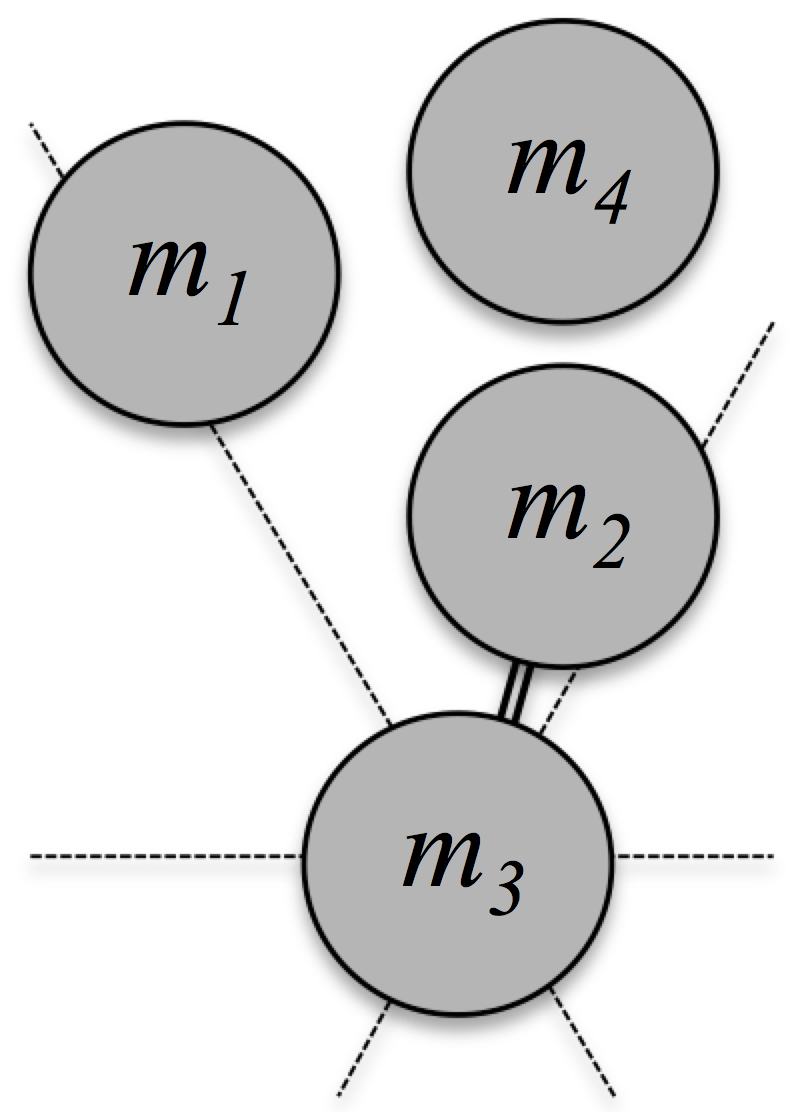}
		\label{fig:neighbour_a}
	}
	\hspace{0.05\hsize}
	\subfigure[]{
		\includegraphics[width=0.39\hsize]{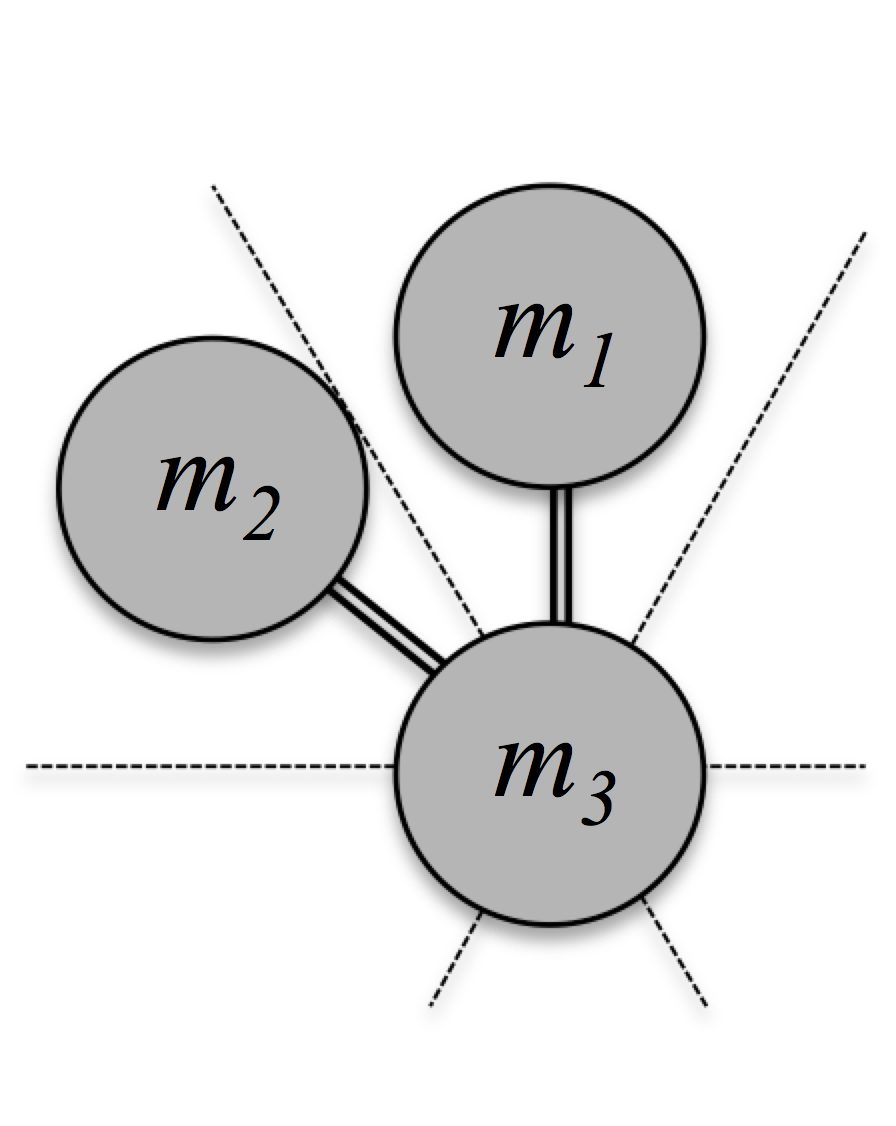}
		\label{fig:neighbour_b}
	}
	\caption{Two examples of neighbour relationships: (a) the line segment between two module's centres, if not interrupted, determines the neighbour relationships; (b) the closest module is defined as the neighbour.}
	\label{fig:neighbour}
\end{figure}
Figure \ref{fig:neighbour} shows two examples of neighbour relationships.
On the left, we notice that only $N(m_2, m_3)$ holds, whereas $N(m_1, m_3)$ and $N(m_4, m_3)$ do not hold since $m_1$ and $m_4$ are farther from $m_2$ than $m_3$.
On the right, both $N(m_1, m_3)$ and $N(m_2, m_3)$ hold, since $m_1$ and $m_2$ belong to two different semi-spaces.  

We consider overlap relationships as an indicator of sub-optimal placements.
In fact, in the ideal case of acceptable placements, no overlap relationships are present, because all modules keep the proper distance from their neighbours.
Henceforth, we assume that only neighbour modules can influence each other through pseudo-forces.
This allows \textsc{ItPla} to consider only \emph{local} pseudo-forces, which can be modelled using local information only, such as the current and the theoretically lowest distances between modules.  

\begin{figure}[t!]
	\centering
	\subfigure[]{
		\includegraphics[width=0.75\hsize]{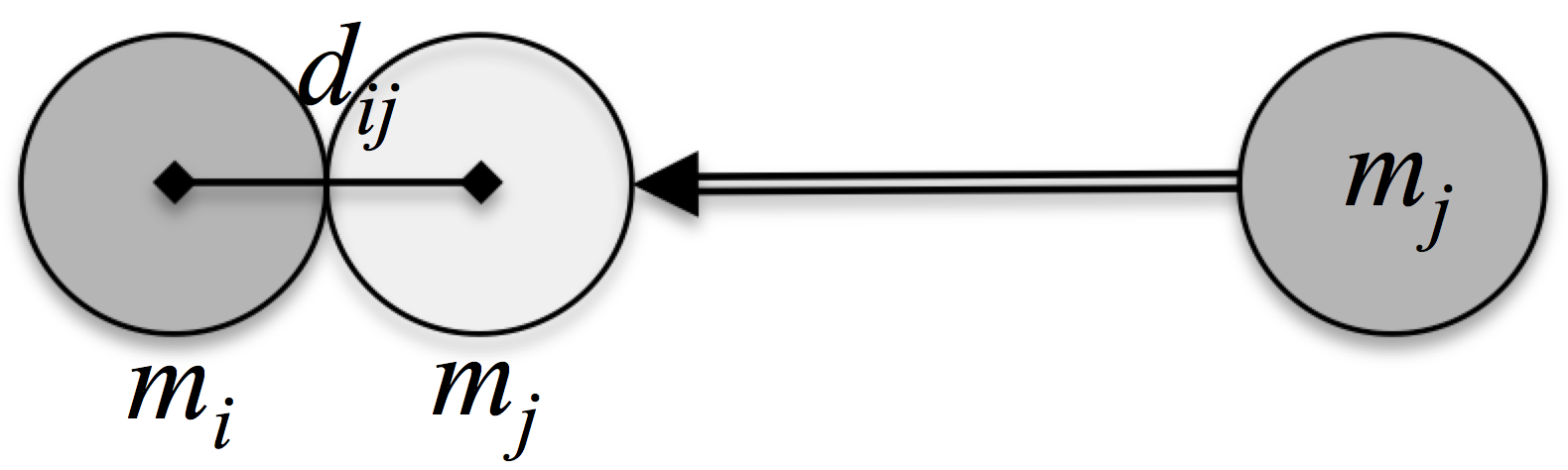}
	}
	\subfigure[]{
		\includegraphics[width=0.75\hsize]{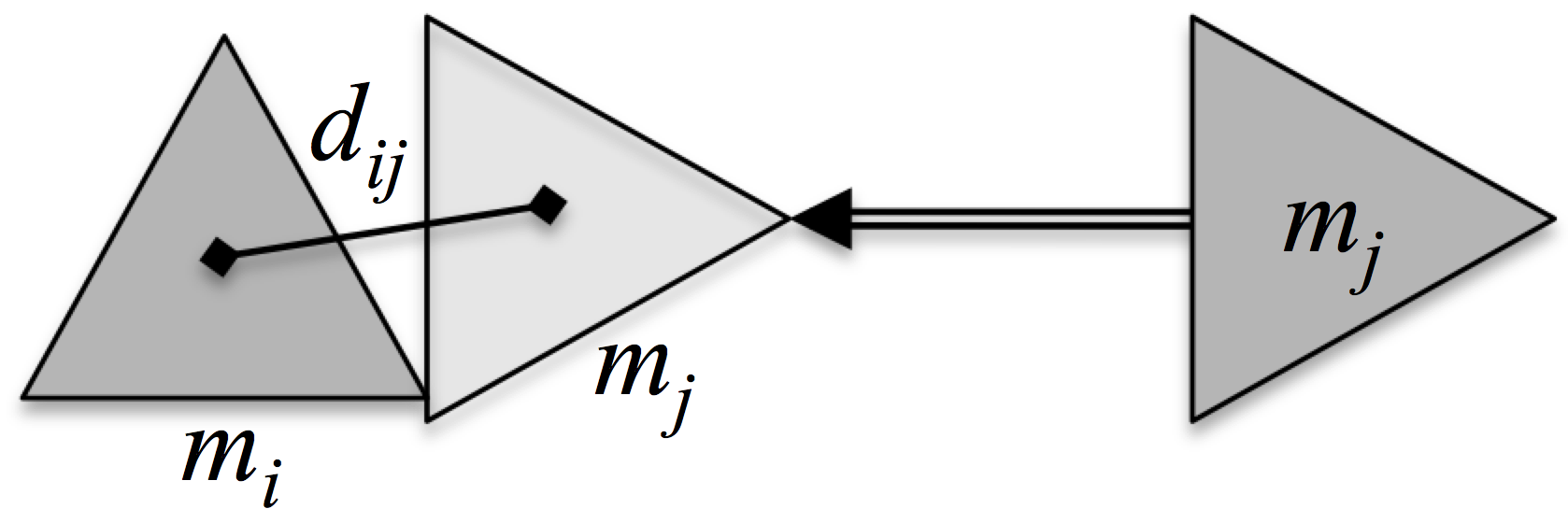}
	}
	\caption{Determining the lowest distance between two modules: (a) circular modules; (b) triangular modules.}
	\label{fig:distancemin}
\end{figure}
Figure \ref{fig:distancemin} shows how the lowest distance between any two modules $m_i$ and $m_j \in s_{i, k}$ is computed, respectively for circular and -- we anticipate -- for triangular modules. 
For circular modules, the lowest Euclidean distance is obviously independent from module's orientations and dependent on module's radii, whereas this is not the case for triangular modules.
In order to avoid any module's shape-specific differences when computing the distance, we adopt a \emph{normalised} distance, described as follows.

\begin{definition}
Given a module $m_i \in \mathbb{S}_P$ and a module $m_j \in s_{i, k}$, we consider the corresponding poses $p_i$ and $p_j$. 
We let $m_j$ translate (and not rotate) along the segment determined by the centres of $p_i$ and $p_j$.
The nearest pose $p^*_j$ such that $m_j$ can reach $m_i$ without overlapping determines the closest location to $m_i$. 
Then, the normalised distance $\overline{d}$ between $m_i$ and $m_j$ can be computed as follows:
\begin{equation}
\overline{d}(p_i, p_j) = \left(\frac{|p_i p_j|}{|p_i p^*_j|}^2 - 1\right)\cdot\frac{d(p_i, p^*_j)}{|p_i p^*_j|}.
\label{equ:f:Delta}
\end{equation}
\end{definition}
It is noteworthy that the normalised distance in \eqref{equ:f:Delta} is not mutual, i.e., $\overline{d}(p_i, p_j) \neq \overline{d}(p_j, p_i)$. 
If $m_i$ and $m_j$ are such close that their distance is lower than $d(p_i, p^*_j)$, then $\overline{d}$ is negative (i.e., there is overlap) and $m_i$ and $m_j$ are subject to a \textit{repulsion} force.
Otherwise, $\overline{d}$ is positive and they are subject to an \textit{attraction} force.
The actual amplitude of the pseudo-force varies according to the difference between the current and the minimum distance as reflected by the normalised distance defined before (i.e., the bigger the difference, the stronger the pseudo-force) and indirectly to the level of overlap between two modules.

A module $m_i$ is influenced by pseudo-forces originating from neighbours of each semi-space $s_{i, k}$.
Assuming that $m_k$ is the neighbour module in $s_{i, k}$, then we refer to $f^d_{k, i}$ as the pseudo-force originating from $m_k$ and acting on $m_i$ related to the normalised distance, as:
\begin{equation}
f^d_{k,i} = \chi_d\cdot\overline{d}(p_i, p_k),
\end{equation}
where $\chi_d$ is a positive real number, possibly used to model a number of features of the pseudo-force, e.g., attraction and repulsion profiles.
Therefore, we label a pseudo-force as \emph{attractive} when $f^d > 0$, and as \emph{repulsive} otherwise.
The overall pseudo-force $\Phi^d_i$ on $m_i$ can be computed as the sum of all pseudo-forces in each semi-space:
\begin{equation}
\Phi^d_i = \sum_{k} f^d_{k,i}.
\label{eq:overallF1}
\end{equation}
The connection between modules must be kept until the repulsive pseudo-force exceeds a parametric amplitude.
We adopt a weighting approach to obtain stable placements in the iterative process.
Specifically, \eqref{eq:overallF1} can be rewritten as follows:
\begin{equation}
\Phi^d_i = \frac{\sum_{k} w^d_{k,i} \cdot f^d_{k,i}}{\sum_{k} w^d_{k, i}},
\label{eq:overallF2}
\end{equation}
where $w^d_{k, i}$ refers to the weight of $f^d_{k, i}$ on $m_i$.

In order to determine such weights, different functions are equally legitimate. 
Here, we adopt a simple function depending on the ratio between distances involving modules $m_i$ and $m_k$ (i.e., the neighbour in $s_{i, k}$), and in particular with respect to their poses:
\begin{equation}
\label{equ:basic_weight}
w^d_{k,i} = \frac{| p_i p^*_k|}{| p_i p_k|}^n,
\end{equation}
where $n$ is an positive integer number.
The weighting mechanism in \eqref{equ:basic_weight} considers the ratio between the theoretically lowest and the current distance of modules $m_i$ and $m_k$, earlier referred to as $| p_i p^*_k|$.
When the current and the lowest distances are the same (i.e., modules are \emph{connected}), then $w^d_{k,i} =1$.
If the actual distance is smaller that the theoretically lowest one (i.e., an overlap occurs), the weighting function increases so that a repulsive pseudo-force $f^d_{k, i}$ caused by the overlapping module is weighted more in \eqref{eq:overallF2}. 
Otherwise, if the distance is greater than the theoretically lowest one, which implies that the corresponding module does not violate any constraints, then the corresponding forces are weighted less.
As a result, \textsc{ItPla} first tends to eliminate overlaps, then maintains the established connections, and finally deals with isolated modules.

\section{\textsc{ItPla-Tri}: Triangular Modules Placement and the Effects of Moments and Translation Offsets}
\label{sec:tri}

In the previous Section, we discussed the basic principles of \textsc{ItPla} in the case of circular modules.
In order to consider a set $\mathbb{S}_P$ of triangular modules covering a polygon $P$, a few differences with respect to the previous case must be discussed.
\begin{itemize}
\item The notions of overlap between modules, total overlap and neighbouring do not change, subject to the different concept of Euclidean distance shown in Figure \ref{fig:distancemin}. However, the normalised distance in (\ref{equ:f:Delta}) is used also in the case of triangular modules.
\item Considering a triangular module $m_i$, the number of semi-spaces $s_{i, k}$ reduces to $3$, because each module can be connected at most to three other modules.
\item The pose $p_i$ must explicitly include the module's orientation $\theta_i$, which is expressed with respect to an external (i.e., absolute) reference frame.
\item Whilst circular modules can be connected together without defining a specific connection location on their border, ROBOSKIN triangular modules can be only connected to each other through the mid points of their sides. 
\end{itemize}
As a consequence, beside the one described in (\ref{eq:overallF2}), two other pseudo-forces are designedAs discussed in \cite{Anghinolfietal2013}, this proves to be equivalent to cover the original 3D surface in so far as the distortions induced by the flattening procedure are minimised. 
 to play a major role in \textsc{ItPla-Tri}, the first related to a triangular module's \emph{moment}, the second to the \emph{translation offset alignment} between adjacent sides of neighbour modules.
On the one hand, the constraints $C_1$ and $C_2$ introduced above are met when there is no overlap nor offset misplacement between neighbour triangular modules, i.e., when close sides are \textit{aligned} with each other.
This implies that it is necessary to induce moments to rotate them of a certain amount.
On the other hand, neighbouring triangular modules must shift along their closest sides to match the corresponding mid points.
This induces an additional interaction constraint, which is related to how real-world modules are \textit{connected} to each other.

%
%

We define first how to characterise the angular difference between two neighbour triangular modules.

\begin{figure}[t!]
\centering
\includegraphics[width=0.55\hsize]{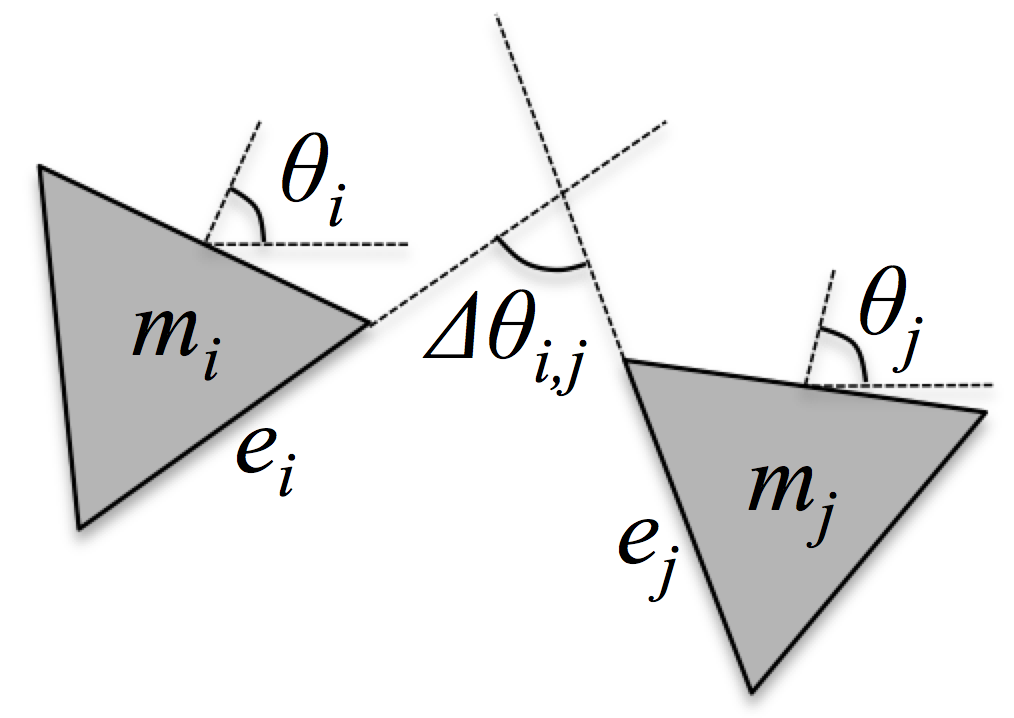}
\caption{The angle $\Delta\theta_{i,j}$ between $m_i$ and $m_j$ as computed using (\ref{eq:angle_diff}).}
\label{fig:angle_diff}
\end{figure}
\begin{definition}
Given two triangular modules $m_i$ and $m_j$ such that $N(m_i, m_j)$ and $N(m_j, m_i)$, let us define $e_i$ as the side of $m_i$ closest to $m_j$, and $e_j$ as the side of $m_j$ closest to $m_i$, i.e., $e_i$ and $e_j$ face each other. Then, the angular difference $\Delta\theta_{i,j}$ between $m_i$ and $m_j$ can be computed as:
\begin{equation}
\Delta\theta_{i,j} = \theta_i - \theta_j + \frac{\Pi}{3} \cdot (\overrightarrow{e_i} -\overrightarrow{e_j}) - \frac{\Pi}{2},
\label{eq:angle_diff}
\end{equation}
where $\overrightarrow{e_i}$ and $\overrightarrow{e_j}$ are vectors associated with modules' sides $e_i$ and $e_j$.
\end{definition}
An example is shown in Figure \ref{fig:angle_diff}.
In order to align two triangular modules, we can use (\ref{eq:angle_diff}) to distribute the variation in modules orientation between $m_i$ and $m_j$ (i.e., their moments $\tau_{i}$ and $\tau_{j}$), as follows:
\begin{equation}
\tau_{i} = \tau_{j} = \frac{\Delta\theta_{i,j}}{2},
\label{eq:equal_angle_diff}
\end{equation}
where the scaling factor $\frac{1}{2}$ is applied so that both $m_i$ and $m_j$ rotate of the same amount to align their closest sides $e_i$ and $e_j$.
Of course, other scaling factors can be chosen.
Here, we assume an equal distribution of angular variations.   

As in the case of circular modules, a module $m_i$ is influenced by the pseudo-forces exerted by its neighbours.
Given a module $m_k$ in a semi-space $s_{i, k}$, we refer to $f^{\tau}_{k, i}$ as the pseudo-force acting on $m_i$ related to the rotation induced by $m_k$, as:
\begin{equation}
f^{\tau}_{k, i} = \chi_{\tau} \cdot \frac{\Delta\theta_{i,k}}{2},
\end{equation} 
where $\chi_{\tau}$ is a positive real number.
In this case, we consider the pseudo-force as attractive when $\theta_i$ and $\theta_k$ converge to each other, and as repulsive otherwise.
Given the contribution of the three semi-spaces around $m_i$, the overall pseudo-force $\Phi^{\tau}_i$ is given by:
\begin{equation}
\Phi^{\tau}_i = \sum_{k} f^{\tau}_{k,i}.
\label{eq:overallFTau1}
\end{equation}
Analogously to (\ref{eq:overallF2}), stable placements can be enforced by introducing weights, such as:
\begin{equation}
\Phi^{\tau}_i = \frac{\sum_{k} w^{\tau}_{k, i} \cdot f^{\tau}_{k,i}}{\sum_{k} w^{\tau}_{k, i}},
\label{eq:overallFTau2}
\end{equation}
where $w^{\tau}_{k, i}$ refers to the weight of $f^{\tau}_{k, i}$ on $m_i$.
Weights can depend on the angular difference between the two modules, for example:
\begin{equation}
w^{\tau}_{k, i} = (\Delta\theta_{i,k})^2,
\label{eq:weightFTau}
\end{equation}
where, when $\theta_i = \theta_k$, then $f^{\tau}_{k,i} = 0$. 

\begin{figure}[t!]
\centering
\includegraphics[width=0.55\hsize]{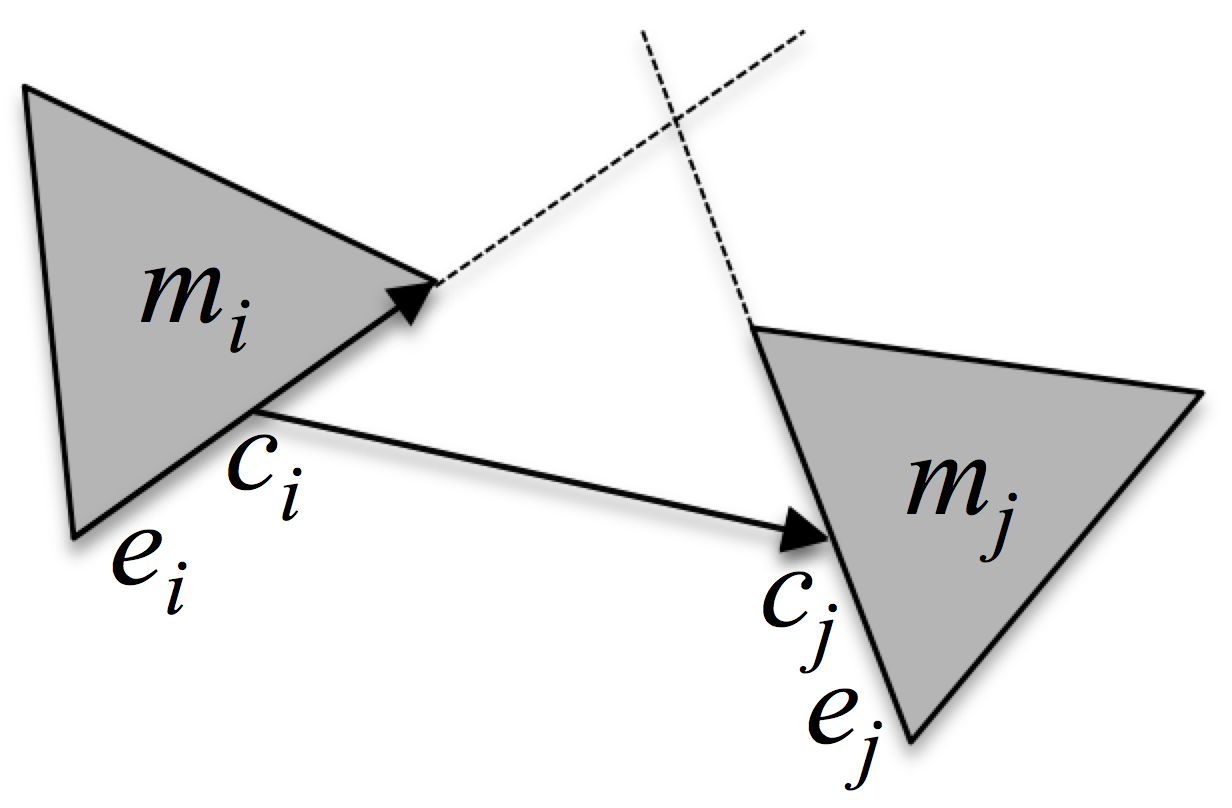}
\caption{The parameters used in offset pseudo-force calculation}
\label{fig:translation_offset}
\end{figure}

Now we define the translation offset between two neighbour triangular modules.
\begin{definition}
Given two triangular modules $m_i$ and $m_j$ such that $N(m_i, m_j)$ and $N(m_j, m_i)$, let us define $e_i$ as the side of $m_i$ closest to $m_j$, and $e_j$ as the side of $m_j$ closest to $m_i$, i.e., $e_i$ and $e_j$ face each other, $c_i$ as the mid point of $e_i$ and $c_j$ as the mid point of $e_j$. Then, the translation offset $\Delta T_{i,j}$ between $m_i$ and $m_j$ is computed as:
\begin{equation}
\Delta T_{i,j} = \frac{\overrightarrow{c_i c_j}\cdot\overrightarrow{e_i}}{2} \cdot \frac{\overrightarrow{e_i}}{|e_i|},
\label{eq:translation_offset}
\end{equation}
where $\overrightarrow{c_i c_j}$ is the vector connecting the two mid points, and $\overrightarrow{e_i}$ is a vector associated with $e_i$.
\end{definition}
The situation is depicted in Figure \ref{fig:translation_offset}.
Analogously to what we do in (\ref{eq:equal_angle_diff}), we can distribute equally the offset reduction to $m_i$ and $m_j$, by imposing:
\begin{equation}
t_{i} = t_{j} = \frac{\Delta T_{i,j}}{2}.
\label{eq:equal_translation_offset}
\end{equation}

Given a module $m_k$ in a semi-space $s_{i, k}$, we refer to $f^{T}_{k, i}$ as the pseudo-force acting on $m_i$ related to the translation offset induced by $m_k$, as:
\begin{equation}
f^{T}_{k, i} = \chi_{T} \cdot \frac{\Delta T_{i,k}}{2},
\label{eq:PseudoForceT}
\end{equation} 
where $\chi_{T}$ is a positive real number.
The pseudo-force is always attractive whether $\Delta T$ reduces or increases.
Given the contribution of the three semi-spaces around $m_i$, the overall pseudo-force $\Phi^{T}_i$ can be computed as:
\begin{equation}
\Phi^{T}_i = \sum_{k} f^{T}_{k,i}.
\label{eq:overallFT1}
\end{equation}
Also in this case we introduce weights to enforce stable placements, as follows:
\begin{equation}
\Phi^{T}_i = \frac{\sum_{k} w^T_{k, i} \cdot f^{T}_{k,i}}{\sum_{k} w^T_{k, i}},
\label{eq:overallFT2}
\end{equation}
where: 
\begin{equation}
w^T_{k, i} = (\Delta T_{i,k})^2.
\label{eq:weightFT}
\end{equation}

The total misplacement $\mathcal{M}_{\mathbb{P}}$ associated with a placement $\mathbb{P}_P$ can be computed as:
\begin{equation}
\mathcal{M}_{\mathbb{P}} = \frac{1}{2} \sum_{i=1}^{\mathcal{B}^*_P} \sum_{j=1, j \neq i}^{\mathcal{B}^*_P} \mathcal{L}(\Delta T_{i,j}),
\end{equation}
where $\mathcal{L}(\cdot)$ denotes the length of a line segment. 
The total misplacement is used to determine whether the associated placement $\mathbb{P}_P$ is acceptable.
This occurs if it is lower than or equal to the threshold $\tau_m$:
\begin{equation}
\mathcal{M}_{\mathbb{P}} \leq \tau_m.
\end{equation}

Once these two pseudo-forces are introduced, we can define the overall pseudo-force acting on a triangular module $m_i$ simply by summing up the contributions of \emph{distances} (\ref{eq:overallF2}), \emph{moments} (\ref{eq:overallFTau2}) and \emph{offsets} (\ref{eq:overallFT2}), as follows:
\begin{equation}
\Phi_i = \Phi^d + \Phi^{\tau}_i + \Phi^{T}_i. 
\label{eq:SumPseudoForces}
\end{equation}

\begin{figure}[t!]
\centering
\includegraphics[width=0.90\hsize]{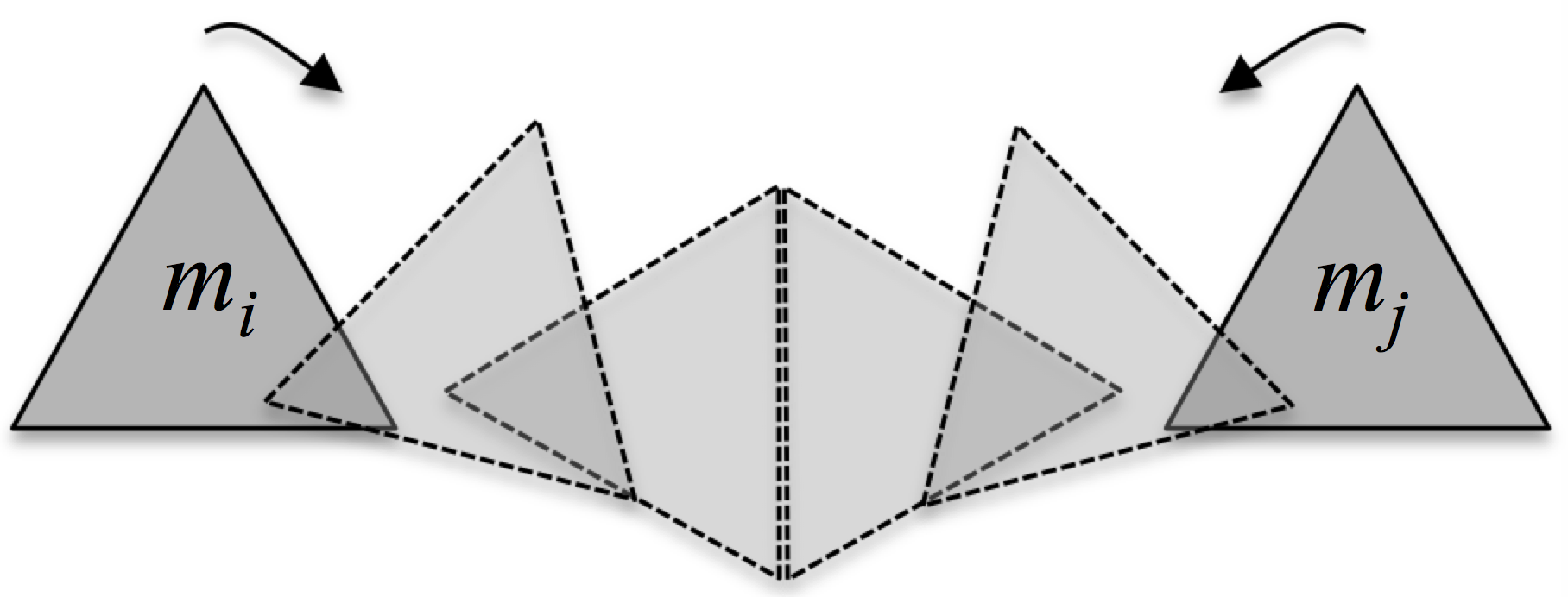}
\caption{An example of the combined effect of the three pseudo-forces $\Phi^d, $ $\Phi^{\tau}$ and $\Phi^{T}$.}
\label{fig:force}
\end{figure}
Figure \ref{fig:force} shows an example of the effects of $\Phi_i$ on two triangular modules. 
It is possible to note that distances, moments and translation offset alignments contribute to the mutual position of two modules $m_i$ and $m_j$.

\section{\textsc{ItPla}: Iterative Placement}
\label{sec:ItPla}

Having devised the main roles of pseudo-forces, we can now introduce the whole \textsc{ItPla} algorithm.
It is necessary to determine how the algorithm deals with the effects of the polygon's border on modules, how stable placements are determined, and how modules in overlap (which cannot be accommodated) are removed from the placement.

\subsection{The Influence of the Polygon's Border on Modules}
\label{fig:InfluenceBorderPolygon}
In this Section we describe how a polygon's border influences the poses of the modules inside it.
As we anticipated, the border constitutes a hard constraint, since no overlap is allowed with its edges.
However, if the algorithm is used to cover a robot's body part using the ROBOSKIN technology, a small portion of triangular module corners can be cut, and in practice we can allow module's corners to overlap with polygon's edges \cite{Anghinolfietal2013}. 
This is not true in general, though.

Let us define first the neighbour relationship between modules and edges.
\begin{definition}
Given a module $m_i \in \mathbb{S}_P$ and an edge $e_j \in \mathbb{E}_P$, let us consider the open subset of $m_i$ as $\wp m_i$, its pose $p_i$, and let us define $c_j$ as the mid point of $e_j$.
For each semi-space $s_{i, k}$ around $m_i$, we classify $e_j$ as belonging to strictly one semi-space $s_{i, k}$, such that $e_j \in s_{i, k}$, depending on the position of $c_j$. 
Then, for each $s_{i, k}$, we define $e^*_j$ as a neighbour of $m_i$, and we write $N(e^*_j, m_i)$, such as:
\begin{equation}
e^*_j = \argmin_j d(c_j, p_i),
\end{equation}
where $d$ is the Euclidean distance and $e_j \in s_{i, k}$.
\end{definition}

Henceforth, we refer to $\mathbb{N}_{m_i}$ as the set of all $m_j$ for which a $N(m_i, m_j)$ relationship holds, plus the set of all $e_j$ for which a $N(e_j, m_i)$ relationship holds.

Given a polygon $P$, each edge $e_j \in \mathbb{E}_P$ contributes to the pose $p_i$ of a neighbour module $m_i$, both with an induced attraction or repulsion, as well as with a moment pseudo-force.
In order to define how an edge $e_j$ attracts or repels a module $m_i$, we must define the notion of distance between them.
We recur to the notion of generalised distance previously introduced in Definition $4$.
\begin{definition}
Given a module $m_i \in \mathbb{S}_P$ and a polygon's edge $e_j \in \mathbb{E}_P$, we consider the pose $p_i$ and the point $c_j$ of $e_j$ where $p_i$ can be projected along the normal direction. 
We let $m_i$ translate (and not rotate) along the segment determined by $p_i$ and $c_j$.
The nearest location $p^*_i$ such that $m_i$ can reach $e_j$ without overlapping determines the closest location to $e_j$. 
We refer to $\overline{p^*_i c_j}$ as the Euclidean distance between $p^*_i$ and $c_j$.
Then, the normalised distance $\overline{d}$ between $m_i$ and $e_j$ can be computed as follows:
\begin{equation}
\label{eq:DistModuleEdge}
\overline{d}(p_i, e_j) = \left(\frac{d(p_i, c_j)}{| p^*_i c_j|}^2 - 1\right)\cdot\frac{d(p^*_i, c_j)}{|p^*_i c_j|}.
\end{equation}
\end{definition}
This notion of distance can be visualised as a particular case of what happens in Figure \ref{fig:distancemin}, where the edge $e_j$ acts as if it were a triangular module's edge.

The attractive or repulsive pseudo-force originating from $e_j$ and acting on $m_i$ depends on the distance $\overline{d}$, as follows:
\begin{equation}
f^{P,d}_{j,i} = \chi_{P,d}\cdot\overline{d}(p_i, e_j),
\end{equation}
where $\chi_{P,d}$ is a positive real number.
It is noteworthy that, if the actual distance between $m_i$ and $e_j$ is lower than $\overline{p^*_i c_j}$, the distance is negative (i.e., there is overlap), then $m_i$ and $e_j$ are subject to a repulsion force; otherwise, to an attractive force. 
The overall weighted pseudo-force $\Phi^{P,d}_i$ on $m_i$ due to polygon's edges can be computed as:
\begin{equation}
\Phi^{P,d}_i = \frac{\sum_{j} w^{P,d}_{j,i} \cdot f^{P,d}_{j,i}}{\sum_{j} w^{P,d}_{j, i}},
\label{eq:WeightedOverallDistanceForceEdge}
\end{equation}
where $w^{P,d}_{j, i}$ refers to the weight of $f^{P,d}_{j, i}$ on $m_i$, i.e., the contribution of $e_j$ related to its distance with respect to the module $m_i$.


As far as the moment pseudo-force is concerned, we can reuse the notion of angular difference of Definition 5, treating again the edge $e_j$ as if it were the edge of a triangular module.
The rotation induced by $e_j$ on $m_i$ is given by the following pseudo-force:
\begin{equation}
f^{P,\tau}_{j,i} = \chi_{P,\tau}\cdot \Delta \theta_{i,j},
\end{equation}
where, differently from (\ref{eq:equal_angle_diff}), all the rotation is induced on $m_i$.
The overall weighted pseudo-force $\Phi^{P,\tau}_i$ on $m_i$ due to polygon's edges is computed as:
\begin{equation}
\Phi^{P,\tau}_i = \frac{\sum_{j} w^{P, \tau}_{j,i} \cdot f^{P,\tau}_{j,i}}{\sum_{j} w^{P, \tau}_{j, i}}.
\label{eq:WeightedOverallMomentForceEdge}
\end{equation}
Analogously to what we do in (\ref{eq:WeightedOverallDistanceForceEdge}), we introduce weights to balance the contributions of different edges.
As a result, (\ref{eq:SumPseudoForces}) can be updated as:
\begin{equation}
\Phi_i = \Phi^d + \Phi^{\tau}_i + \Phi^{T}_i + \Phi^{P,d}_i + \Phi^{P,\tau}_i,
\label{eq:SumPseudoForcesWithEdges}
\end{equation}
where we can identify the contributions of modules and edges, respectively, to a module's translation and rotation.


\begin{algorithm}[t]
\caption{GenerateNextPlacement()}
\label{alg:step}
\begin{algorithmic}[5]
	\REQUIRE a polygon $P$, a set of modules $\mathbb{S}_P$, the current placement $\mathbb{P}^c_P$, the time quantum $\Delta t$
	\ENSURE a new placement $\mathbb{P}^n_P$
	\FORALL {$m_i \in \mathbb{S}_P$}
		\FORALL {$m_k \in s_{i,k}$}
			\STATE Compute $\overline{d}(p_i,p_k)$, $f^d_{k,i}$ and $w^d_{k,i}$
		\ENDFOR
		\STATE Compute $\Phi^d_i$
		\FORALL {$m_k \in s_{i,k}$}
			\STATE Compute $\Delta \theta_{i,k}$, $f^{\tau}_{k,i}$ and $w^{\tau}_{k,i}$
		\ENDFOR
		\STATE Compute $\Phi^{\tau}_i$
		\FORALL {$m_k \in s_{i,k}$}
			\STATE Compute $\Delta T_{i,k}$, $f^{T}_{k,i}$ and $w^{T}_{k,i}$
		\ENDFOR
		\STATE Compute $\Phi^{\tau}_i$
		\FORALL {$e_j \in \mathbb{E}_P$}
			\STATE Compute $\overline{d}(p_i,e_j)$, $f^{P,d}_{j,i}$ and $w^{P,d}_{j,i}$
		\ENDFOR
		\STATE Compute $\Phi^{P,d}_i$
		\FORALL {$e_j \in \mathbb{E}_P$}
			\STATE Compute  $\Delta \theta_{i,j}$, $f^{P,\tau}_{j,i}$ and $w^{P,\tau}_{j,i}$
		\ENDFOR
		\STATE Compute $\Phi^{P,\tau}_i$
		\STATE $\Phi^{tra}_i \leftarrow \Phi^d_i + \Phi^{\tau}_i + \Phi^{P,d}_i$
		\STATE $\Phi^{rot}_i \leftarrow \Phi^{\tau}_i + \Phi^{P,\tau}_i$			
    	\ENDFOR
	\FORALL {$m_i \in \mathbb{S}_P$}
		\STATE $\mathbb{P}^n_P$ $\leftarrow$ TranslateModule($P$, $m_i$, $\Phi^{tra}_i \Delta t$)
		\STATE $\mathbb{P}^n_P$ $\leftarrow$ RotateModule($P$, $m_i$, $\Phi^{rot}_i \Delta t$)
	\ENDFOR
\end{algorithmic}
\end{algorithm}
\subsection{Generation of New Placements}
\label{sec:PlacementAlgorithm}
Each placement step requires computing all the pseudo-forces, and then to use \eqref{eq:SumPseudoForcesWithEdges} to translate and rotate each module (Algorithm \ref{alg:step}).
The step assumes the availability of the polygon $P$ associated with the robot body part to cover (in particular, its edges in $\mathbb{E}_P$), the set of modules $\mathbb{S}_P$ constituting the current placement $\mathbb{P}^c_P$, and the time quantum $\Delta t$.

For each module $m_i$, GenerateNextPlacement() proceeds by computing all the pseudo-forces, respectively $\Phi^d_i$ using (\ref{eq:overallF2}) in line $5$, $\Phi^{\tau}_i$ using (\ref{eq:overallFTau2}) in line $9$, $\Phi^{\tau}_i$ using (\ref{eq:overallFT2}) in line $13$, $\Phi^{P,d}_i$ using (\ref{eq:WeightedOverallDistanceForceEdge}) in line $17$ and $\Phi^{P,\tau}_i$ using (\ref{eq:WeightedOverallMomentForceEdge}) in line $21$.
Then, a resultant pseudo-force for the translation component $\Phi^{tra}_i$ is computed in line $22$, and a for the rotation component $\Phi^{rot}_i$ in line $23$.
Finally, two functions, namely TranslateModule() and RotateModule() are used to generate the next placement $\mathbb{P}^n_P$ (lines 26 and 27).
In these two functions, pseudo-forces (i.e, velocity vectors) are transformed in displacement by time multiplication for the time quantum $\Delta t$. 

\begin{algorithm}[t]
\caption{GenerateStablePlacements()}
\label{alg:SetSolutions}
\begin{algorithmic}[5]
	\REQUIRE a polygon $P$, a set of modules $\mathbb{S}_P$, the number of modules $\mathcal{B}^c_P$, the time quantum $\Delta t$, the desired number of stable solutions $\tau_{\mathcal{E}}$
	\ENSURE the set of acceptable candidate placements $\mathcal{E}_P$
	\STATE $\mathcal{E}_P \leftarrow \{\emptyset\}$
	\STATE $i \leftarrow 0$
	\STATE $\mathbb{P}^c_P \leftarrow$ GenerateInitialPlacement($P$, $\mathbb{S}_P$, $\mathcal{B}^c_P$)
	\WHILE {$i < \tau_{\mathcal{E}}$}
		\STATE $\mathbb{P}^n_P$ $\leftarrow$ GenerateNextPlacement($P$, $\mathbb{S}_P$, $\mathbb{P}^c_P$, $\Delta t$)
		\STATE Compute $\mathcal{O}^c_{\mathbb{P}}$, $\mathcal{O}^n_{\mathbb{P}}$, $\mathcal{M}^c_{\mathbb{P}}$ and $\mathcal{M}^n_{\mathbb{P}}$
		\STATE $\Delta E \leftarrow \alpha_{\mathcal{O}} ( \mathcal{O}^n_{\mathbb{P}} - \mathcal{O}^c_{\mathbb{P}} ) + \alpha_{\mathcal{M}} ( \mathcal{M}^n_{\mathbb{P}} - \mathcal{M}^c_{\mathbb{P}}$ )
		\IF {$e^{-\frac{\Delta E}{\mathcal{F}(\mathcal{O}^c_{\mathbb{P}}, \mathcal{M}^c_{\mathbb{P}})}} < $ random(0, 1)}
			\STATE $\mathcal{E}_P \leftarrow \mathcal{E}_P \cup \mathbb{P}^n_P$
			\STATE $i \leftarrow i + 1$
		\ENDIF
		\STATE $\mathbb{P}^c_P \leftarrow \mathbb{P}^n_P$
	\ENDWHILE
\end{algorithmic}
\end{algorithm}
\subsection{Generation of Stable Placements}
\label{sec:GenerationOfSolutions}
When a new polygon $P$ is given, \textsc{ItPla} first computes with (\ref{eq:UpperBound}) the upper bound $\mathcal{B}_P$ to the number of modules to use.
\textsc{ItPla} tries to find a stable placement $\mathbb{P}_P$, whose acceptability must be determined.
If the total overlap area $\mathcal{O}_{\mathbb{P}}$ exceeds $\tau_o$ or the total misplacement $\mathcal{M}_{\mathbb{P}}$ exceeds $\tau_m$, a module to remove must be chosen (see next Section), and the algorithm keeps iterating on possible solutions.
Determining whether $\mathbb{P}_{P}$ is acceptable is a crucial step in the iteration process: on the one hand, removing a module implicitly prunes the solutions space, therefore making possibly good solutions unreachable; on the other hand, not removing a module can lead \textsc{ItPla} to bias on a specific portion of the solutions space, which may not contain the best solutions. 

We employ an adaptive simulated annealing approach to generate the set of acceptable candidate solutions, from which to select the best one. 
In our case, as described in Algorithm \ref{alg:step}, the iterative generation of the next solution (i.e., a placement $\mathbb{E}^n_P$) depends on the polygon $P$, the set of modules $\mathbb{S}_P$ and the current solution $\mathbb{E}^c_P$ through the combined effect of pseudo-forces.
Differently from the standard simulated annealing procedure, where the acceptability of a solution is subject to a certain randomness degree, in our case we can tie it also to the total overlap area $\mathcal{O}_{\mathbb{P}}$ and the total misplacement $\mathcal{M}_{\mathbb{P}}$, which are related to the specific placement. 
 
GenerateStablePlacements() (Algorithm \ref{alg:SetSolutions}) assumes a polygon $P$, the number of modules $\mathcal{B}^c_P$ used to generate the current coverage, and the time quantum $\Delta t$, to forward to GenerateNextPlacement() (Algorithm \ref{alg:step}), as well as the number of desired stable solutions $\tau_{\mathcal{E}}$.
At the beginning, the set of solutions $\mathcal{E}_P$ and the solution index $i$ are initialised (lines 1 and 2).
The function GenerateInitialPlacement() is called once to generate a random initial placement of modules, i.e., the \emph{current} placement $\mathbb{P}^c_P$ and the associated set of modules $\mathbb{S}_P$ (line 3).
Then, at least $\tau_{\mathcal{E}}$ iterations of lines $5$-$12$ are executed.
For each iteration, the GenerareNextPlacement() function is called, which generates a \emph{new} placement $\mathbb{P}^n_P$ (line 5).
The two total overlaps (respectively, for the new and current placements) $\mathcal{O}^n_{\mathbb{P}}$ and $\mathcal{O}^c_{\mathbb{P}}$, as well as the two total misplacements $\mathcal{M}^n_{\mathbb{P}}$ and $\mathcal{M}^c_{\mathbb{P}}$ are computed (line 6).
Their weighted difference $\Delta E$ is computed (line 7).
Parameters $\alpha_{\mathcal{O}}$ and $\alpha_{\mathcal{M}}$ can be chosen to weight differently area overlap and misplacements. 
When the total overlap and the total misplacement tend to stabilise, then $\Delta E$ decreases, leading to a \emph{stable} solution with a given probability.  
If this happens, the placement $\mathbb{P}^n_P$ is added to the set of stable solutions $\mathcal{E}_P$ and $i$ is increased (lines 9 and 10).
Otherwise, the new solution becomes the current one and a new placement for the index $i$ is determined (line 12).
It is noteworthy that the annealing effect is tied to both total area and total misplacement using a function $\mathcal{F}$: the lower their values, the greater the annealing effect (line 8).

\begin{algorithm}[t]
\caption{\textsc{ItPla}()}
\label{alg:Acceptable}
\begin{algorithmic}[5]
	\REQUIRE a polygon $P$, the time quantum $\Delta t$, the desired number of stable solutions $\tau_{\mathcal{E}}$
	\ENSURE an acceptable placement $\mathbb{P}^*_P$
	\STATE $\mathbb{P}^*_P \leftarrow$ null 
	\STATE Compute $\mathbb{B}_P$
	\STATE $\mathbb{B}^c_P \leftarrow \mathbb{B}^*_P$
	\STATE $\mathbb{S}^c_P \leftarrow$ null
	\WHILE {$\mathbb{P}^*_P$ is null}	
		\STATE $\mathcal{E}_P \leftarrow$ GenerateStablePlacements($P$, $\mathbb{S}^c_P$, $\mathbb{B}^c_P$, $\Delta t$, $\tau_{\mathcal{E}}$)
		\STATE $\mathbb{P}^+_P \leftarrow$ SelectFrom($\mathcal{E}_P$)
		\IF{$\mathcal{O}_{\mathbb{P}} \leq \tau_o$ and $\mathcal{M}_{\mathbb{P}} \leq \tau_m$ }
			\STATE $\mathbb{P}^*_P \leftarrow \mathbb{P}^+_P$ 
		\ELSE
			\STATE $\mathbb{B}^c_P \leftarrow \mathbb{B}^c_P - 1$
			\STATE $\mathbb{S}^c_P \leftarrow$ RemoveModuleFrom($\mathbb{S}^+_P$) 
		\ENDIF 
	\ENDWHILE
\end{algorithmic}
\end{algorithm}
\subsection{Generation of Acceptable Placements}
\label{sec:GenerationAccPlacements}
With the generation of acceptable placements, we fully implement the $\textsc{ItPla}$ procedure, as outlined in Figure \ref{fig:procedure}.
$\textsc{ItPla}$ (Algorithm \ref{alg:Acceptable}) assumes as input a polygon $P$, the time quantum $\Delta t$ for the iterative procedure and the number of stable solutions to choose from.
It produces an acceptable placement $\mathbb{P}^*_P$ that can be evaluated by a human designer.

First of all, the upper bound to the number of modules is computed using (\ref{eq:UpperBound}) in line 2.
A number of steps are executed until an acceptable placement $\mathbb{P}^*_P$ is found (lines 5 to 14).
The GenerateStablePlacement() function is executed to generate a set $\mathcal{E}_P$ of stable placements (line 6).
When the set of stable solutions $\mathcal{E}_P$ is determined, the \emph{best candidate} for acceptability $\mathbb{P}^+_P$ can be selected (line 7).
Different strategies may be involved.
For instance, the solution minimising a weighted sum of total overlap $\mathcal{O}_{\mathbb{P}}$ and total misplacement $\mathcal{M}_{\mathbb{P}}$ can be selected, as follows:
\begin{equation}
\mathbb{P}^+_{P} = \argmin_{\mathcal{O}_{\mathbb{P}}, \mathcal{M}_{\mathbb{P}}} \mathcal{E}_P.
\label{eq:GettingBest}
\end{equation}
Obviously, other approaches can be preferred as well.
These can be related to the peculiar geometry of the problem, distortions in $P$, adopted technology and common sense.

If $\mathbb{P}^+_P$ meets the requirements related to overlap and misplacement, then an acceptable solution is found (line 9) and the algorithm exits.
Otherwise, it is necessary to remove one module from the corresponding set $\mathbb{S}^+_{P}$.
To this aim, in general we induce a \emph{total order} on the pairs $(|\mathcal{O}_{m_i}|, |\mathcal{N}_{m_i}|) \in \mathbb{R}^2$, where $m_i \in \mathbb{S}_{P}$.
\begin{definition}
Given two modules $m_i$ and $m_j \in \mathbb{S}_P$, we define a lexicographic order on $\mathbb{S}_P$, as we define it as $(\mathbb{S}_P, \geq)$, such that:
\begin{equation*}
(|\mathcal{O}_{m_i}|, |\mathcal{N}_{m_i}|) \geq (|\mathcal{O}_{m_j}|, |\mathcal{N}_{m_j}|)
\end{equation*}
if an only if:
\begin{enumerate}
\item $|\mathcal{N}_{m_i}| < |\mathcal{N}_{m_j}|$, 
\item $|\mathcal{N}_{m_i}| = |\mathcal{N}_{m_j}|$ and $|\mathcal{O}_{m_i}| \geq |\mathcal{O}_{m_j}|$.
\end{enumerate}
\end{definition}
Since we want to remove the module $m^*_i \in \mathbb{S}^+_{P}$ which has the \emph{lowest} number of neighbours and an above-than-average overlap, RemoveModuleFrom() selects one such module such that (i) $m^*_i \in \mathbb{S}_{P}$ and (ii) $\mathcal{O}_{m^*_i} \geq avg(\mathcal{O}_{m^*_i})$ (line 12).
When the module is removed, a new set $\mathbb{S}^c_{P}$ is obtained and the algorithm iterates.

\section{Use Cases}
\label{sec:cases}

This Section introduces first a few notes about the implementation of the algorithm, and then presents three uses cases involving different robot body parts, one from the iCub robot, and two from a Schunk manipulator.

\subsection{Implementation}

\textsc{ItPla} has been implemented both in C++ and Python to broaden its users base.

The C++ version integrates the CGAL computational geometry library\footnote{Web: \url{www.cgal.org}.} and Box2D\footnote{Web: \url{box2d.org/}.} to simulate polygon's motions. 
This version also includes a graphical user interface developed using the QT libraries\footnote{Web: \url{www.qt.io/}.}.
The C++ version can be considered as a sort of mock-up:
the runtime environment is not easy to set-up, it may be difficult to add new features because of poor architectural design choices, and the computation of minimum distances is just an approximation.
As far as the Python version is concerned, the algorithm natively computes 2D calculations as well as physics simulations, and integrates a Python-based graphical user interface. 
It uses the pip package manager, and integrates such libraries as NumPy\footnote{Web: \url{www.numpy.org/}.}, Box2D and PySDL2\footnote{Web: \url{www.pygame.org/}.}.
With respect to the C++ version, it has a number of advantages:
(i) setting up the environment is quite easy;
(ii) the code is simpler and more clear;
(iii) the computation of the lowest distance is accurate;
(iv) the number of lines of code is around $200$ versus more than $1000$ of the C++ version.  
As anticipated in Section \ref{sec:introduction}, both versions are available open source.

\begin{figure*}[t!]
\centering
\subfigure[]{
    \includegraphics[width=0.14\hsize]
        {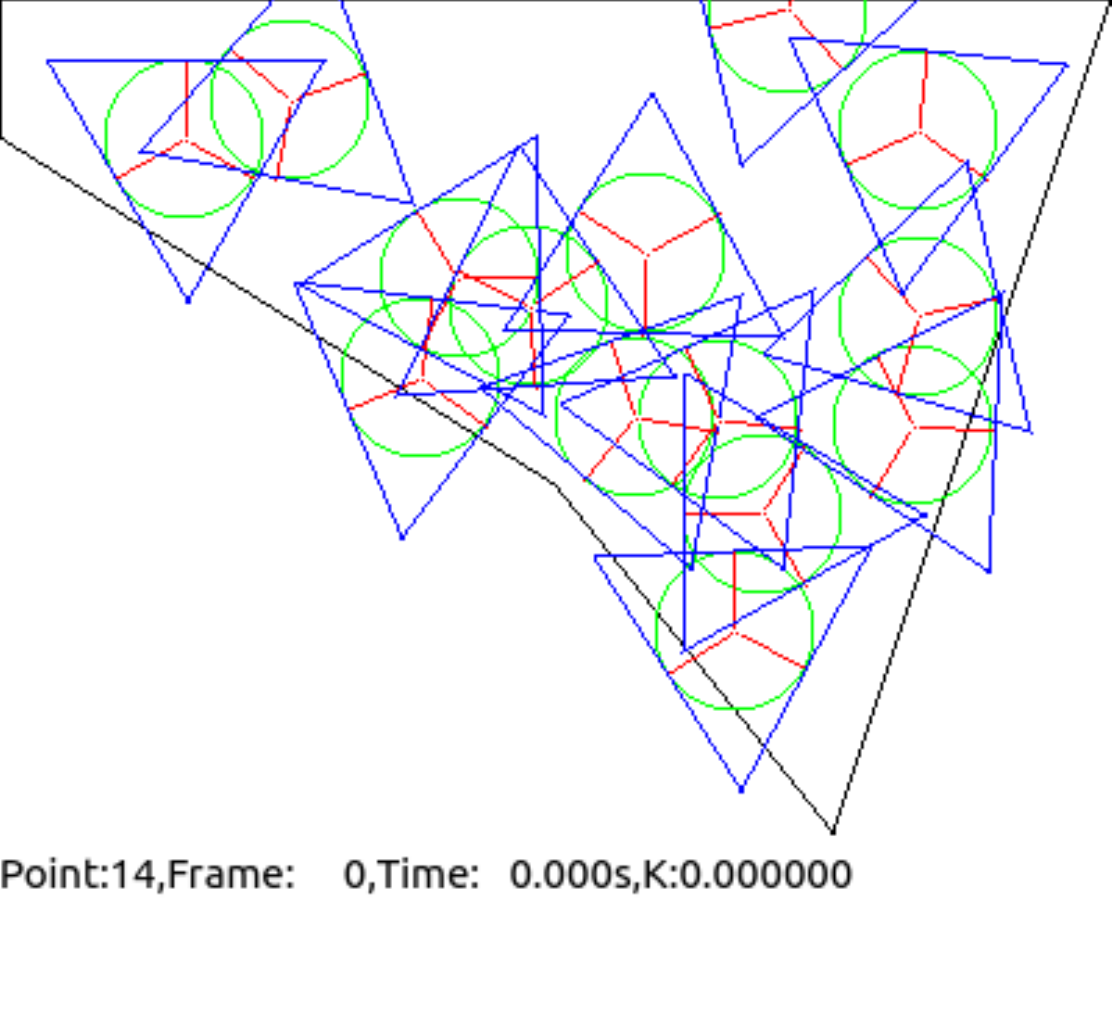}
    \label{fig:result:0}
}
\hspace{0\hsize}
\subfigure[]{
    \includegraphics[width=0.14\hsize]
        {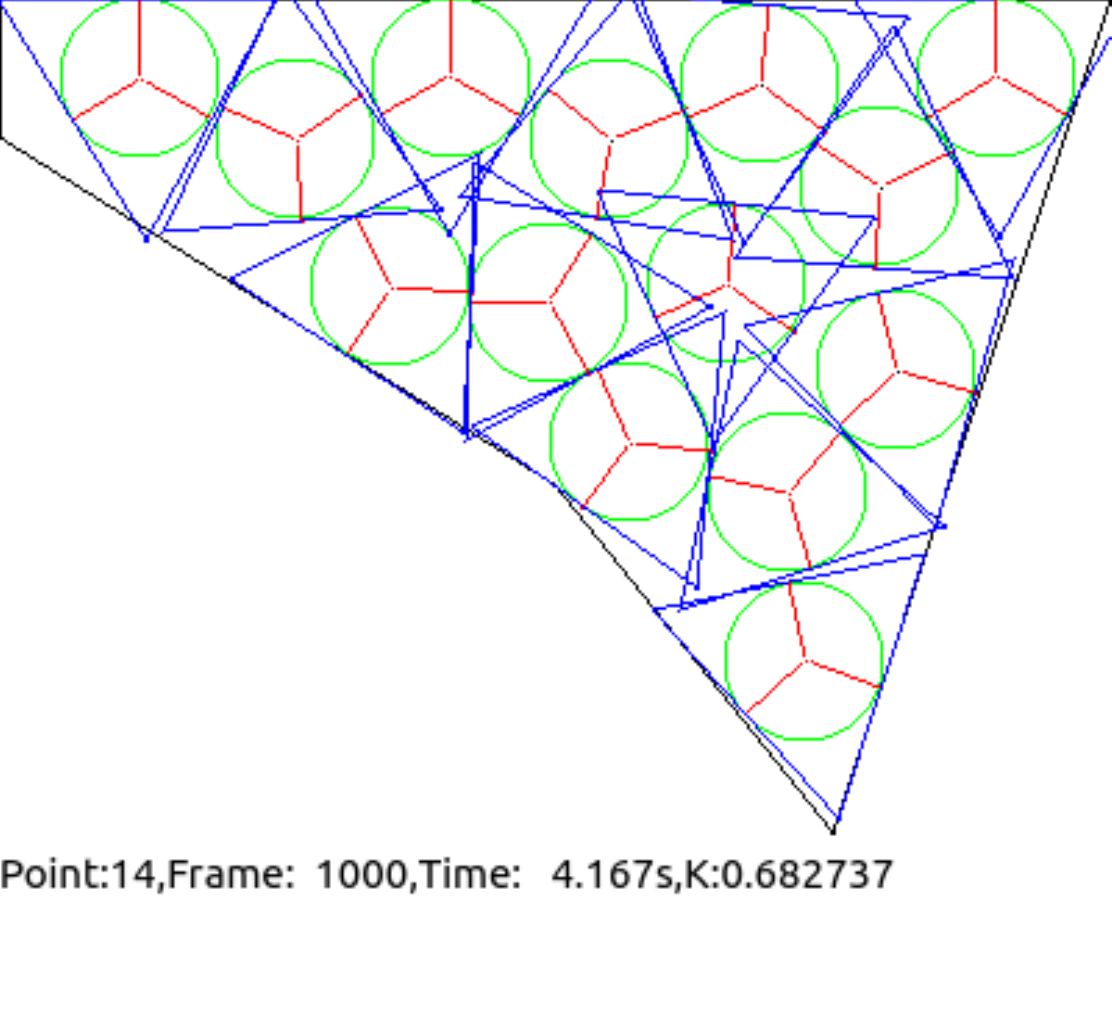}
    \label{fig:result:1}
}
\hspace{0\hsize}
\subfigure[]{
    \includegraphics[width=0.14\hsize]
        {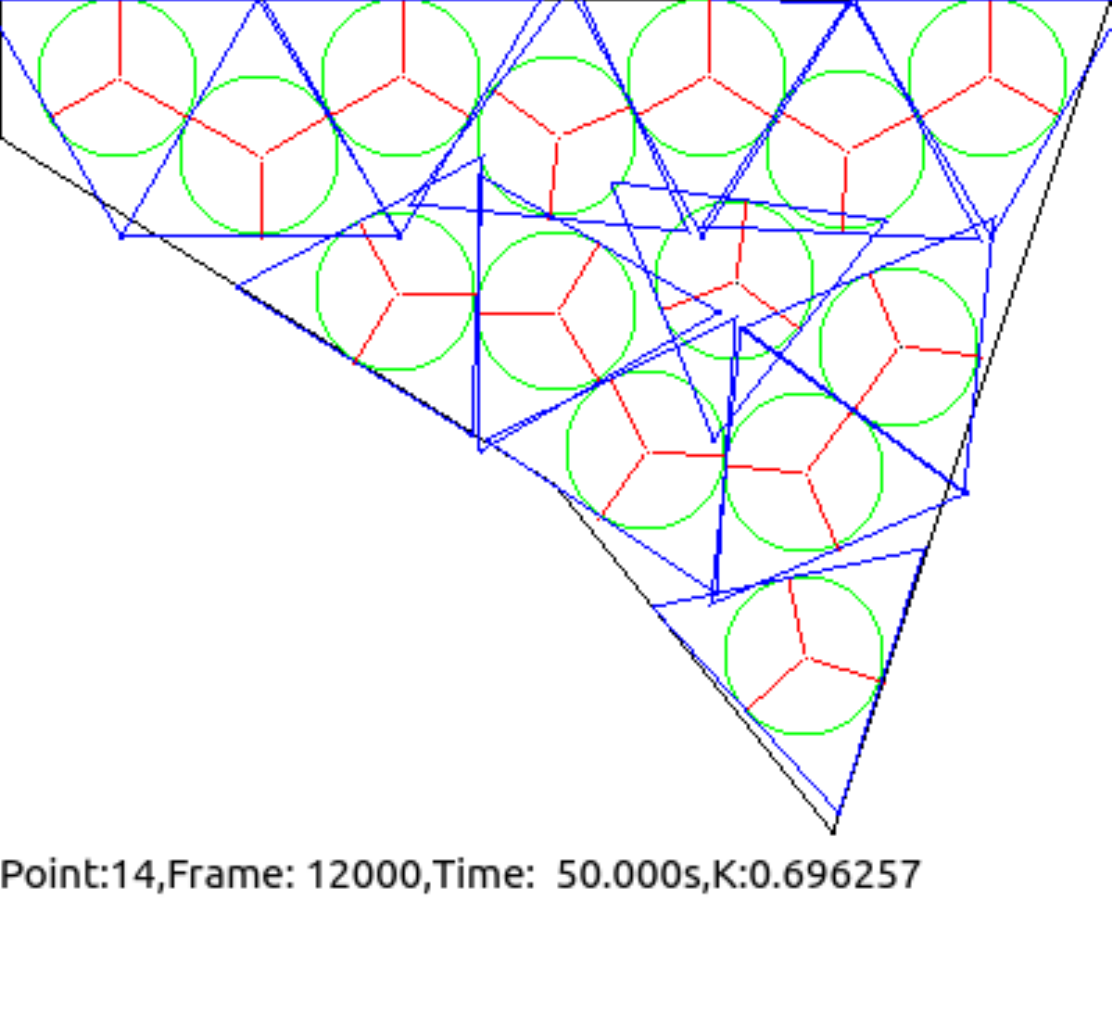}
    \label{fig:result:2}
}
\hspace{0\hsize}
\subfigure[]{
    \includegraphics[width=0.14\hsize]
        {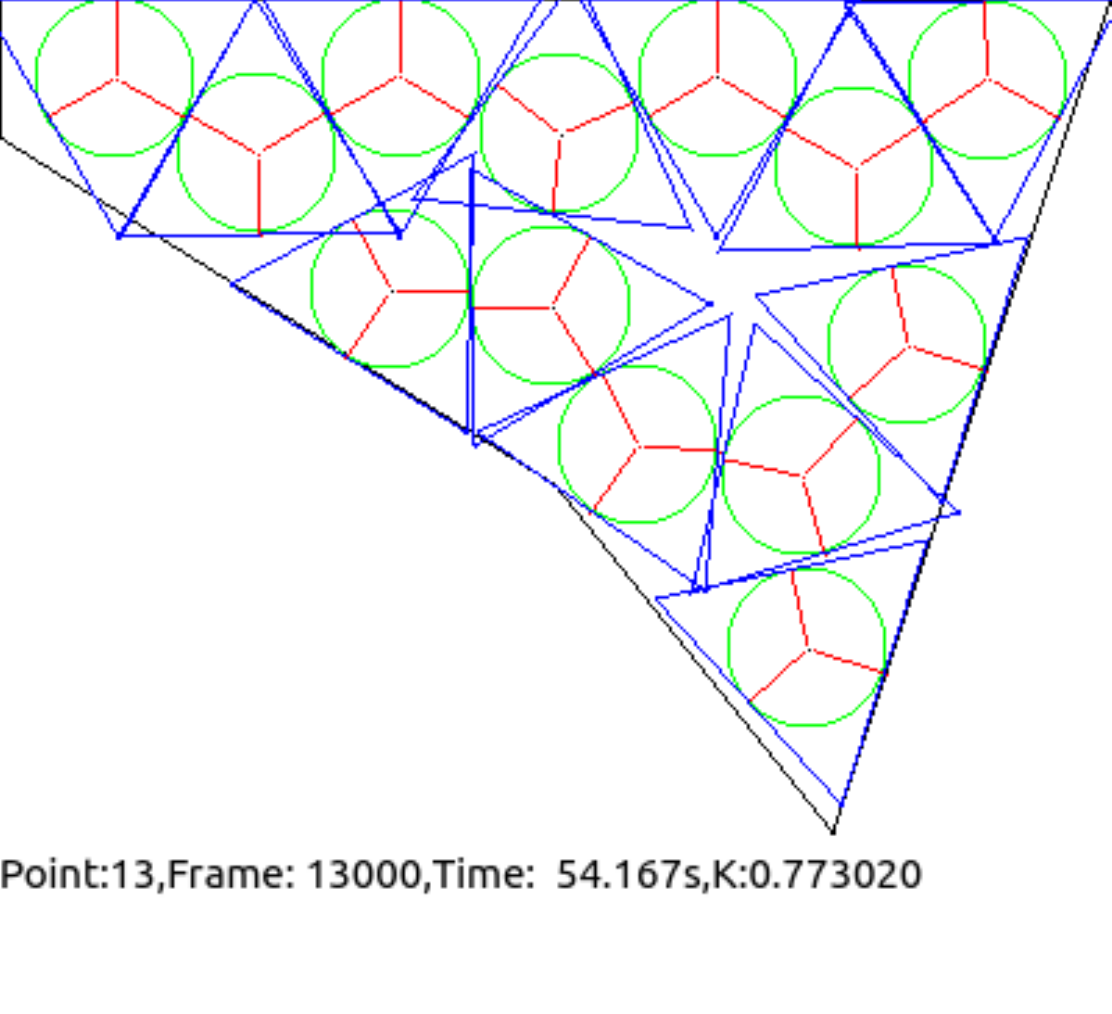}
    \label{fig:result:3}
}
\subfigure[]{
    \includegraphics[width=0.14\hsize]
        {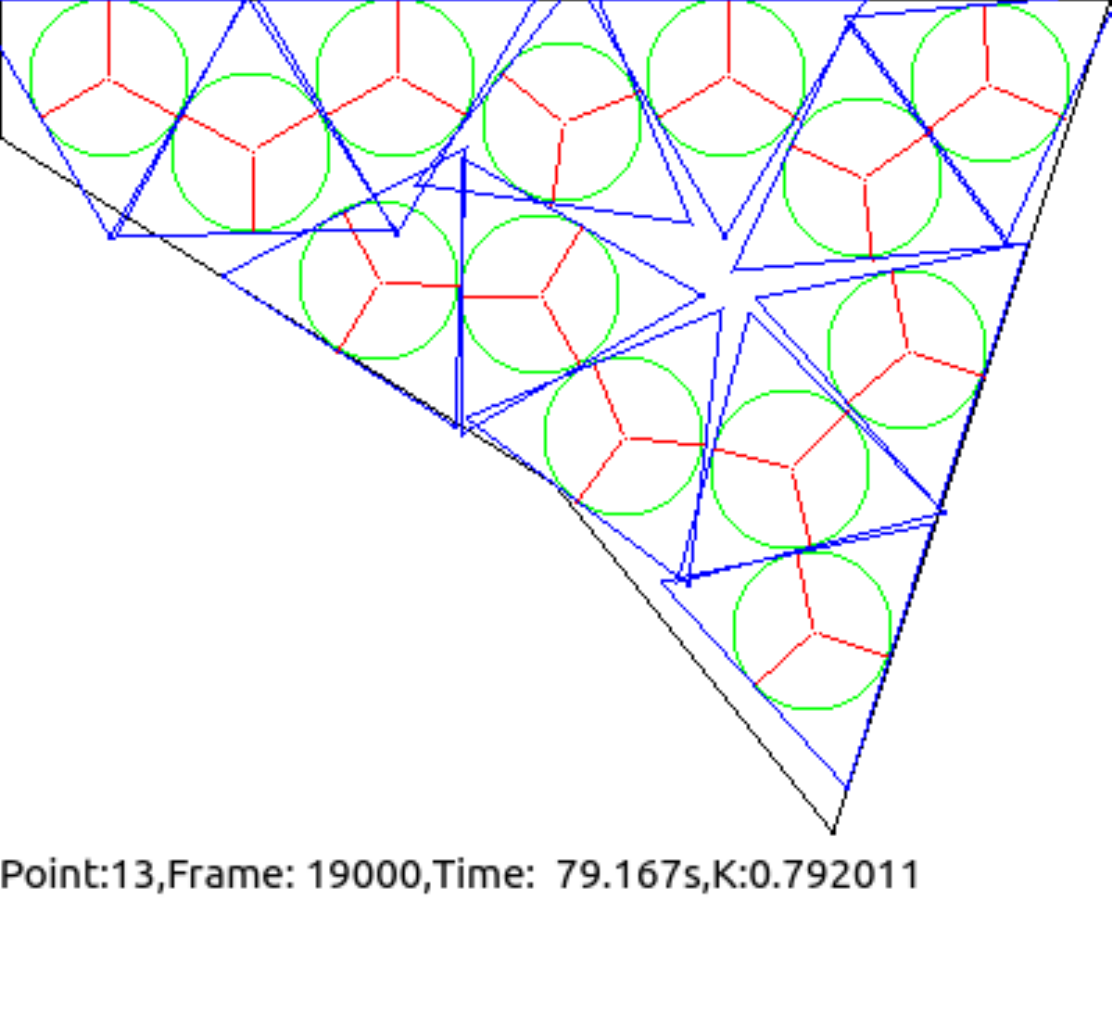}
    \label{fig:result:4}
}
\hspace{0\hsize}
\subfigure[]{
    \includegraphics[width=0.14\hsize]
        {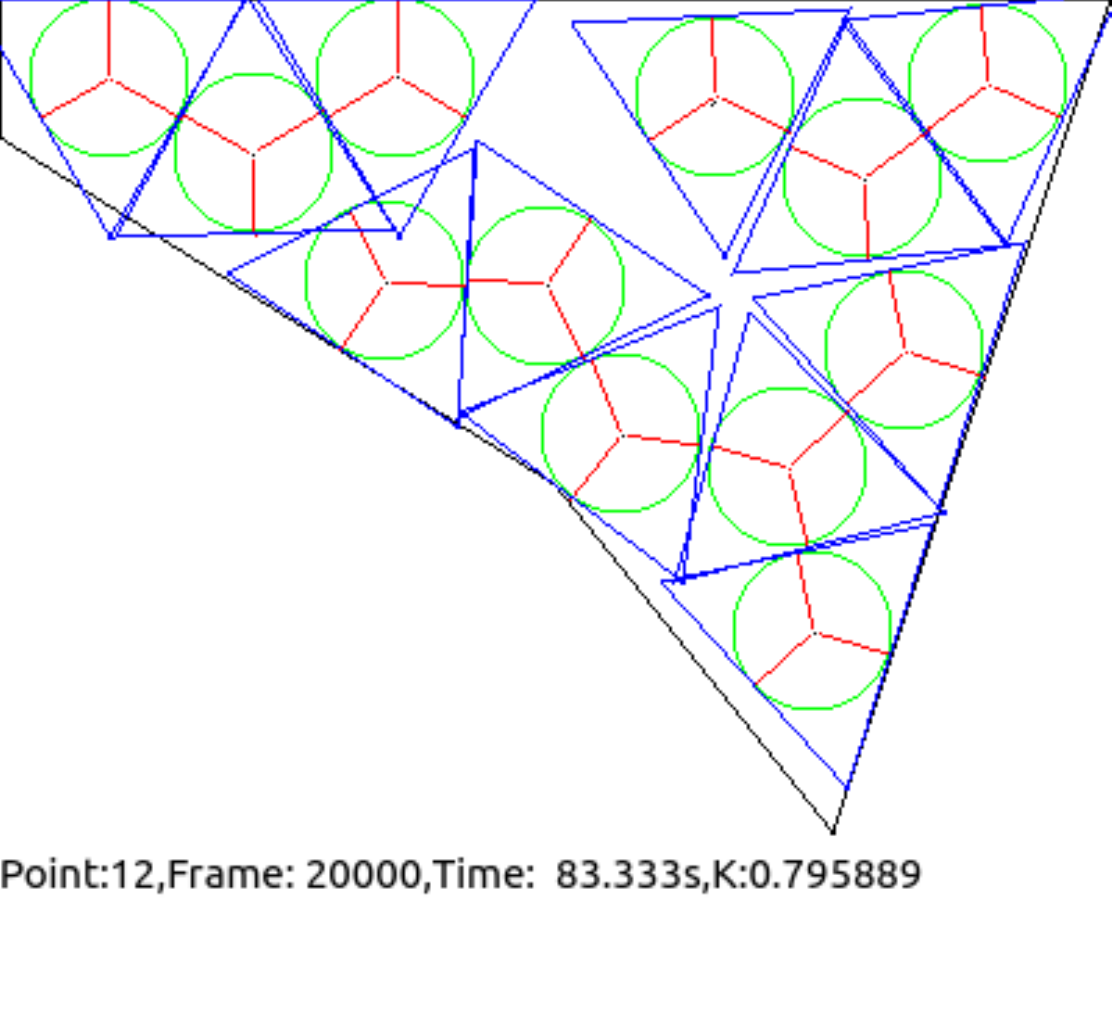}
    \label{fig:result:5}
}
\\
\hspace{0\hsize}
\subfigure[]{
    \includegraphics[width=0.14\hsize]
        {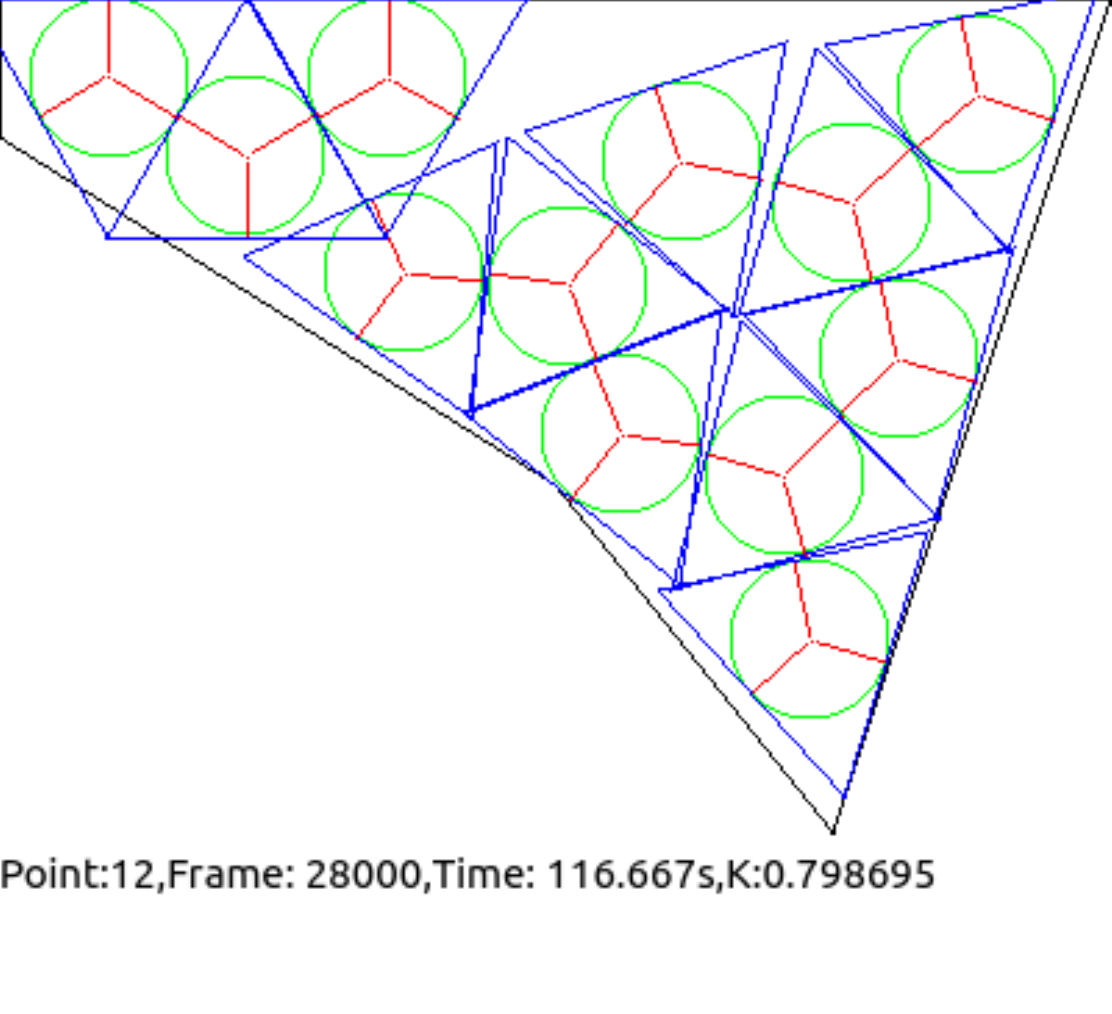}
    \label{fig:result:6}
}
\hspace{0\hsize}
\subfigure[]{
    \includegraphics[width=0.14\hsize]
        {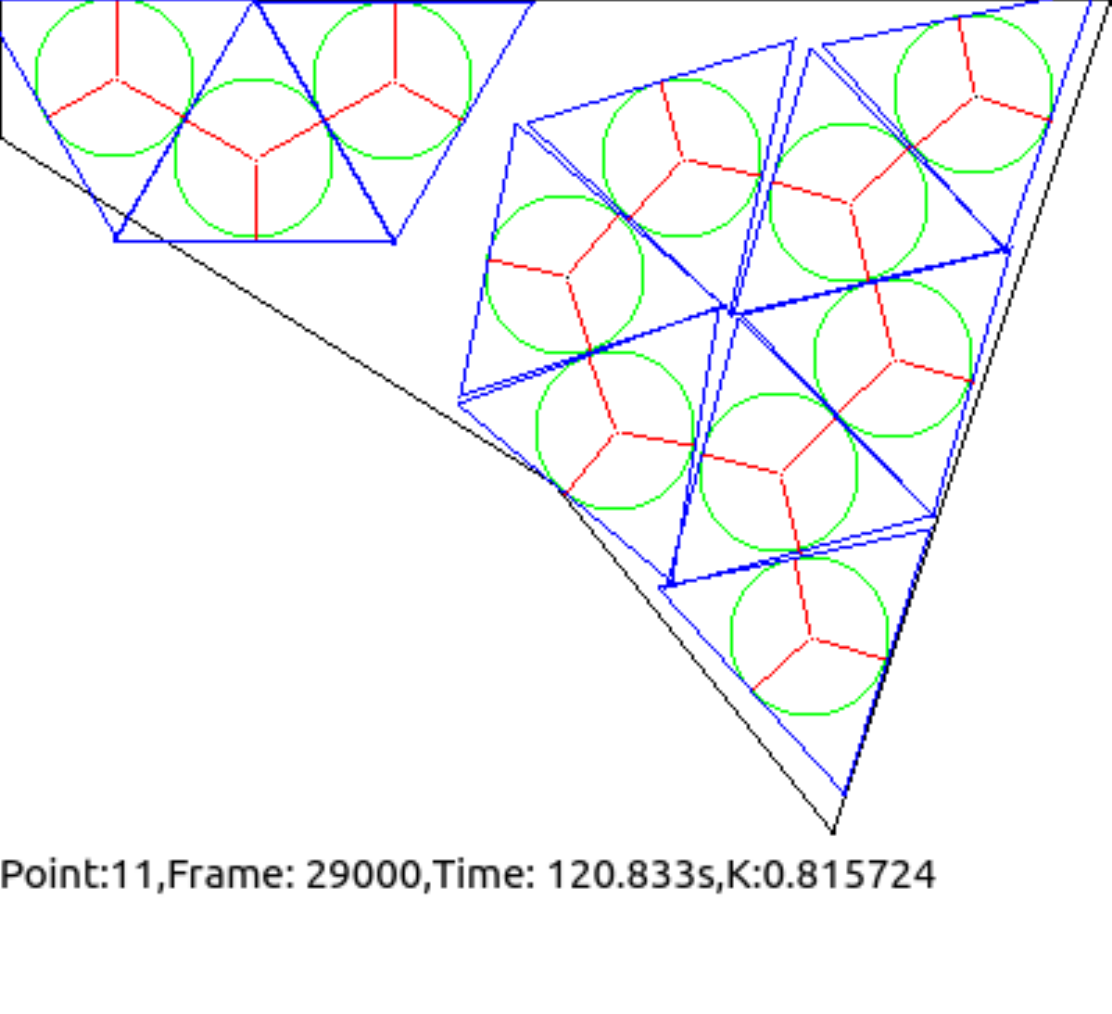}
    \label{fig:result:7}
}
\subfigure[]{
    \includegraphics[width=0.14\hsize]
        {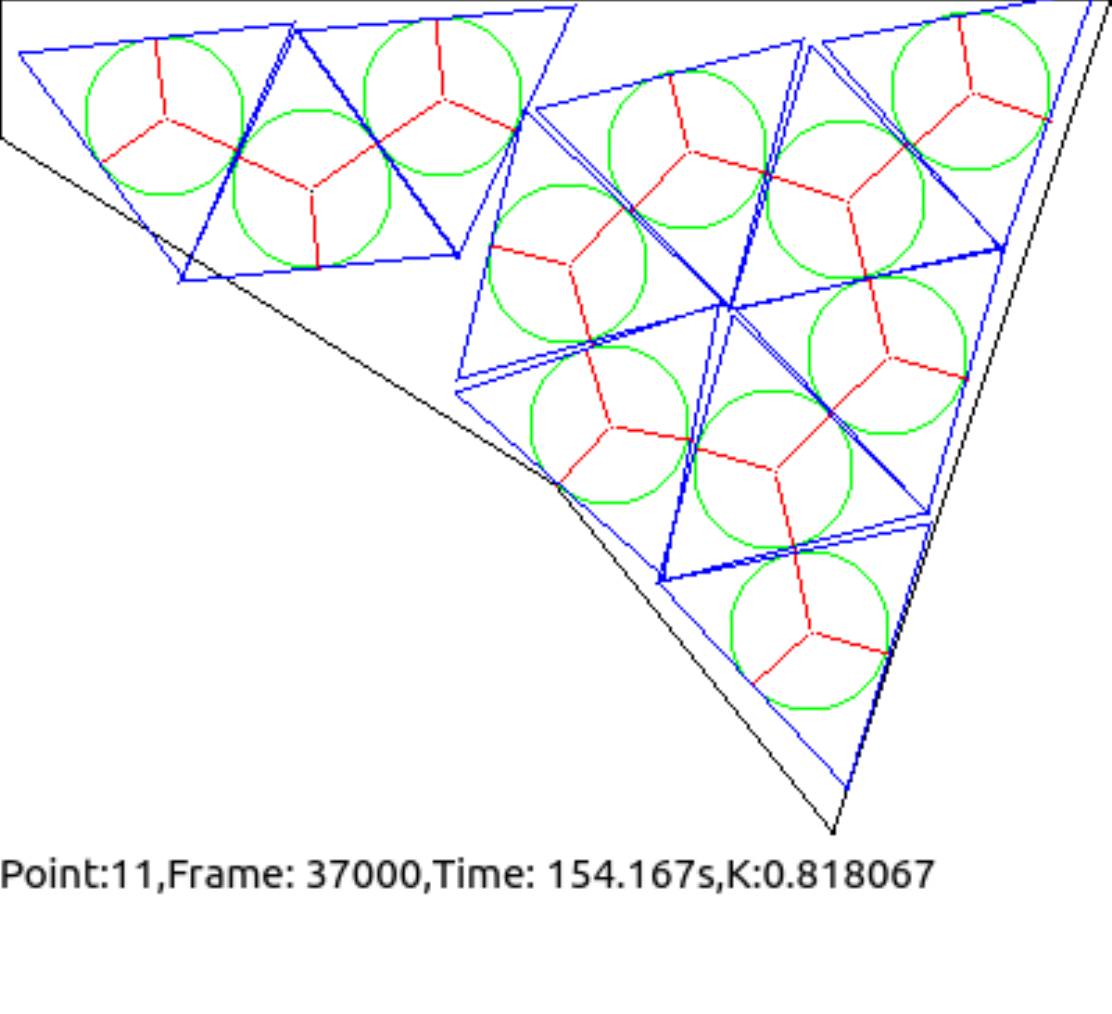}
    \label{fig:result:8}
}
\hspace{0\hsize}
\subfigure[]{
    \includegraphics[width=0.14\hsize]
        {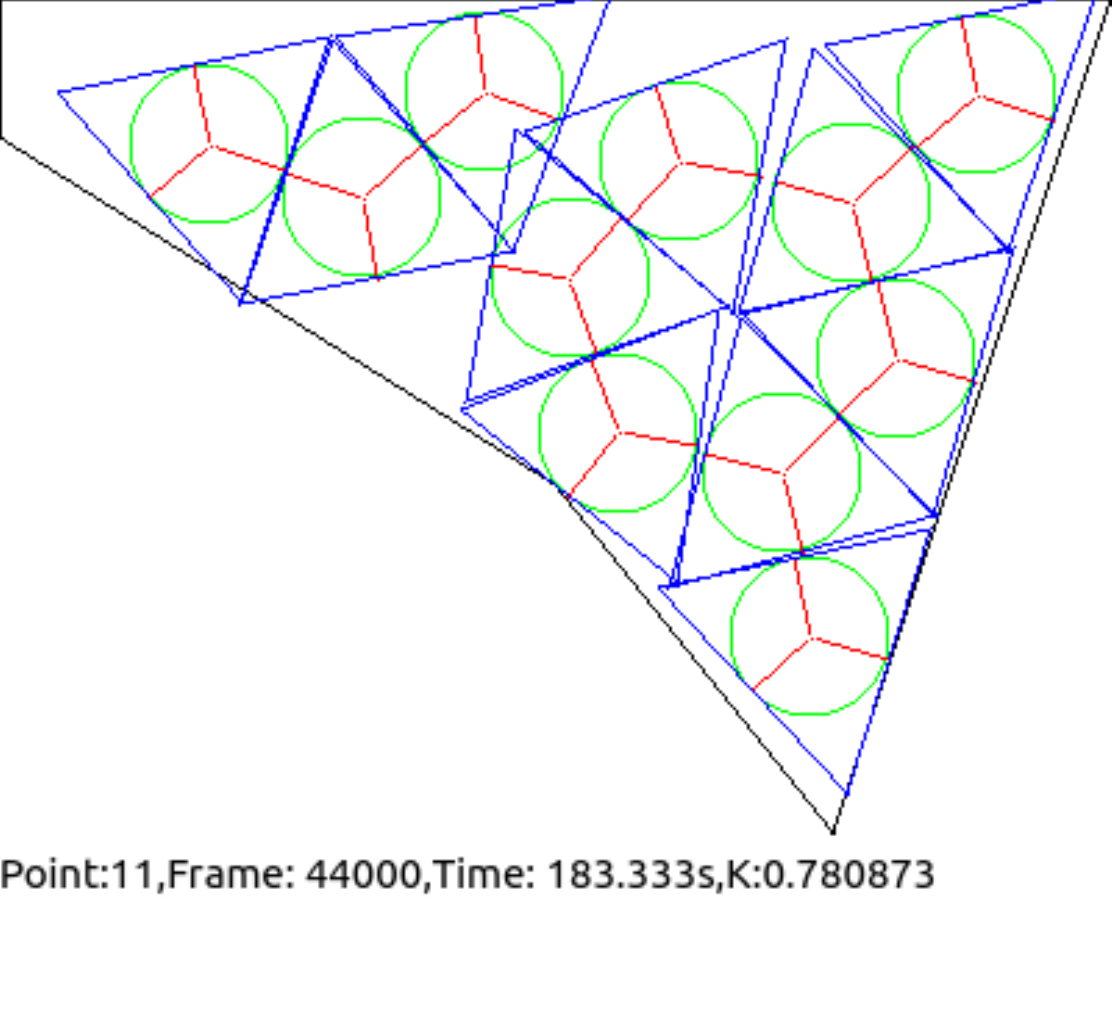}
    \label{fig:result:9}
}
\hspace{0\hsize}
\subfigure[]{
    \includegraphics[width=0.14\hsize]
        {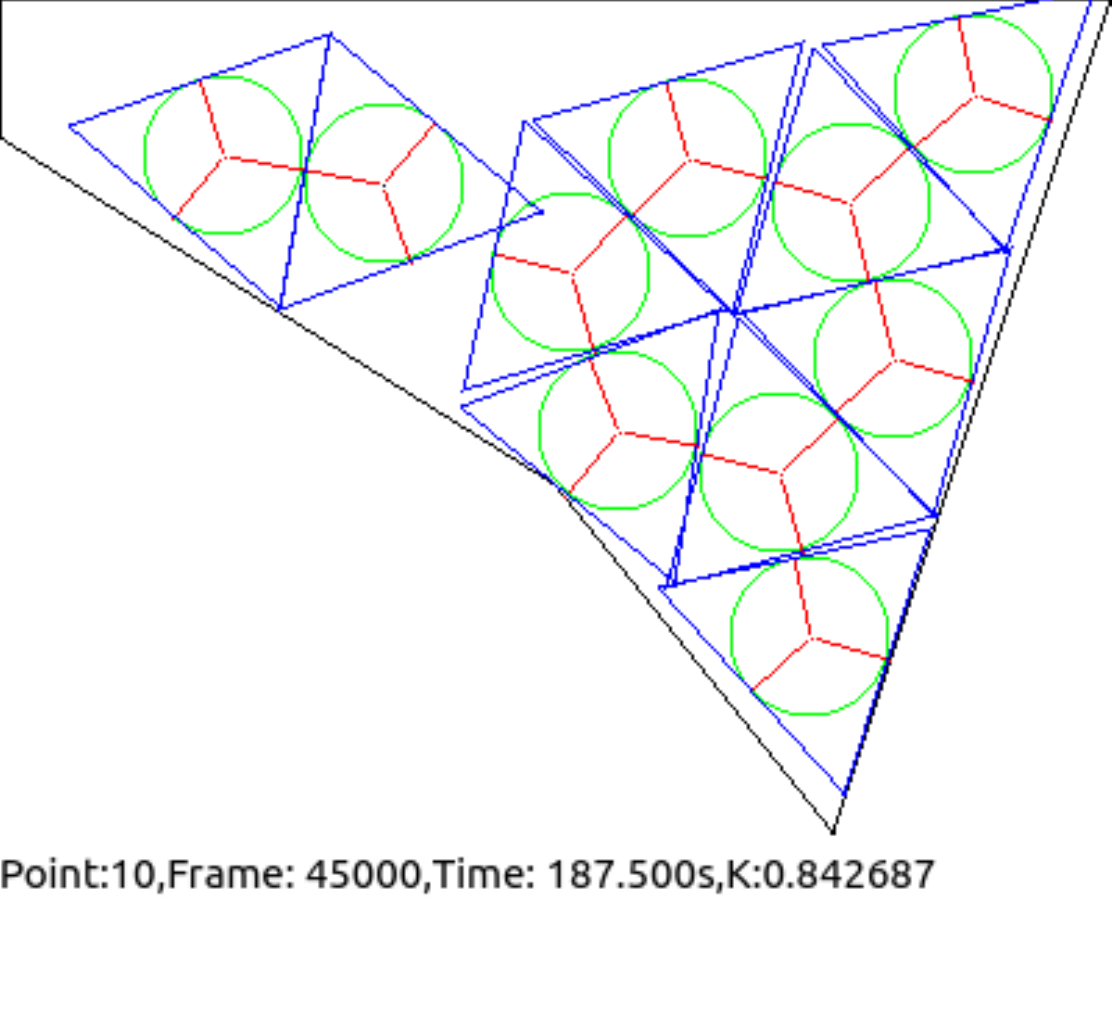}
    \label{fig:result:10}
}
\hspace{0\hsize}
\subfigure[]{
    \includegraphics[width=0.14\hsize]
        {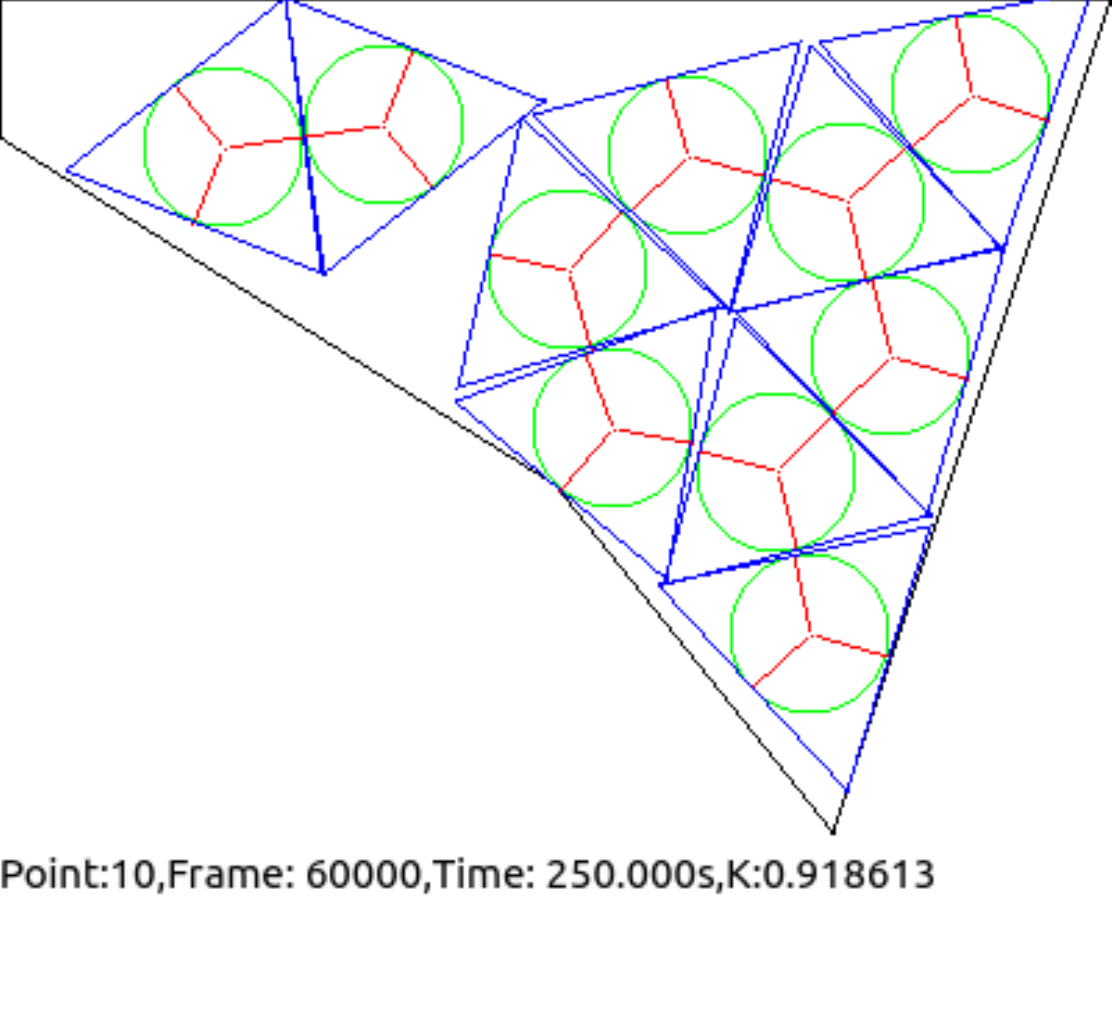}
    \label{fig:result:11}
}
\caption{The \textsc{ItPla}'s evolution process for the iCub robot's left hip.}
\label{fig:result_iCub}
\end{figure*}

\subsection{Use Case $1$: the iCub's Left Hip}
This use case illustrates how \textsc{ItPla} behaves using the simplified version of the iCub's left hip used in \cite{Anghinolfietal2013}.

Figure \ref{fig:result_iCub} shows a few significant steps in the iteration process.
First of all, given a flat representation of the hip surface as a polygon, the theoretically maximum number of triangular modules is computed using \eqref{eq:UpperBound}.
As a consequence, a certain amount of triangular modules are generated and randomly placed inside the polygon, as shown in Figure \ref{fig:result:0}.
Obviously enough, triangle's sides can actually stick out from the polygon's area.
Then, the iteration phase of the algorithm starts, and the modules tend to distribute uniformly as per effect of the attractive and repulsive forces, leading to the situation in Figure \ref{fig:result:1}, as described in \eqref{eq:SumPseudoForcesWithEdges}.
In this case, a part from a few notable exceptions, all modules turn to be uniformly distributed.
In Figure \ref{fig:result:2}, it is possible to see how pseudo-forces tend to remove overlaps, although some of them seem to be unavoidable. 
Since a stable but unacceptable (i.e., it contains overlaps) placement is obtained, \textsc{ItPla} needs to remove one triangular module, which is done as shown in Figure \ref{fig:result:3}. 
As discussed above, the removed module is the one with the lowest number of neighbours and the highest overlap area. 
Similar removal operations are performed in the situations depicted in Figure \ref{fig:result:5}, Figure \ref{fig:result:7}, and Figure \ref{fig:result:10}.
The acceptable placement shown in Figure \ref{fig:result:11} is finally obtained, which terminates the algorithm.
It is noteworthy that, with respect to the results in \cite{Anghinolfietal2013}, the number of triangular modules that are accommodated inside the polygon raises from $9$ to $10$, which amounts to a $11.11\%$ increase in coverage.

\begin{figure}[t!]
\centering
\includegraphics[width=0.9\hsize]
{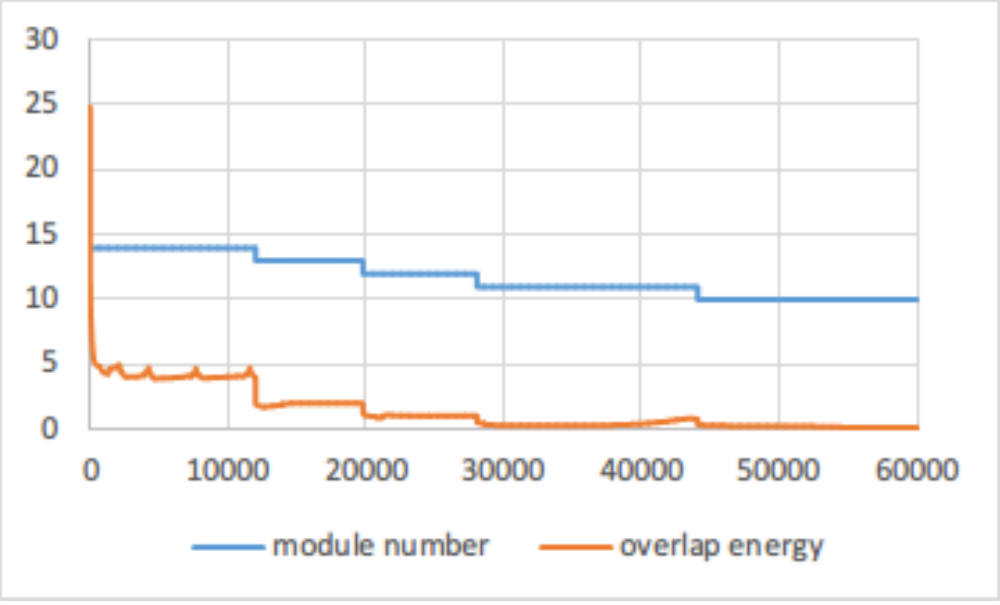}
\caption{The evolution of the number of modules and the overlap area during the process shown in Figure \ref{fig:result_iCub}.}
\label{fig:chart}
\end{figure}
The trend in both the number of modules and the overlap area associated with the process described above is shown in Figure \ref{fig:chart}.
It is possible to observe that the number of modules decreases as per effect of module removals, whereas the overlap area tends to stabilise and sometimes to increase during the iterative procedure.
This phenomenon is caused by the employed weighting mechanism.
In \textsc{ItPla}, pseudo-force effects are computed taking into account the amount of overlap multiplied by a gain, i.e., the weight, which means that such functions produce gain-dependent results, i.e., their effect is uniquely defined only subject to a specific choice of gains.
However, this does not affect the procedure since the overlap area is used only for evaluations, therefore having little influence on the final results.

\begin{figure}[t!]
\centering
\includegraphics[width=0.9\hsize]
{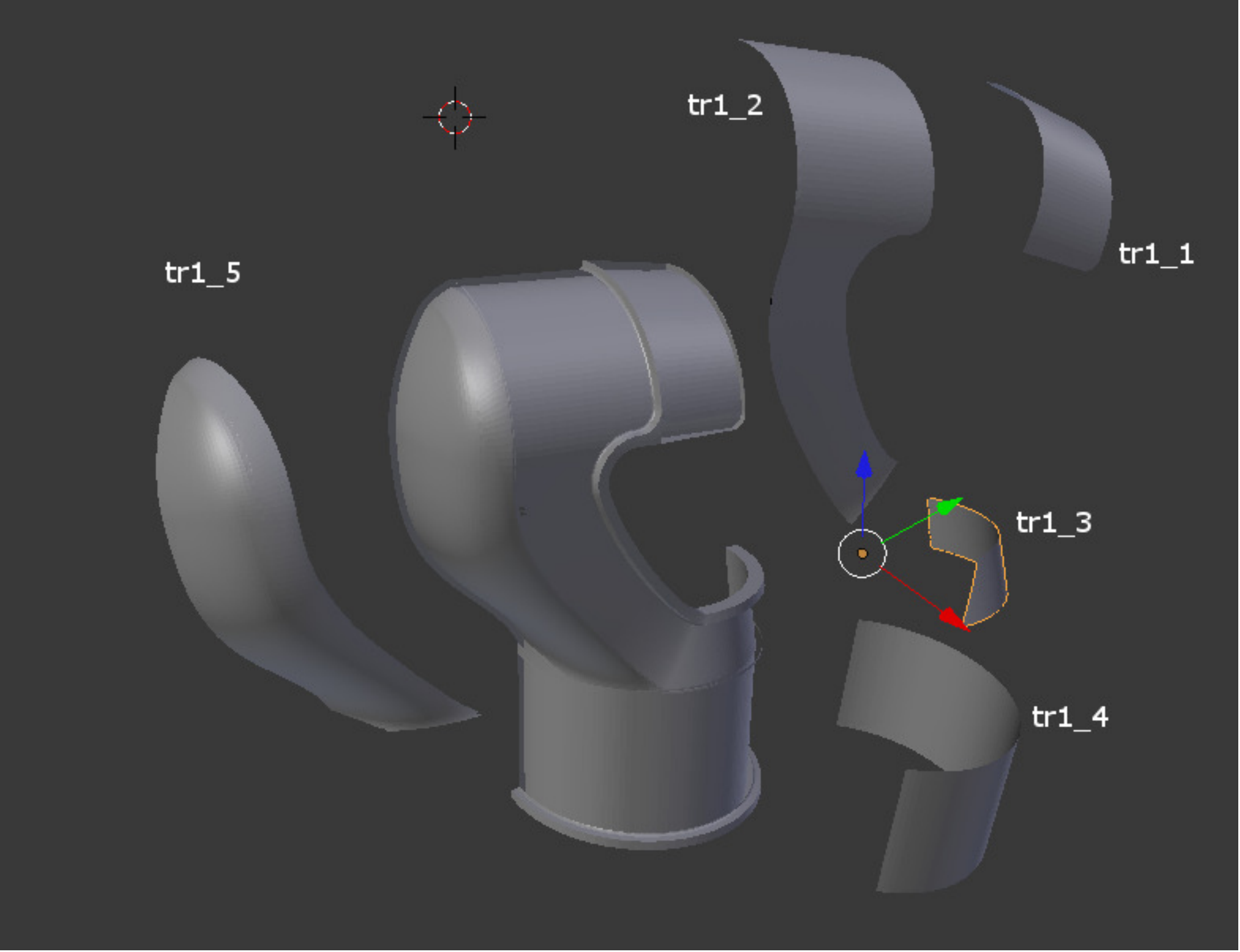}
\caption{A Schunk manipulator's link with five different surfaces to cover.}
\label{fig:tr1}
\end{figure}

\subsection{Use Cases $2$ and $3$: a Schunk Manipulator's Link}

The second and third use cases are related to the coverage of a Schunk manipulator's link, as shown in Figure \ref{fig:tr1}.
The whole link's surface has been \textit{a priori} divided in five different parts, each one subject to a different flattening and coverage process. 
As it can be seen in the Figure, the five surfaces are referred to as $tr1\_1$, \ldots, $tr1\_5$.
Here, we focus only on $tr1\_2$ and $tr1\_5$, which are characterised by challenging \textit{irregular} shapes.
We do not consider $tr1\_3$ because its area is too small, neither do we consider $tr1\_1$ nor $tr1\_4$ because their flat shape is similar to a rectangle, which would be trivial to cover.  

\begin{figure*}[t!]
\centering
\subfigure[]{
    \includegraphics[width=0.18\hsize]
        {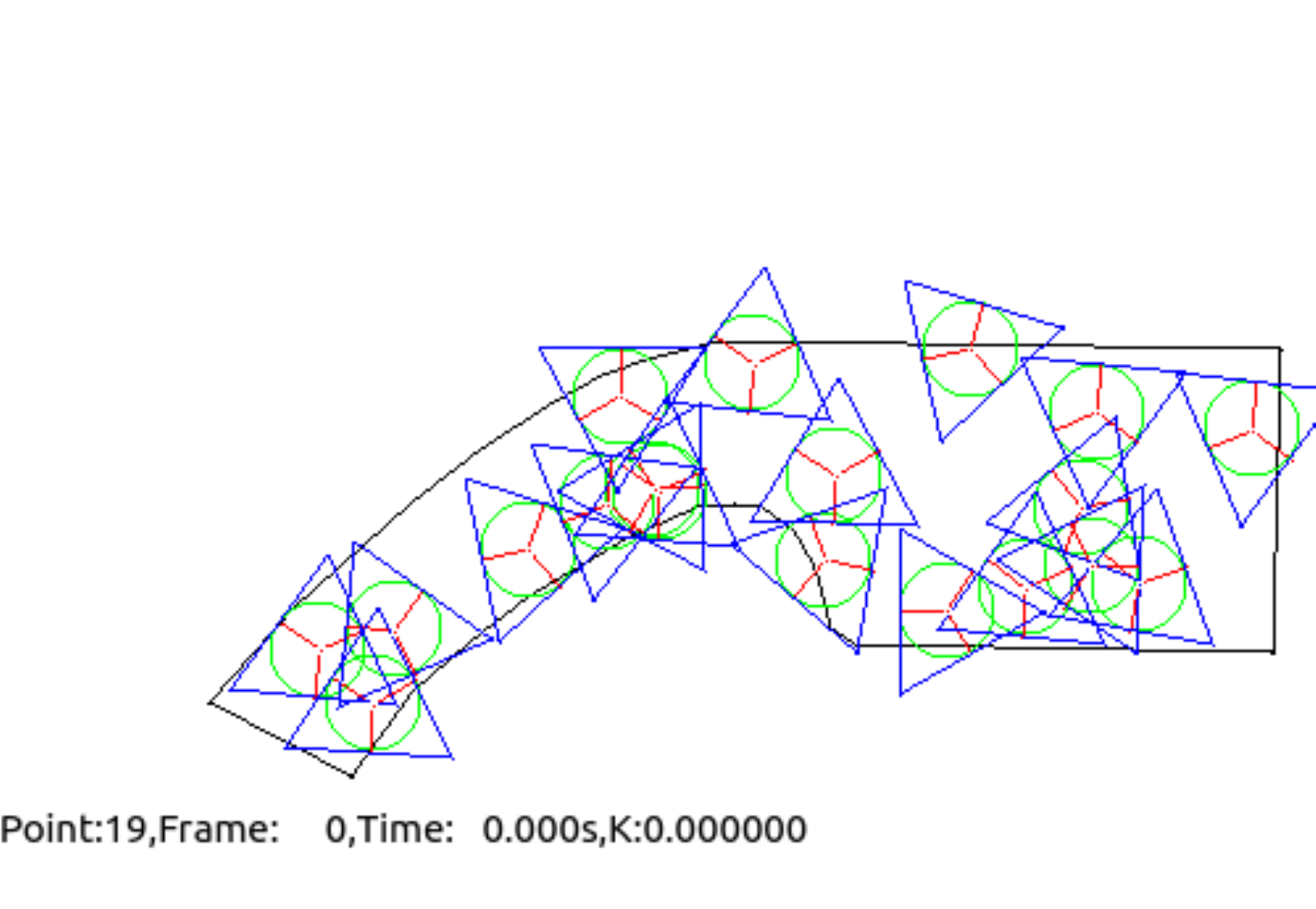}
    \label{fig:tr1_2:0}
}
\hspace{0\hsize}
\subfigure[]{
    \includegraphics[width=0.18\hsize]
        {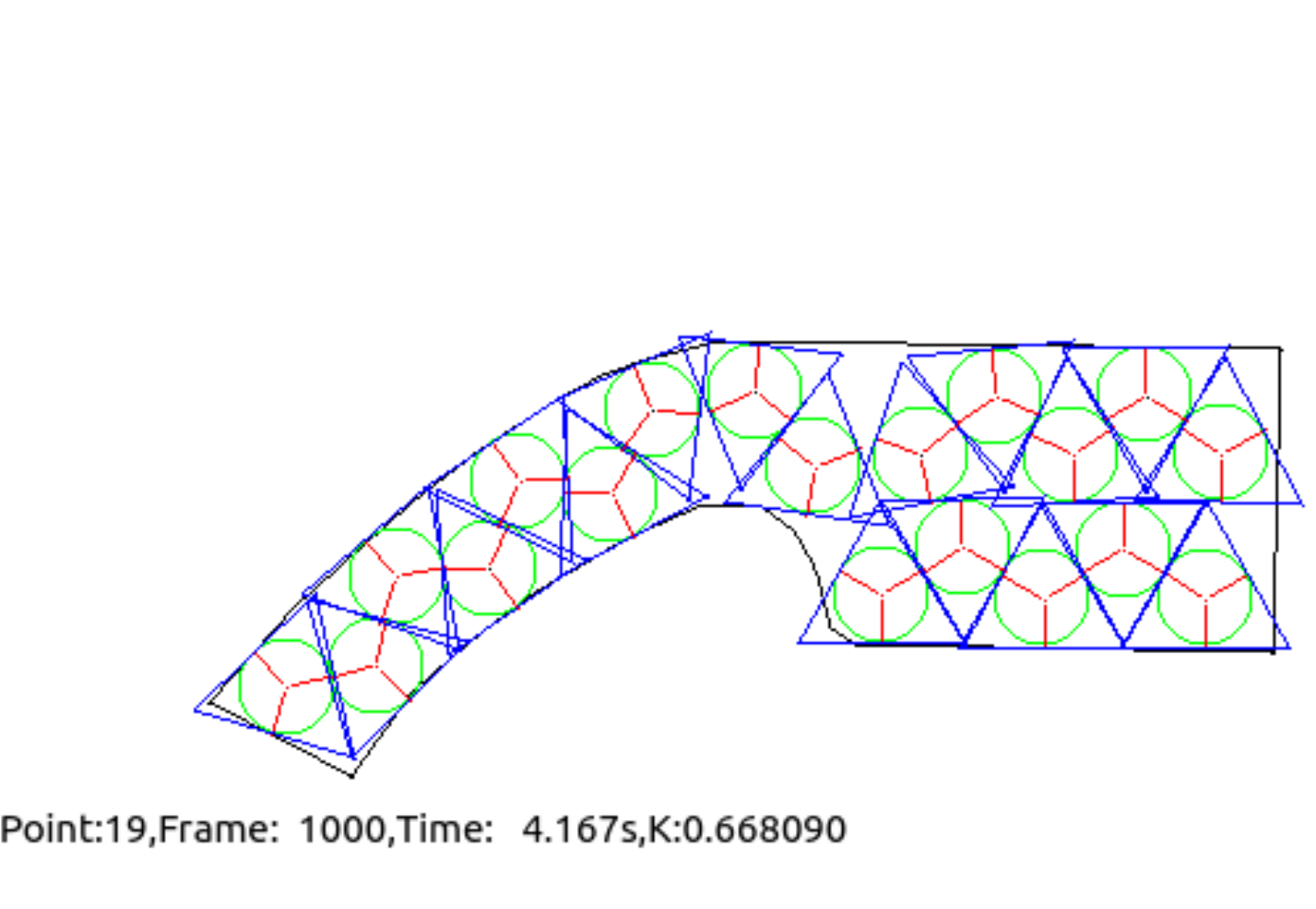}
    \label{fig:tr1_2:1}
}
\hspace{0\hsize}
\subfigure[]{
    \includegraphics[width=0.18\hsize]
        {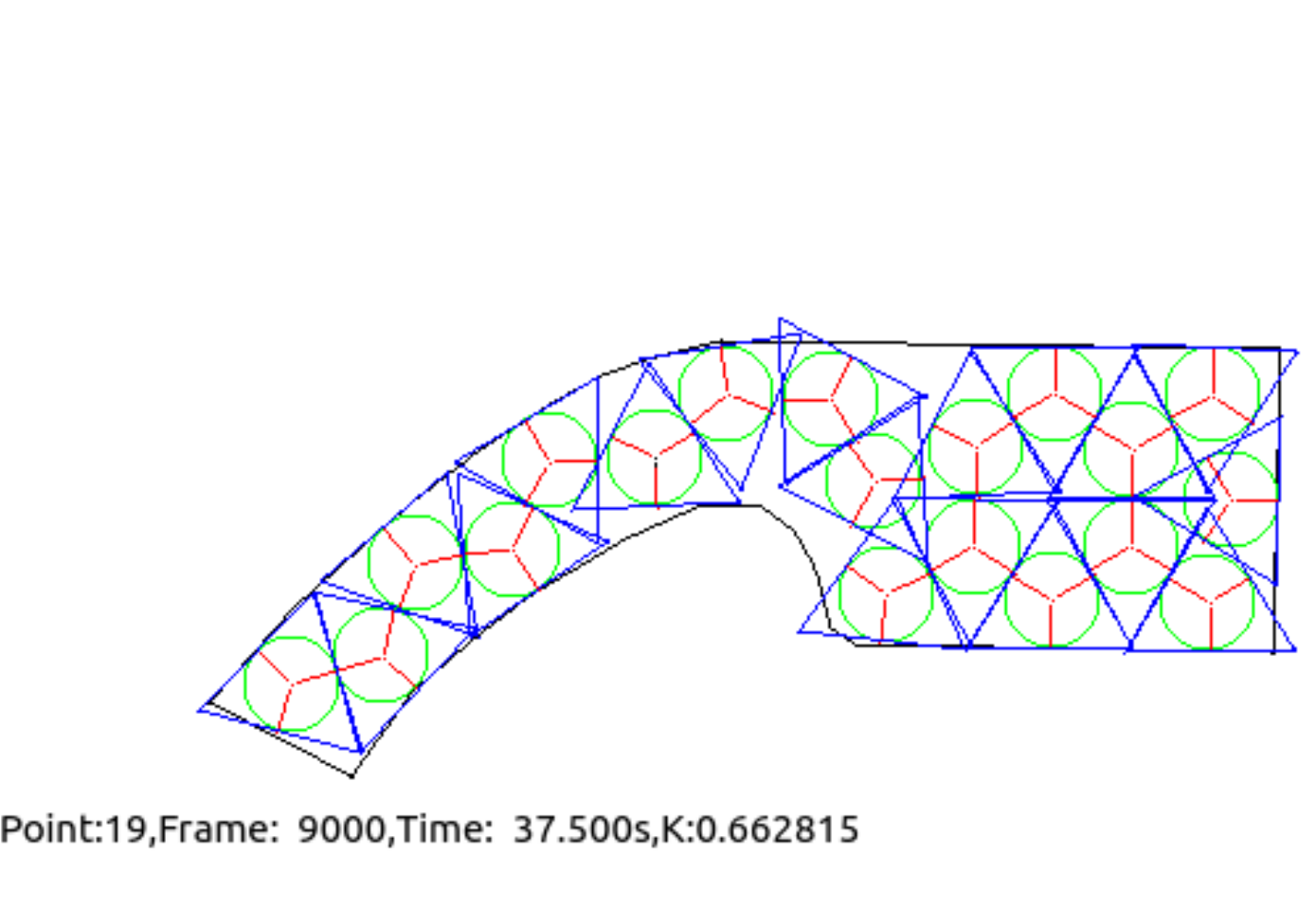}
    \label{fig:tr1_2:2}
}
\subfigure[]{
    \includegraphics[width=0.18\hsize]
        {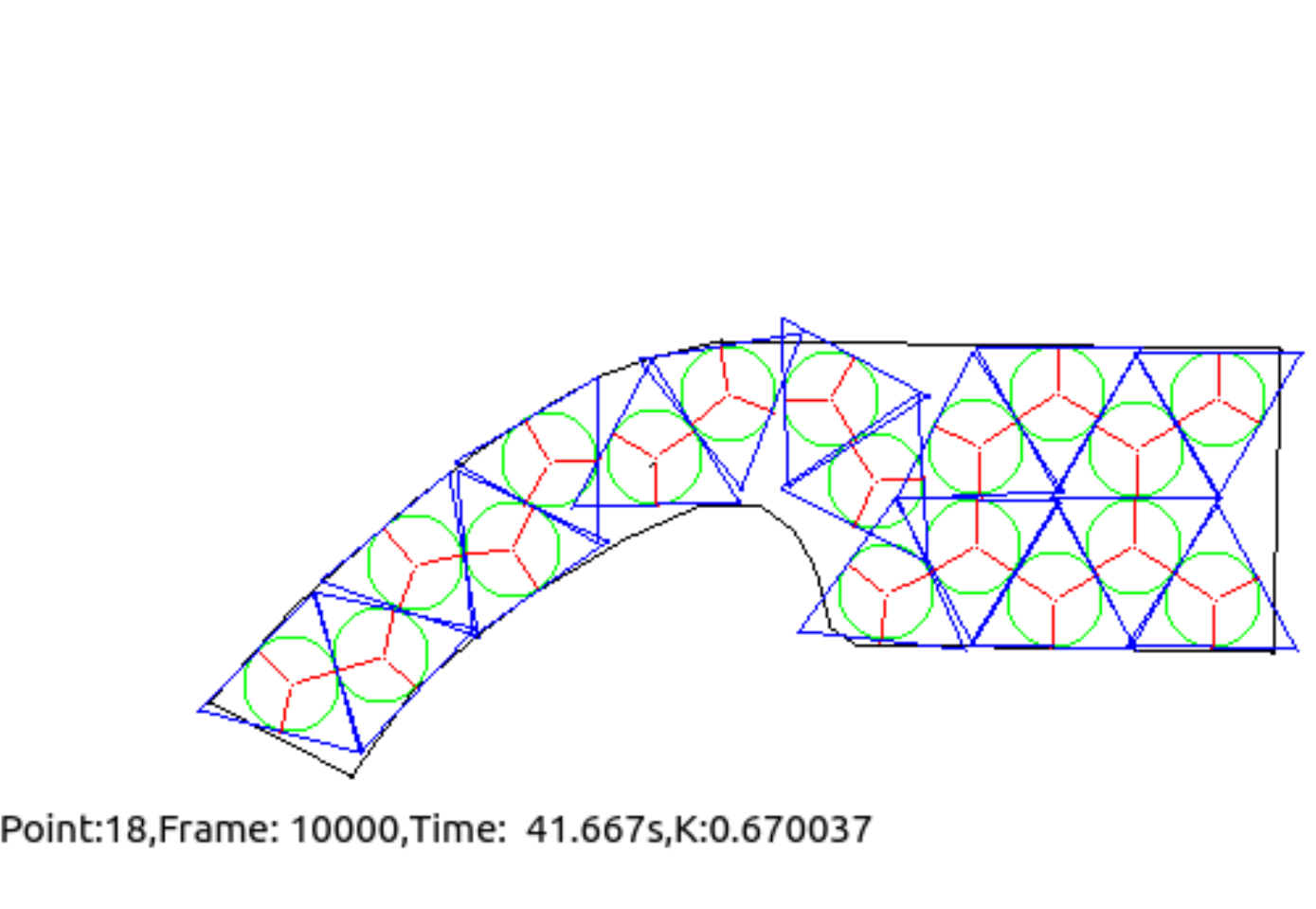}
    \label{fig:tr1_2:3}
}
\hspace{0\hsize}
\subfigure[]{
    \includegraphics[width=0.18\hsize]
        {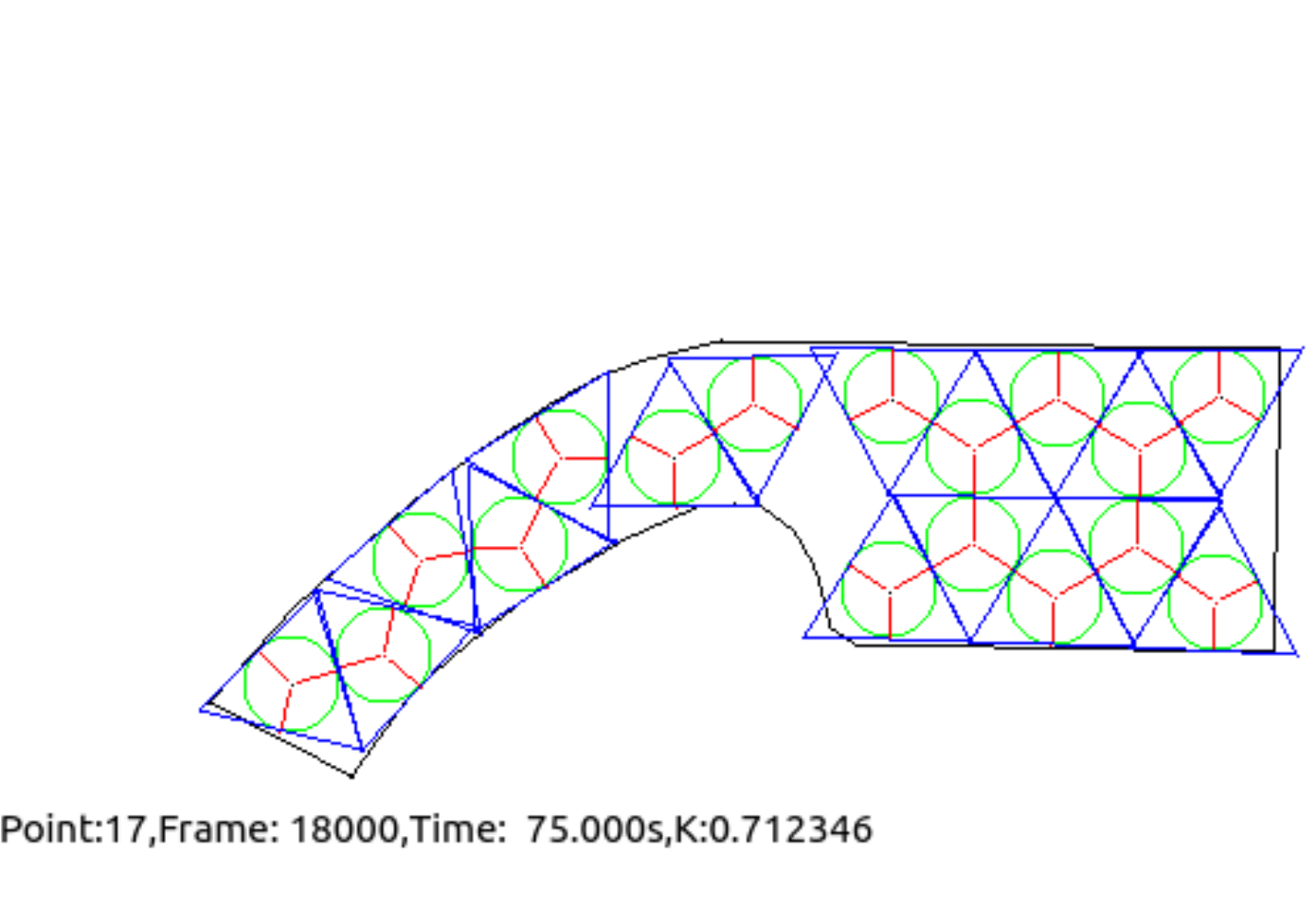}
    \label{fig:tr1_2:5}
}
\\
\subfigure[]{
    \includegraphics[width=0.18\hsize]
        {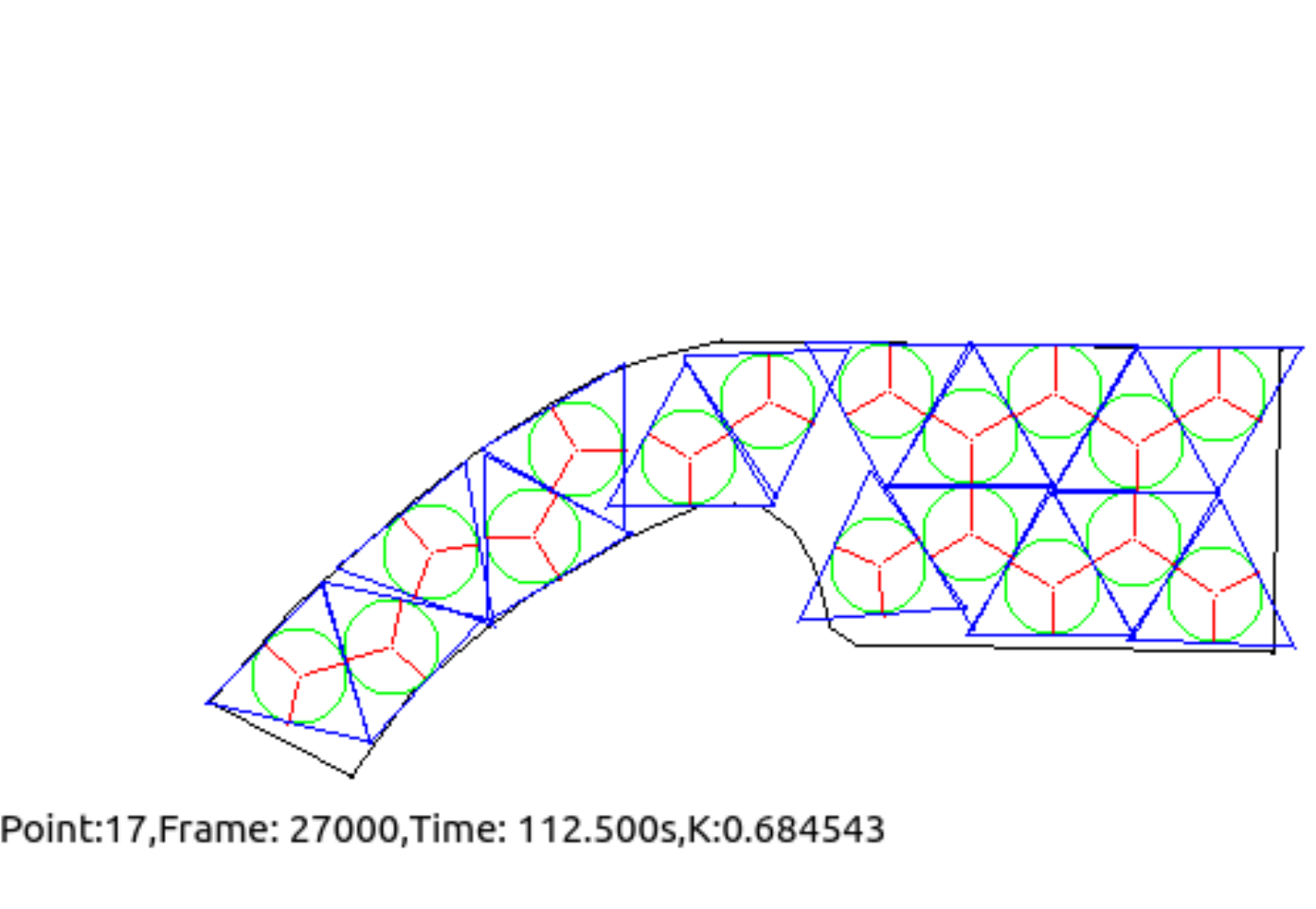}
    \label{fig:tr1_2:6}
}
\hspace{0\hsize}
\subfigure[]{
    \includegraphics[width=0.18\hsize]
        {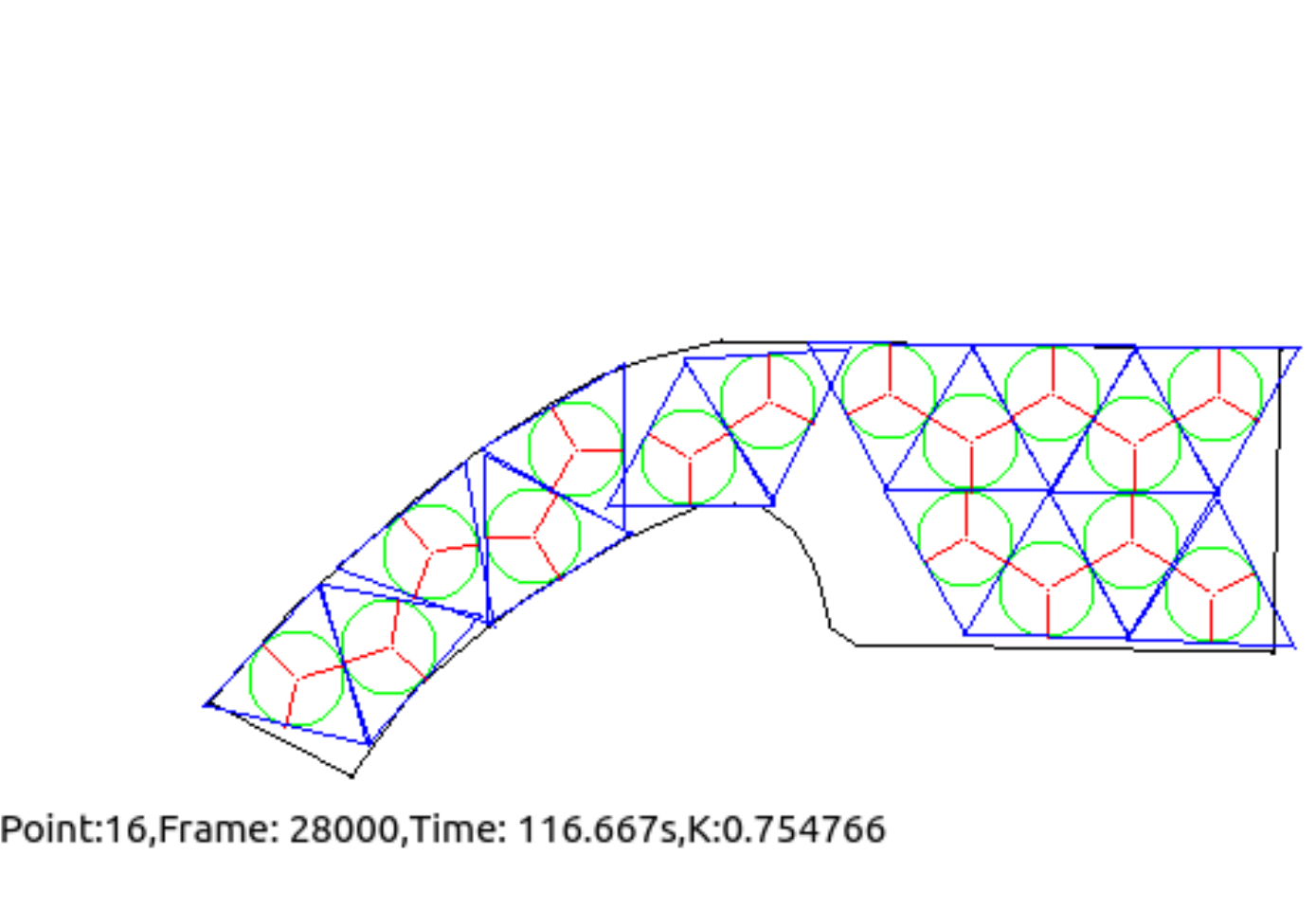}
    \label{fig:tr1_2:7}
}
\hspace{0\hsize}
\subfigure[]{
    \includegraphics[width=0.18\hsize]
        {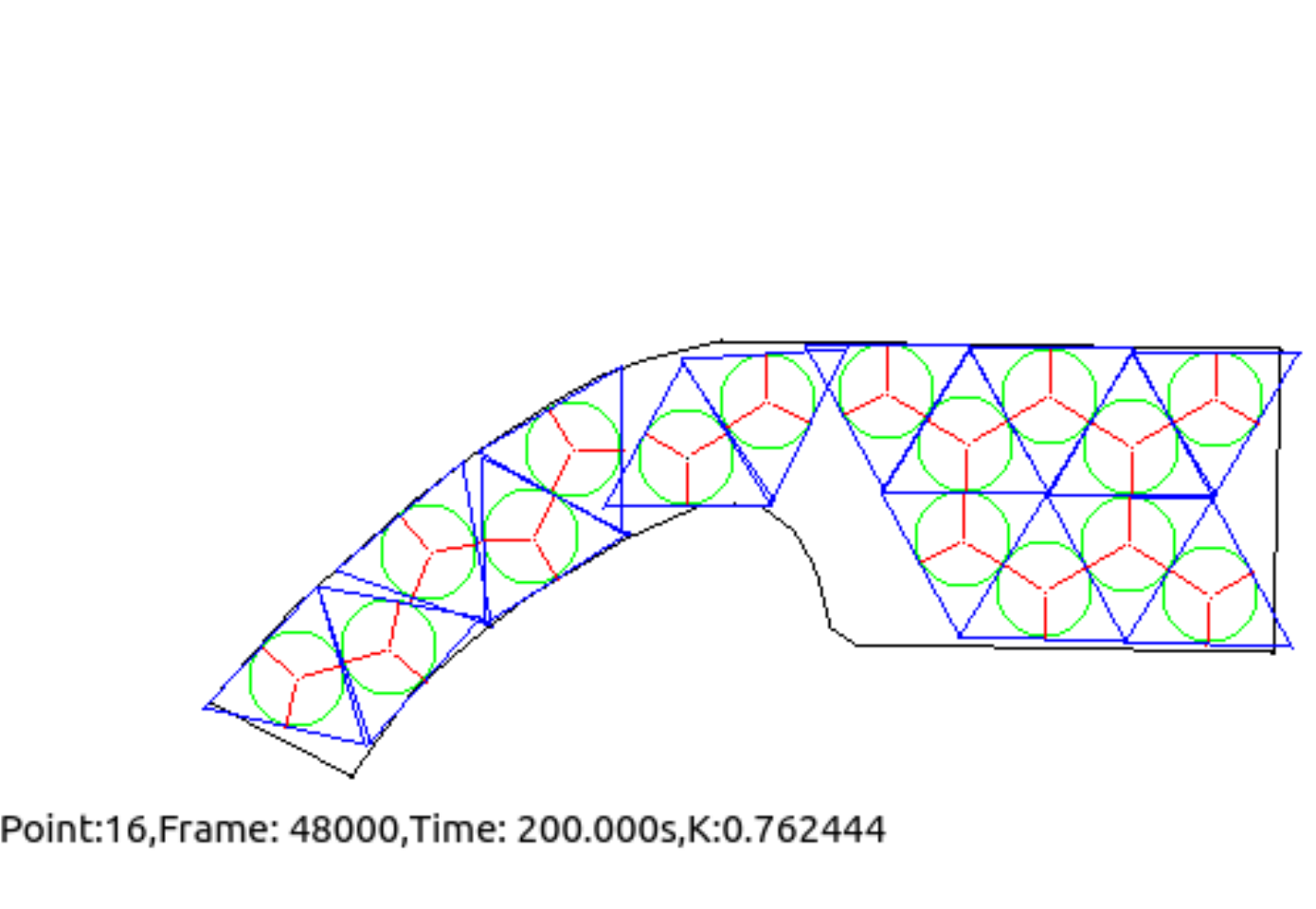}
    \label{fig:tr1_2:8}
}
\subfigure[]{
    \includegraphics[width=0.18\hsize]
        {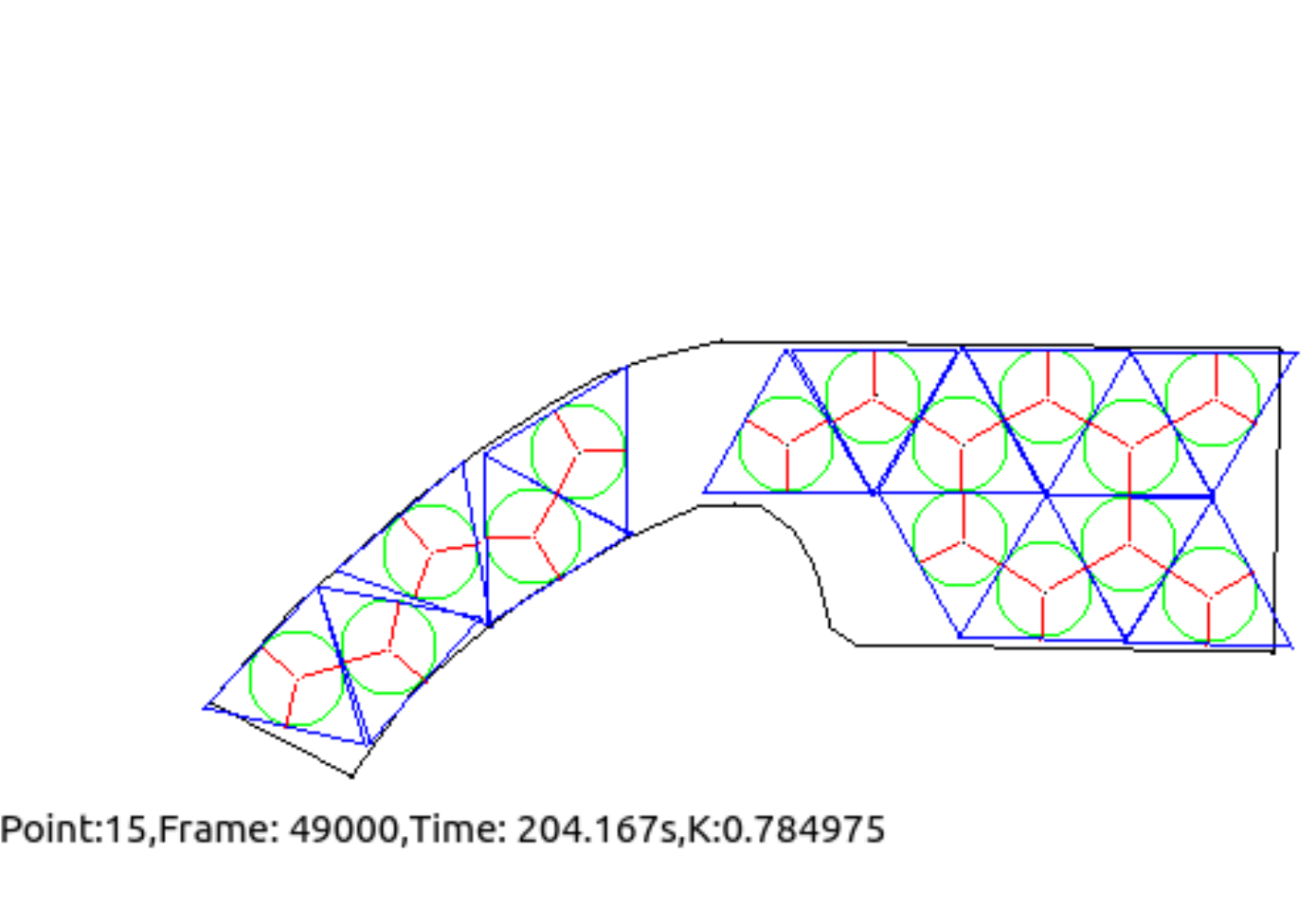}
    \label{fig:tr1_2:9}
}
\hspace{0\hsize}
\subfigure[]{
    \includegraphics[width=0.18\hsize]
        {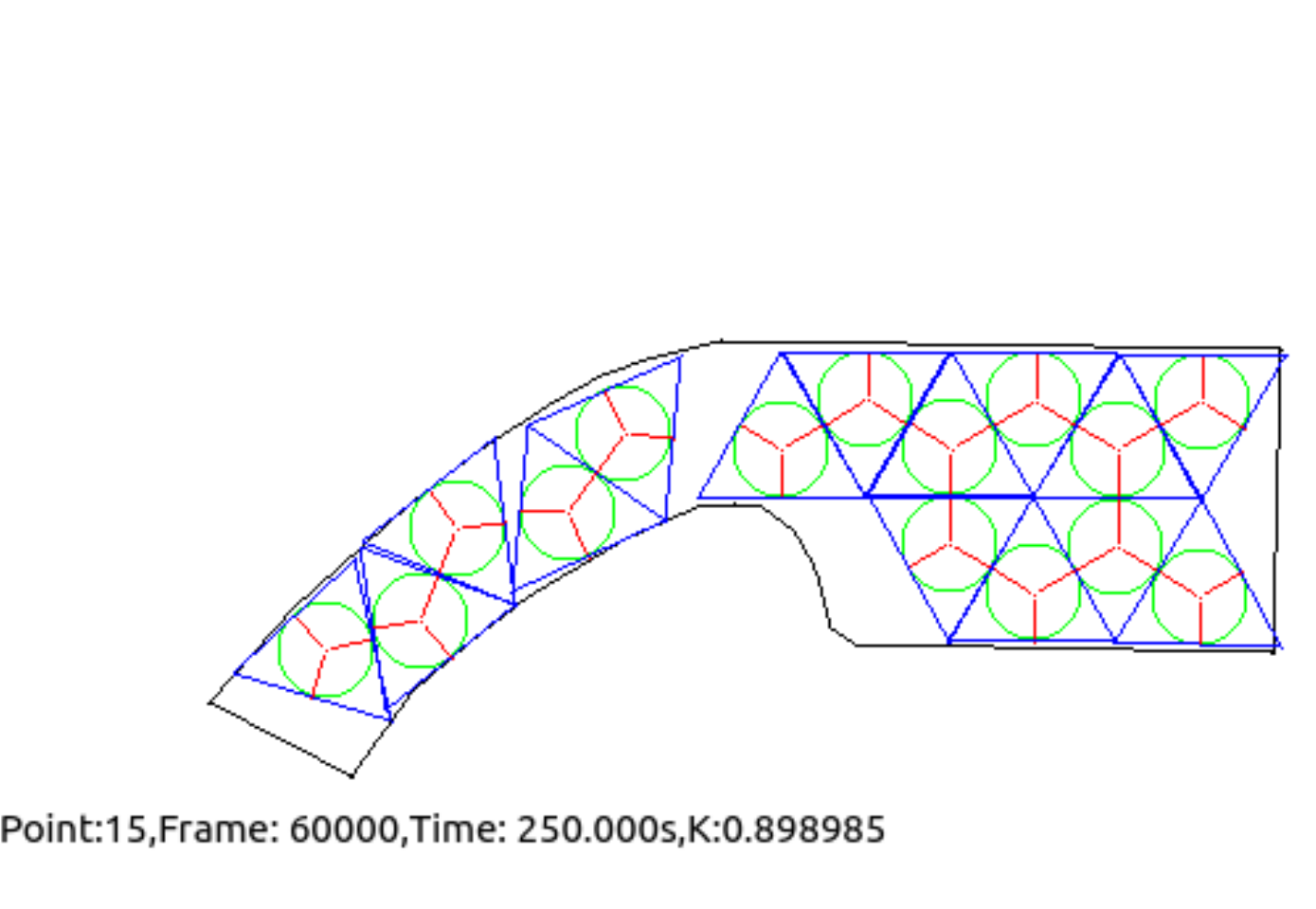}
    \label{fig:tr1_2:10}
}
\caption{The \textsc{ItPla}'s evolution process for the Schunk link's surface called $tr1\_2$.}
\label{fig:tr1_2}
\end{figure*}

Figure \ref{fig:tr1_2} shows a sequence of snapshots related to the coverage of $tr1\_2$.
As in the previous case, the number of triangular modules which -- in principle -- would cover at best the polygon is computed, and the corresponding number of modules is randomly placed within the polygon (Figure \ref{fig:tr1_2:0}).
As soon as the iterative process starts, we can observe that the leftmost and the rightmost regions of the polygon become soon well-covered, whereas overlaps occur in the central region (Figure \ref{fig:tr1_2:2}).
The algorithm alternates iterations necessary to stabilise the placement, which can be shown in Figure \ref{fig:tr1_2:2}, 
Figure \ref{fig:tr1_2:6}, and Figure \ref{fig:tr1_2:8}, with events necessary to remove modules, as shown in Figure \ref{fig:tr1_2:3}, Figure \ref{fig:tr1_2:5}, Figure \ref{fig:tr1_2:7}, and Figure \ref{fig:tr1_2:9}.
As a comparison with the results obtained using the approach described in \cite{Anghinolfietal2013}, \textsc{ItPla} generates two different patches (instead of one) for a total of $15$ triangular modules instead of $11$, thereby obtaining a $36.36\%$ coverage increase rate.

\begin{figure*}[t!]
\centering
\subfigure[]{
    \includegraphics[width=0.18\hsize]
        {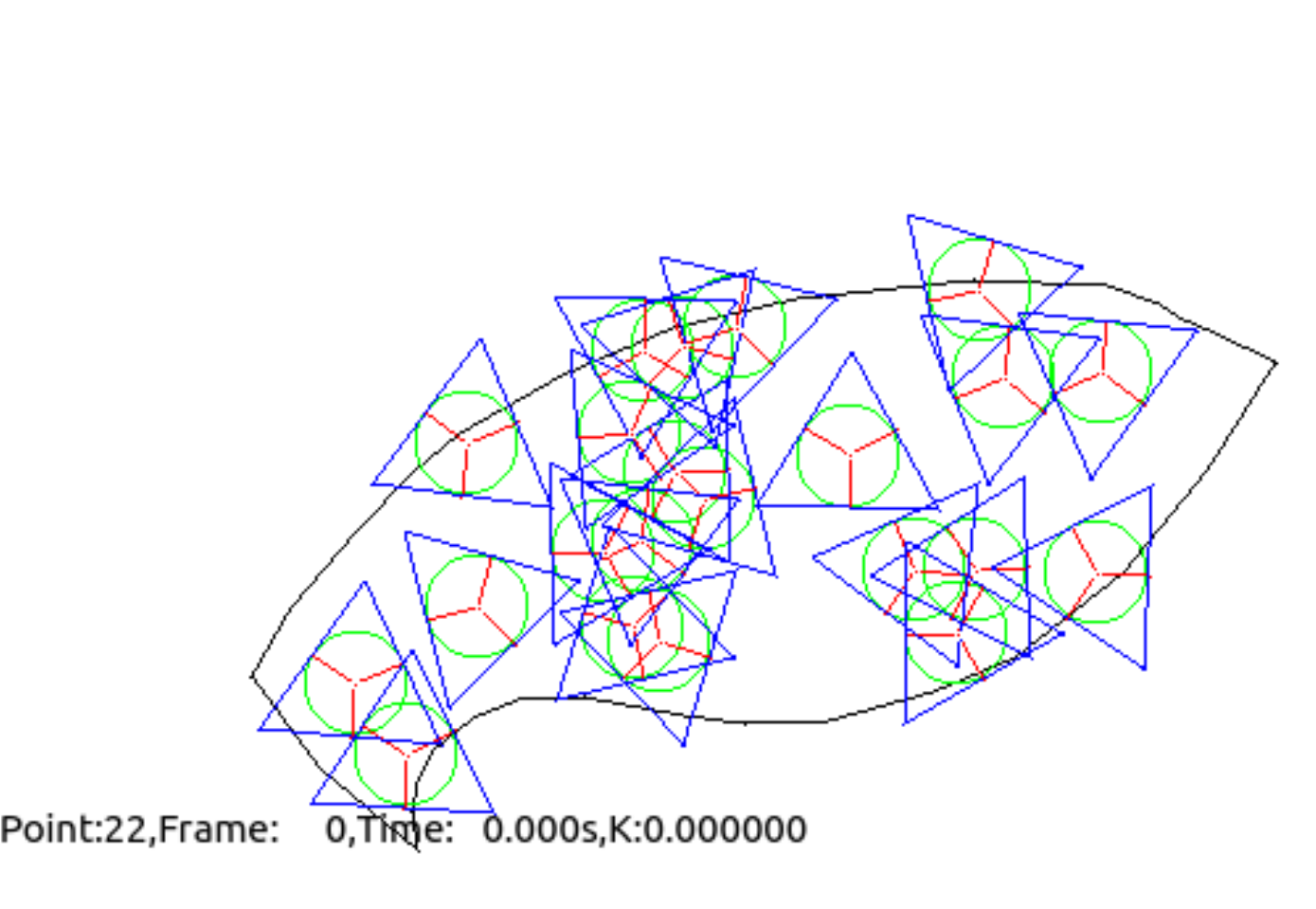}
    \label{fig:tr1_5:0}
}
\hspace{0\hsize}
\subfigure[]{
    \includegraphics[width=0.18\hsize]
        {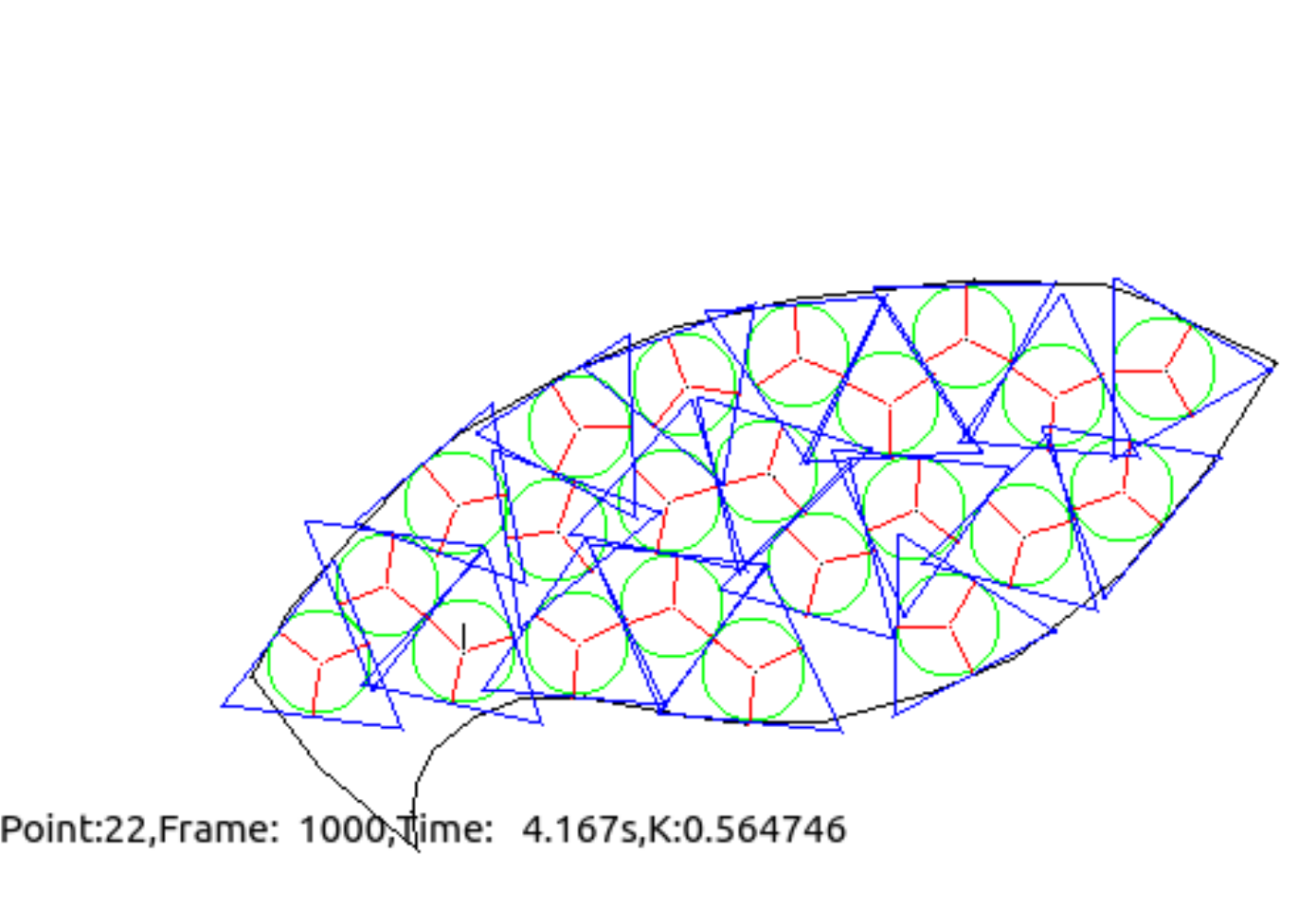}
    \label{fig:tr1_5:1}
}
\hspace{0\hsize}
\subfigure[]{
    \includegraphics[width=0.18\hsize]
        {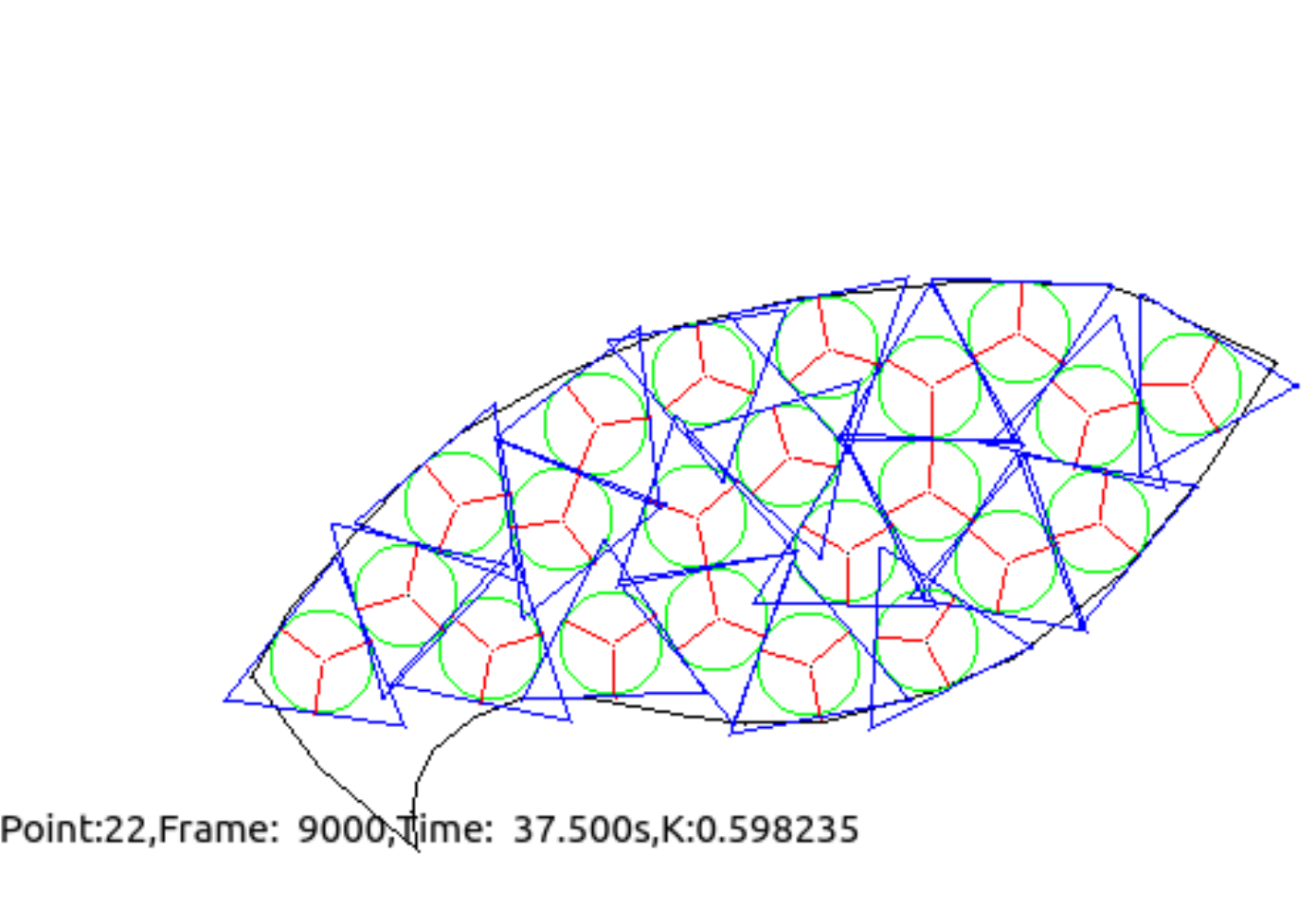}
    \label{fig:tr1_5:2}
}
\subfigure[]{
    \includegraphics[width=0.18\hsize]
        {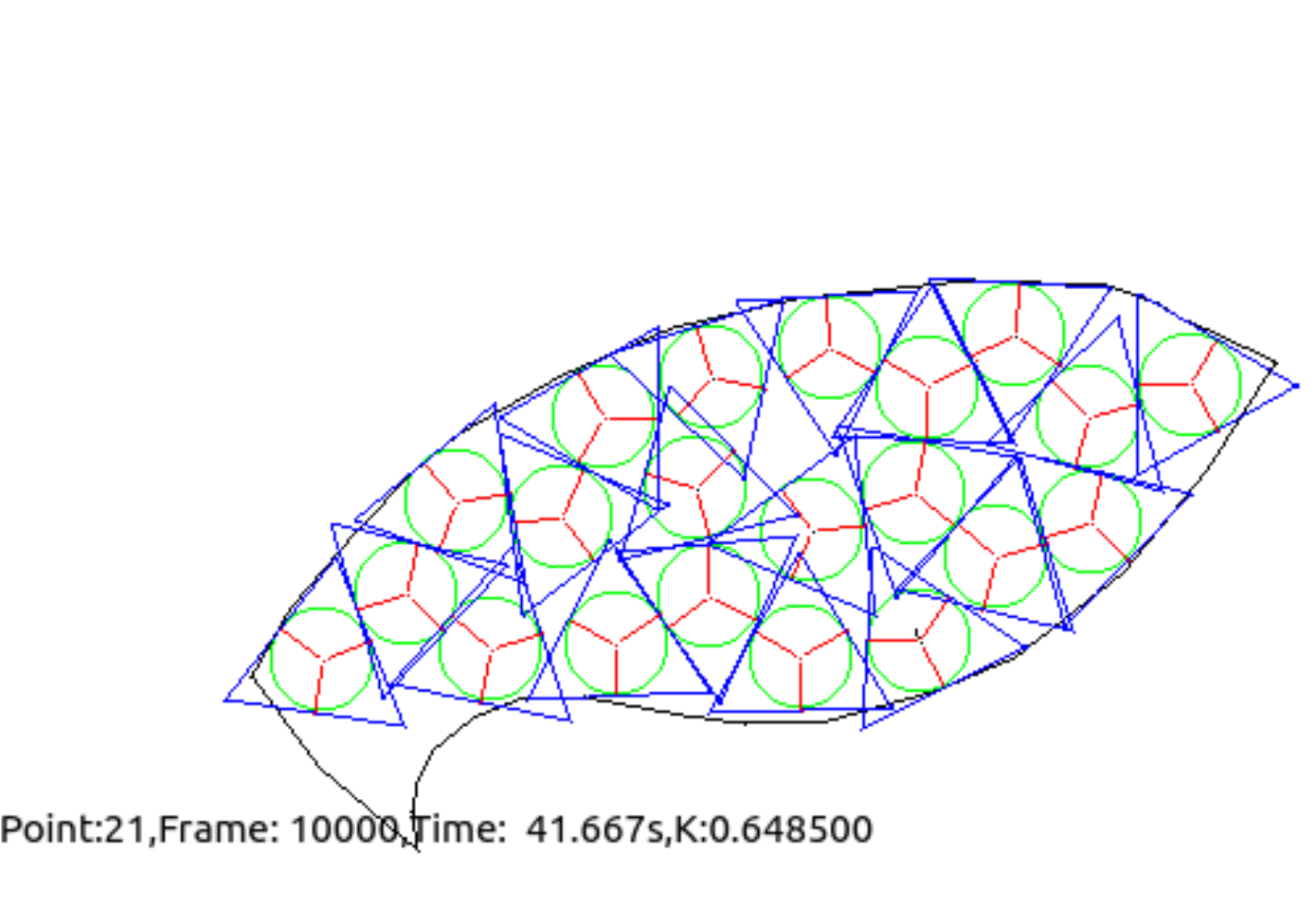}
    \label{fig:tr1_5:3}
}
\hspace{0\hsize}
\subfigure[]{
    \includegraphics[width=0.18\hsize]
        {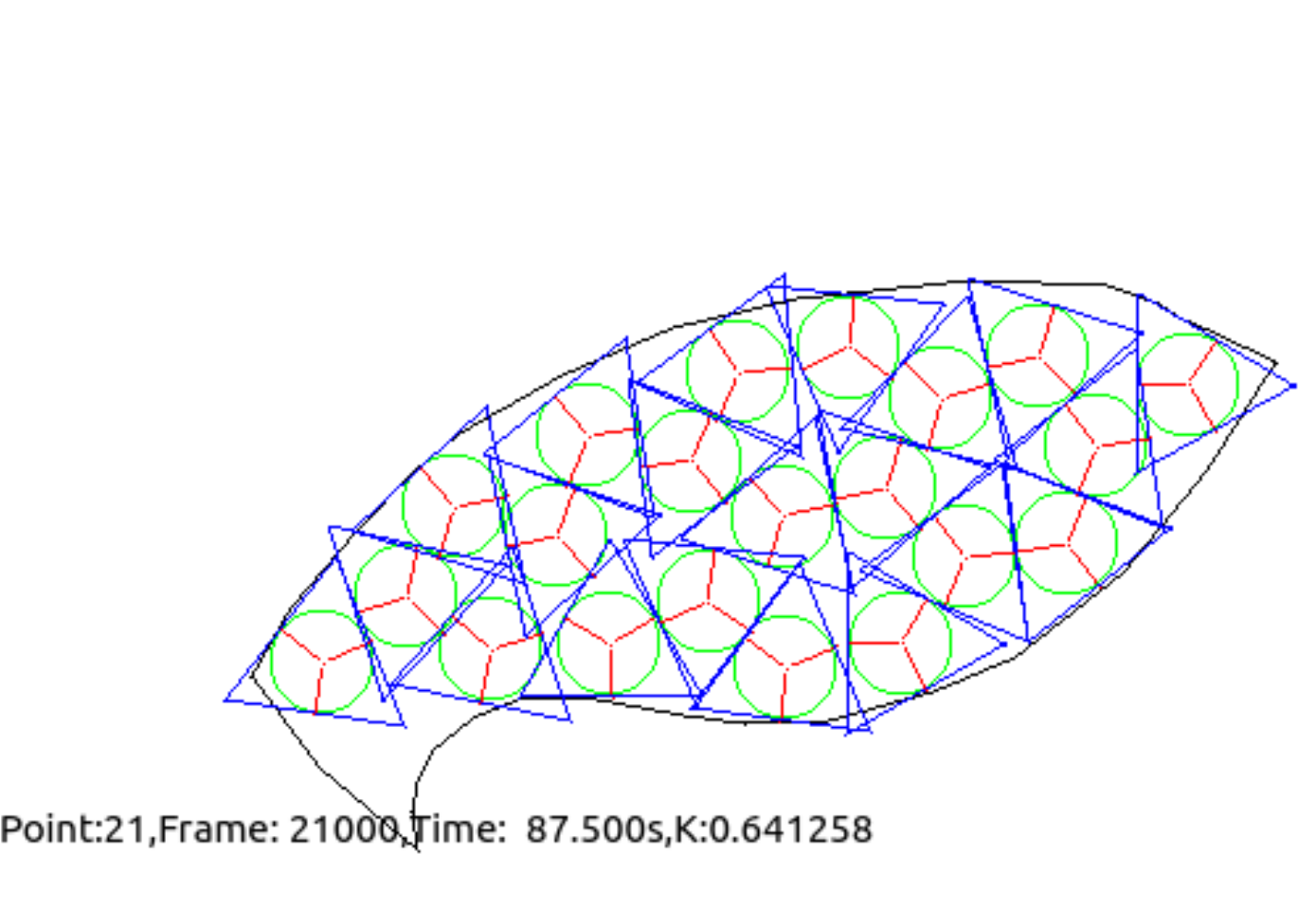}
    \label{fig:tr1_5:4}
}
\\
\hspace{0\hsize}
\subfigure[]{
    \includegraphics[width=0.18\hsize]
        {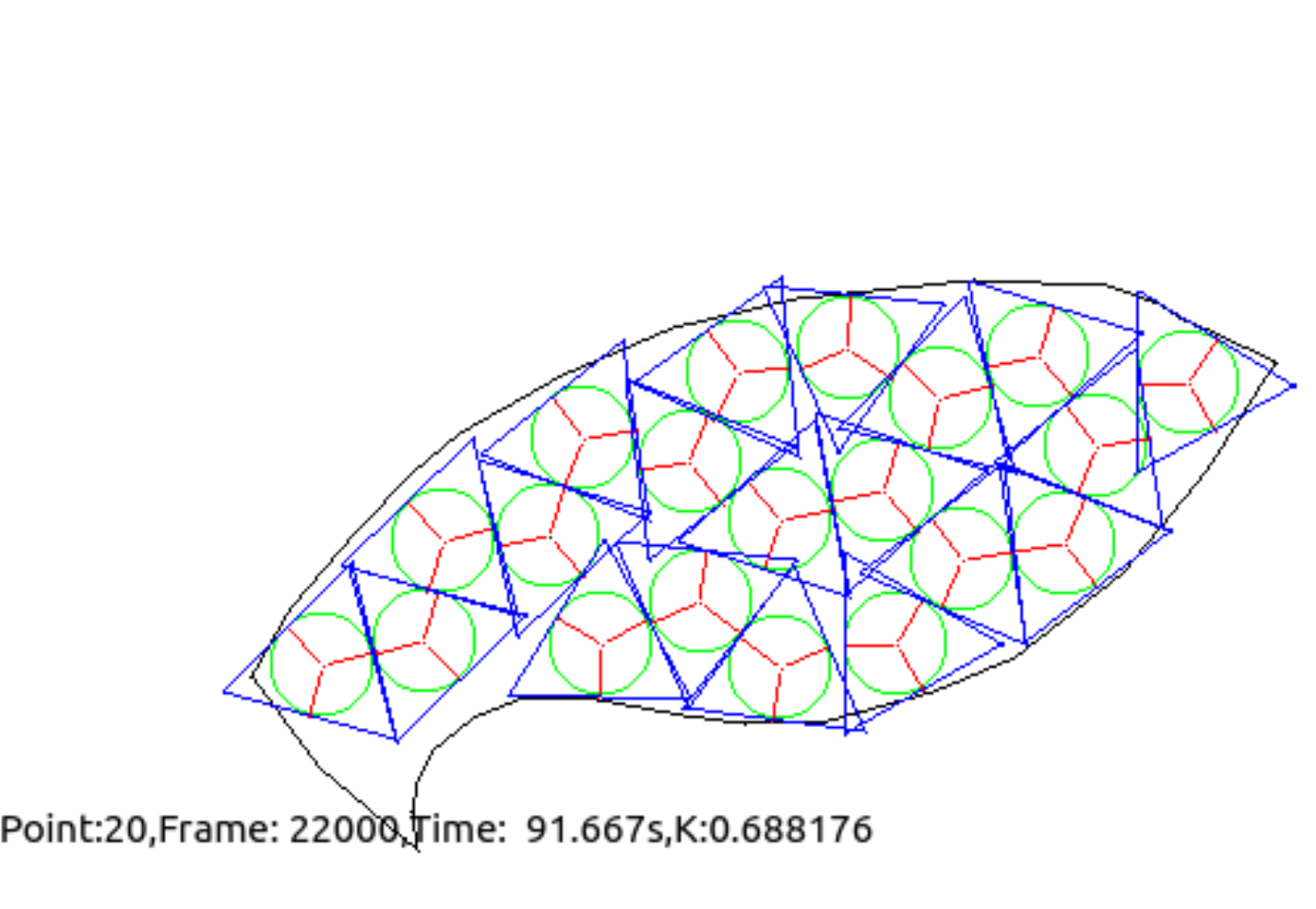}
    \label{fig:tr1_5:5}
}
\hspace{0\hsize}
\subfigure[]{
    \includegraphics[width=0.18\hsize]
        {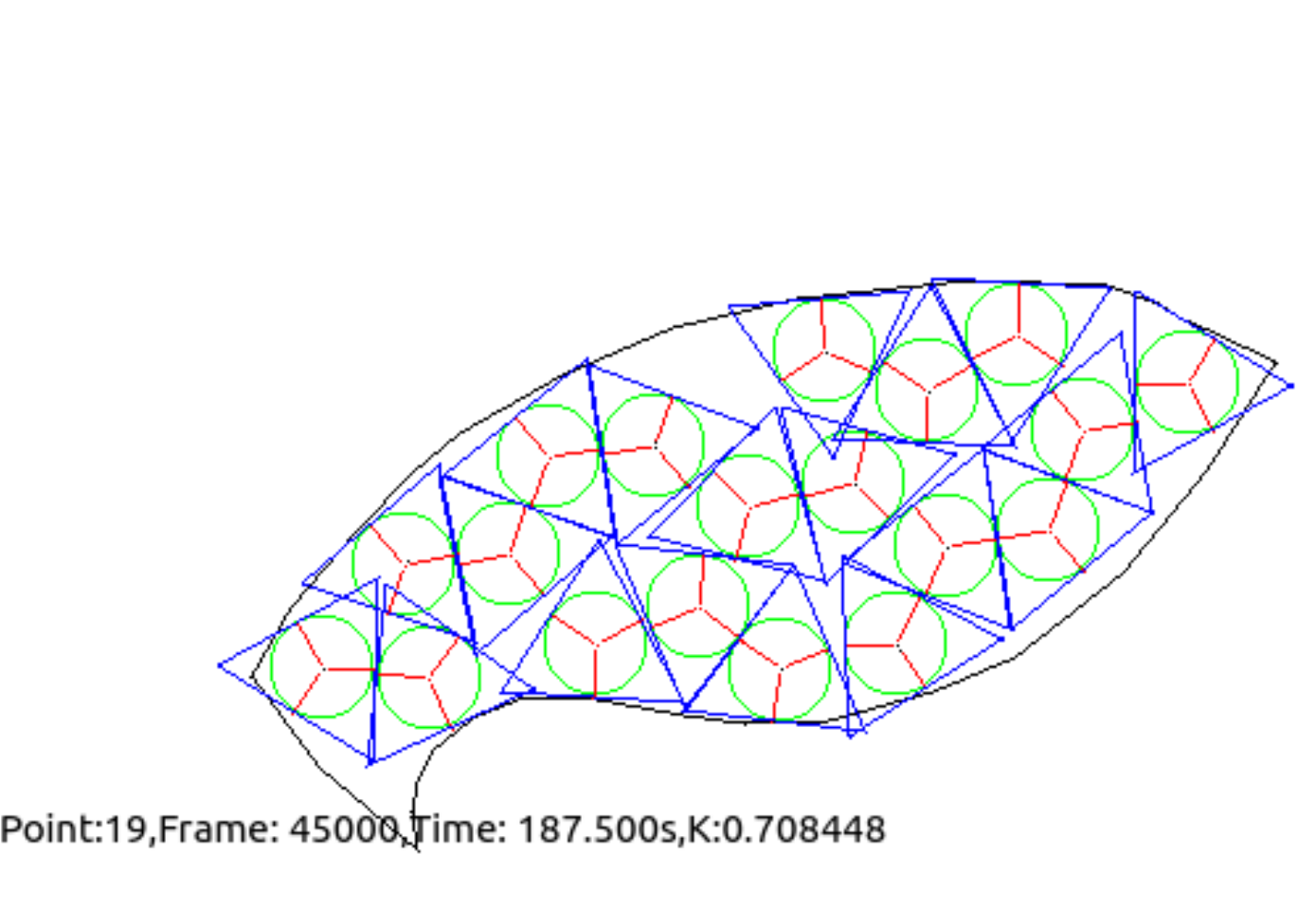}
    \label{fig:tr1_5:7}
}
\hspace{0\hsize}
\subfigure[]{
    \includegraphics[width=0.18\hsize]
        {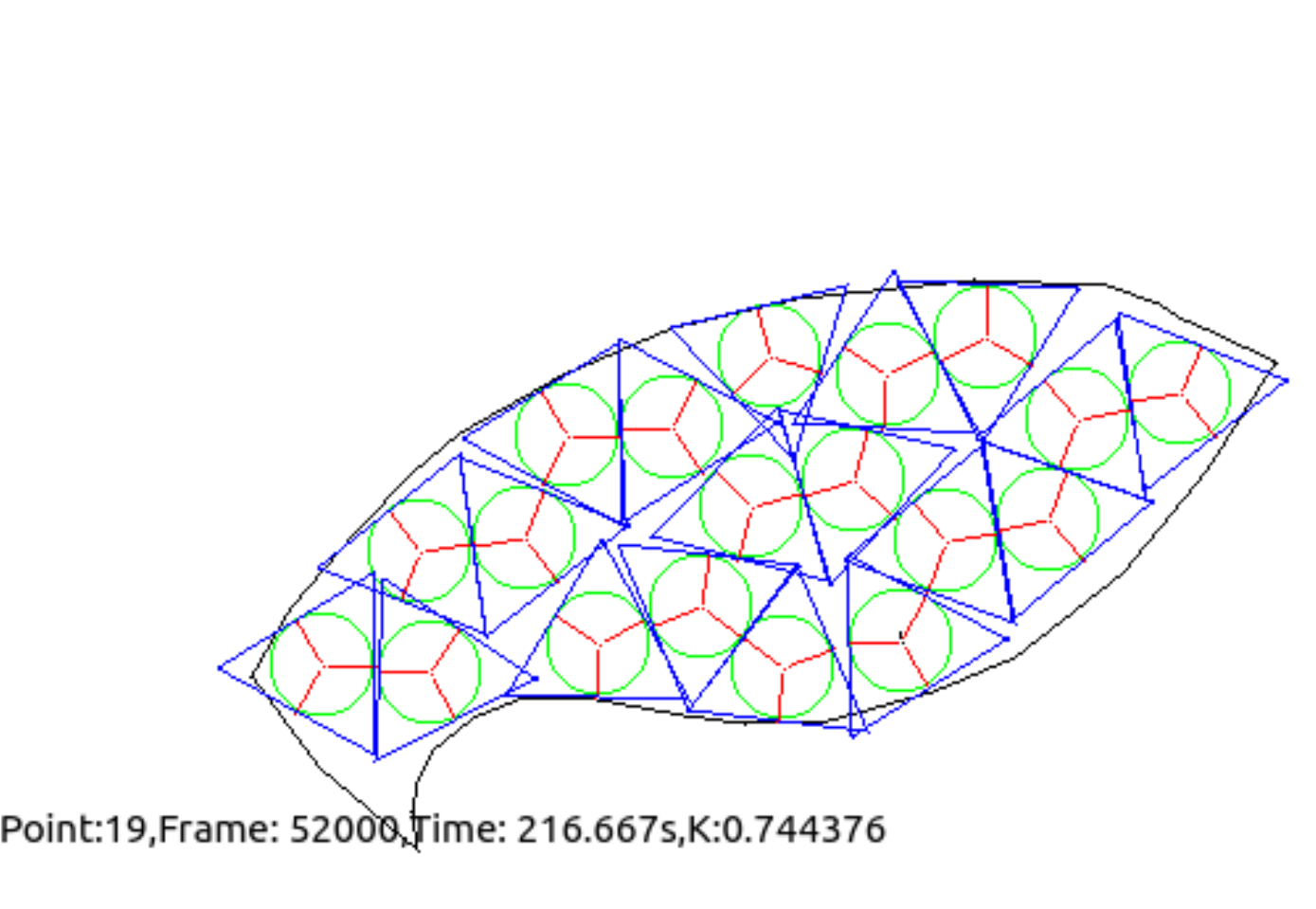}
    \label{fig:tr1_5:8}
}
\hspace{0\hsize}
\subfigure[]{
    \includegraphics[width=0.18\hsize]
        {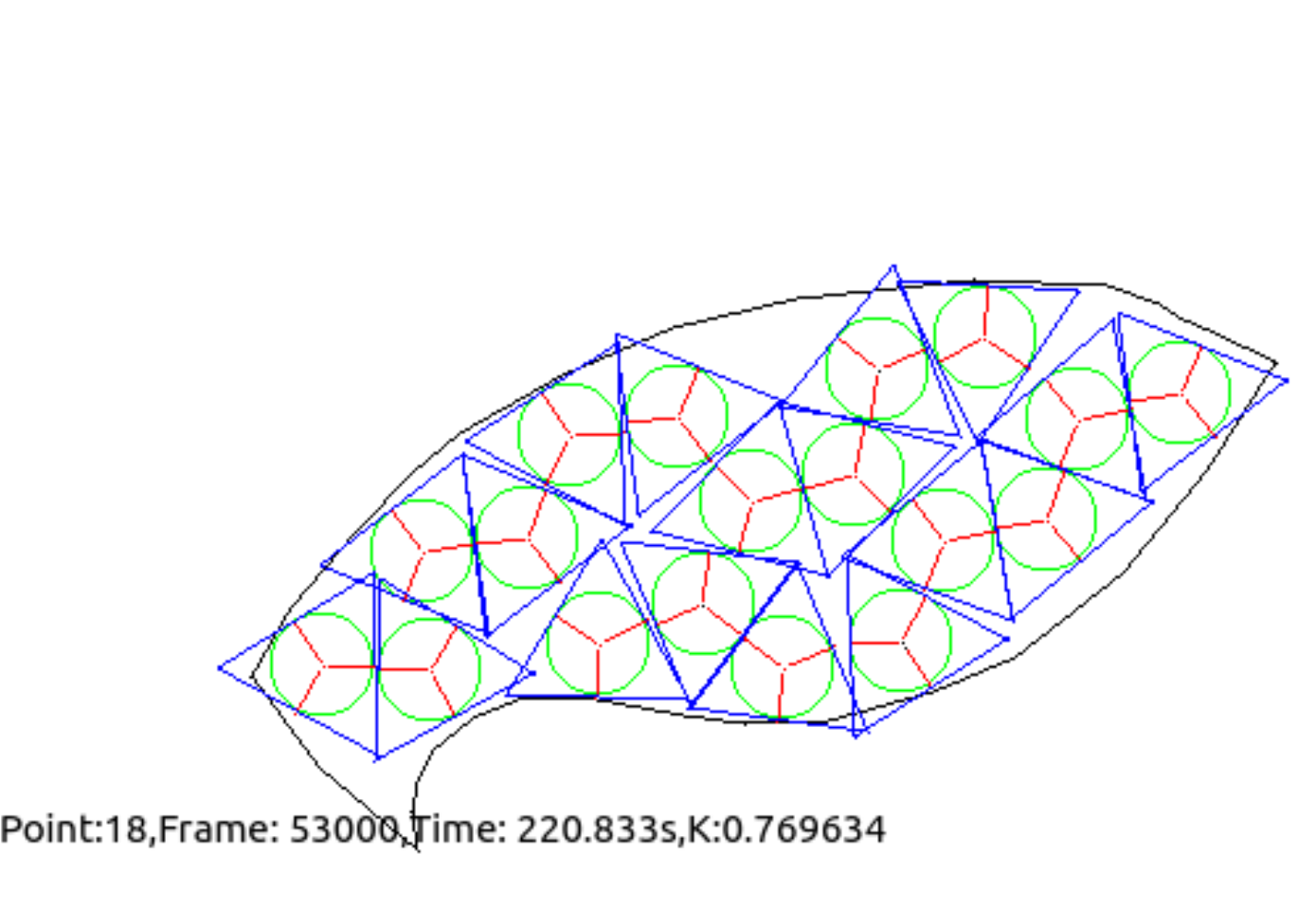}
    \label{fig:tr1_5:9}
}
\hspace{0\hsize}
\subfigure[]{
    \includegraphics[width=0.17\hsize]
        {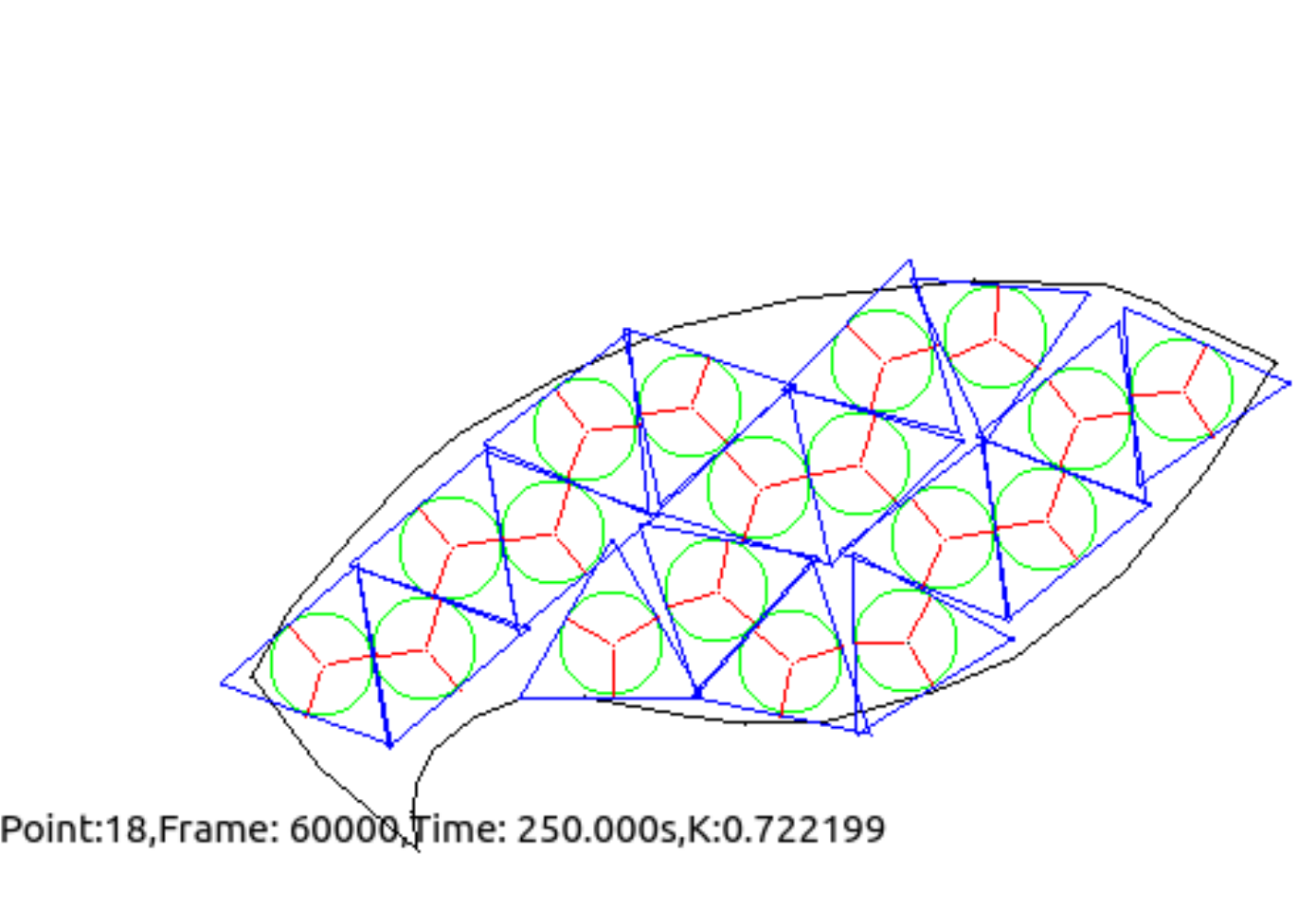}
    \label{fig:tr1_5:10}
}
\caption{The \textsc{ItPla}'s evolution process for the Schunk link's surface called $tr1\_5$.}
\label{fig:tr1_5}
\end{figure*}

Figure \ref{fig:tr1_5} shows a sequence of snapshots related to the coverage of $tr1\_5$.
Analogously to what happens in other cases, the iterative process alternates phases where pseudo-forces arrange triangular modules towards a stable placement, with others where one module is removed to decrease the overall overlap area.
In the case of Figure \ref{fig:tr1_5}, we observe an increase in coverage of about $20\%$ (from $15$ to $18$ modules) with respect to the baseline solution presented in \cite{Anghinolfietal2013}.

\subsection{Discussion}

The overall algorithm behaviour discussed in the previous sections leads us to a number of considerations about both the results and the possible follow-ups.

\begin{itemize}
\item
\textsc{ItPla} is able to generate placements where multiple patches are present.
Patches are distributed on the polygon in order to maximise the number of contained modules.
Attractive pseudo-forces seem to be appropriate to guarantee an intra-patch connectivity, whereas repulsive pseudo-forces are involved in separating patches. 
For different patches to be generated, a non-negligible effect is played by the polygon's boundaries.
This can be observed in the example involving $tr1\_2$.
In this case, the polygon is clearly divided in two regions: the leftmost part is narrower than the rightmost one, which causes an unsolvable overlap situation at the junction of the two regions.
Two patches are also generated in the polygon associated with the iCub's left hip, where the top left corner cannot accommodate a protrusion of the \textit{main} patch.
An irregular (although convex) shape like the one associated with $tr1\_5$ originates approximately one patch.  
\item
The patches generated by \textsc{ItPla} are not constrained to share the same regular grid structure, as it is required in the approach proposed in \cite{Anghinolfietal2013}.
Because of attractive pseudo-forces, triangular modules forming the same patch tend to be aligned along the principal directions of a regular isometric grid.
However, because of repulsive pseudo-forces and the effect of polygon's edges, different patches become aligned along different directions.
Therefore, each patch is aligned with respect to a different regular isometric grid.
This effect allows \textsc{ItPla} to be quite flexible in the process of placements generation and patches formation, and it is the main reason why it produces  results better than those obtained by the baseline approach discussed in \cite{Anghinolfietal2013}, above all in such cases as $tr1\_2$, where polygons are characterised by strong irregular shapes.  
\item
As it is possible to observe from the various snapshots in Figure \ref{fig:result_iCub}, Figure \ref{fig:tr1_2} and Figure \ref{fig:tr1_5}, the overall overlap area is not a monotone decreasing function of the iteration procedure.
On the contrary, as per effect of attractive and repulsive pseudo-forces, it can increase as long as the placement gets stable.
This behaviour can be observed also in Figure \ref{fig:chart} for the specific case of the iCub's left hip.
However, the overlap is reduced from now and then when a module is removed.
Module removals are of the utmost importance for the generation of stable and acceptable placements, as they drastically impact on the overlap area.
Furthermore, by initialising the number of modules to the theoretical upper bound subject to the polygon's area, we enforce (although this cannot be theoretically guaranteed) the maximum coverage.  
\item
Other module's shapes are possible, subject to the fact that they are radially symmetrical at least with respect to certain orientations.
For instance, circular modules are radially completely symmetrical, whereas triangular modules are radially symmetrical with respect to three equidistant radii. 
Since many of the \textsc{ItPla}'s properties are based on the abstract circular shape, from it many other shapes can be derived. 
One notable example would be the use of hexagonal shapes (which are radially symmetrical with respect to six radii), since a modular robot skin based on such a shape has been proposed in \cite{Kabolietal2015}.
Obviously enough, distance information as well as the effects of pseudo-forces should be modified to reflect the new mechanical constraints associated with the employed technology.
\item
Although we did not discuss it in the paper, and subject to meeting the possible associated mechanical constraints, module's sizes need not to be uniform, as it is instead implicitly posed by the authors of \cite{Anghinolfietal2013} with the introduction of an isometric grid. 
Modules with different sizes in the same placement are allowed in \textsc{ItPla}.
Such capability may prove to be useful in covering certain large-scale robot surfaces, where large regions would be covered by big modules and small regions by modules of a reduced size. 
\end{itemize}

\section{Conclusions}

In this paper, we present and discuss \textsc{ItPla}, an algorithm for the design of layouts related to the so-called robot skin placement problem.
The robot skin placement problem is a central issue to provide robots with the sense of touch, and in particular to attain robots with large-scale tactile sensing capabilities, which is fundamental in a wide range of service applications.

Differently from algorithms previously discussed in the literature, which are based on purely geometrical considerations, \textsc{ItPla} produces placements by the combined effect of a number of pseudo-forces.
Pseudo-forces locally translate and rotate modules to fit them inside the area of a polygon representing the robot's surface to cover.
The algorithm is characterised by a number of advantages when compared to previous approaches, namely: generating placements where multiple patches are present, which are not constrained to be aligned with respect to the same isometric grid, as well as accommodating for different module's shapes and sizes.
Given these features, \textsc{ItPla} proves to behave better than previous algorithms in the performance indicator of area coverage percentage.
The algorithm is available open source. 
%
%


\end{document}